\documentclass{article}
\usepackage{spconf,amsmath,graphicx}
\usepackage[ruled,linesnumbered]{algorithm2e}
\usepackage{dsfont}

\usepackage[utf8]{inputenc} 
\usepackage[T1]{fontenc}    
\usepackage{hyperref}       
\usepackage{url}            
\usepackage{booktabs}       
\usepackage{amsfonts}       
\usepackage{nicefrac}       
\usepackage{microtype}      
\usepackage{xcolor}         
\usepackage{amsmath}
\usepackage{amssymb}
\usepackage{mathtools}
\usepackage{amsthm}
\usepackage[capitalize,noabbrev]{cleveref}
\usepackage{graphicx}
\usepackage{aliascnt}
\usepackage{float}
\usepackage{xr-hyper}
\usepackage{hyperref}
\usepackage{todonotes}
\usepackage{varwidth}
\usepackage{multicol}
\usepackage{changepage}

\usepackage{tikz-timing}[2009/12/09]
\usetikztiminglibrary[new={char=Q,reset char=R}]{counters}

\usepackage{autonum}
\makeatletter
\newtheorem{theorem}{Theorem}
\crefname{theorem}{theorem}{Theorems}
\Crefname{Theorem}{Theorem}{Theorems}

\newaliascnt{lemma}{theorem}
\newtheorem{lemma}[lemma]{Lemma}
\aliascntresetthe{lemma}
\crefname{lemma}{lemma}{lemmas}
\Crefname{Lemma}{Lemma}{Lemmas}

\newaliascnt{corollary}{theorem}
\newtheorem{corollary}[corollary]{Corollary}
\aliascntresetthe{corollary}
\crefname{corollary}{corollary}{corollaries}
\Crefname{Corollary}{Corollary}{Corollaries}

\newaliascnt{proposition}{theorem}

\aliascntresetthe{proposition}
\crefname{proposition}{proposition}{propositions}
\Crefname{Proposition}{Proposition}{Propositions}

\newaliascnt{definition}{theorem}

\aliascntresetthe{definition}
\crefname{definition}{definition}{definitions}
\Crefname{Definition}{Definition}{Definitions}

\newtheorem{assumption}{\textbf{A}\hspace{-3pt}}
\Crefname{assumption}{\textbf{A}\hspace{-3pt}}{\textbf{A}\hspace{-3pt}}
\crefname{assumption}{\textbf{A}}{\textbf{A}}

\newaliascnt{remark}{theorem}
\newtheorem{remark}[remark]{Remark}
\aliascntresetthe{remark}
\crefname{remark}{remark}{remarks}
\Crefname{Remark}{Remark}{Remarks}

\crefname{example}{example}{examples}
\Crefname{Example}{Example}{Examples}

\crefname{algorithm}{algorithm}{algorithms}
\Crefname{Algorithm}{Algorithm}{Algorithms}

\crefname{figure}{figure}{figures}
\Crefname{Figure}{Figure}{Figures}

\usepackage{caption}
\usepackage{subcaption}
\usepackage{multirow}

\def\rset{\mathbb{R}}

\def\nset{\mathbb{N}}

\def\param{w}

\def\Algo{\texttt{FAVANO}}

\newcommand{\Set}{\mathcal{S}}
\def\Geom{\operatorname{Geom}}

\def\var{\operatorname{Var}}

\def\objfunc{R}

\usepackage{xargs}
\newcommandx{\CPE}[3][1=]{\PE_{#1}\left[\left. #2 \, \right| #3 \right]}

\def\PE{\mathbb{E}}

\newcommand{\ps}[2]{\left\langle#1\,,\,#2\right\rangle}

\title{\Algo: Federated AVeraging with Asynchronous Nodes}
\name{Louis Leconte$^{1, 2}$ \qquad Van Minh Nguyen$^{2}$ \qquad Eric Moulines$^{3}$}
\address{$^{1}$ Lisite, Isep, Sorbonne University, $^{3}$ CMAP, Ecole Polytechnique \\$^{2}$ Mathematical and Algorithmic Sciences Lab, Huawei Technologies}
\begin{document}
%
\maketitle
\begin{abstract}
In this paper, we propose a novel centralized Asynchronous Federated Learning (FL) framework, \Algo\, for training Deep Neural Networks (DNNs) in resource-constrained environments. Despite its popularity, ``classical'' federated learning faces the increasingly difficult task of scaling synchronous communication over large wireless networks. Moreover, clients typically have different computing resources and therefore computing speed, which can lead to a significant bias (in favor of ``fast'' clients) when the updates are asynchronous. Therefore, practical deployment of FL requires to handle users with strongly varying computing speed in communication/resource constrained setting.  We provide convergence guarantees for \Algo\ in a smooth, non-convex environment and carefully compare the obtained convergence guarantees with existing bounds, when they are available. Experimental results show that the \Algo\ algorithm outperforms current methods on standard benchmarks.
\end{abstract}

\section{Introduction}
\label{sec:intro}

Federated learning, a promising approach for training models from networked agents, involves the collaborative aggregation of locally computed updates, such as parameters, under centralized orchestration \cite{konevcny2015federated, mcmahan2017communication, kairouz2021advances}. The primary motivation behind this approach is to maintain privacy, as local data is never shared between agents and the central server \cite{zhao2018federated, horvath2022fedshuffle}. However, communication of training information between edge devices and the server is still necessary. The central server aggregates the local models to update the global model, which is then sent back to the devices. Federated learning helps alleviate privacy concerns, and it distributes the computational load among networked agents. However, each agent must have more computational power than is required for inference, leading to a computational power bottleneck. This bottleneck is especially important when federated learning is used in heterogeneous, cross-device applications.

Most approaches to centralized federated learning (FL) rely on synchronous operations, as assumed in many studies \cite{mcmahan2017communication, wang2021field}. At each global iteration, a copy of the current model is sent from the central server to a selected subset of agents. The agents then update their model parameters using their private data and send the model updates back to the server. The server aggregates these updates to create a new shared model, and this process is repeated until the shared model meets a desired criterion. However, device heterogeneity and communication bottlenecks (such as latency and bandwidth) can cause delays, message loss, and stragglers, and the agents selected in each round must wait for the slowest one before starting the next round of computation. This waiting time can be significant, especially since nodes may have different computation speeds.

To address this challenge, researchers have proposed several approaches that enable asynchronous communication, resulting in improved scalability of distributed/federated learning \cite{xie2019asynchronous, chen2020asynchronous, chen2021towards, xu2021asynchronous}. In this case, the central server and local agents typically operate with inconsistent versions of the shared model, and synchronization in lockstep is not required, even between participants in the same round. As a result, the server can start aggregating client updates as soon as they are available, reducing training time and improving scalability in practice and theory.

\textbf{Contributions.} Our work takes a step toward answering this question by introducing \Algo, a centralized federated learning algorithm designed to accommodate clients with varying computing resources and support asynchronous communication.
\begin{itemize} \item In this paper, we introduce a new algorithm called \Algo\ that uses an unbiased aggregation scheme for centralized federated learning with asynchronous communication. Our algorithm does not assume that clients computed the same number of epochs while being contacted, and we give non-asymptotic complexity bounds for \Algo\ in the smooth nonconvex setting. We emphasize that the dependence of the bounds on the total number of agents $n$ is improved compared to \cite{zakerinia2022quafl} and does not depend on a maximum delay. 
\item Experimental results show that our approach consistently outperforms other asynchronous baselines on the challenging TinyImageNet dataset \cite{le2015tiny}.
\end{itemize}
Our proposed algorithm \Algo\ is designed to allow clients to perform their local steps independently of the server's round structure, using a fully local, possibly outdated version of the model. Upon entering the computation, all clients are given a copy of the global model and perform at most $K \geq 1$ optimization steps based on their local data. The server randomly selects a group of $s$ clients in each server round, which, upon receiving the server's request, submit an \emph{unbiased} version of their progress. Although they may still be in the middle of the local optimization process, they send reweighted contributions so that fast and slow clients contribute equally. The central server then aggregates the models and sends selected clients a copy of the current model. The clients take this received server model as a new starting point for their next local iteration. 

\section{Related Works}
\label{sec:relatedworks}
Federated Averaging (FedAvg), also known as local SGD, is a widely used approach in federated learning. In this method, each client updates its local model using multiple steps of stochastic gradient descent (SGD) to optimize a local objective function. The local devices then submit their model updates to the central server for aggregation, and the server updates its own model parameters by averaging the client models before sending the updated server parameters to all clients. FedAvg has been shown to achieve high communication efficiency with infrequent synchronization, outperforming distributed large mini-batches SGD \cite{lin2019don}.

However, the use of multiple local epochs in FedAvg can cause each device to converge to the optima of its local objective rather than the global objective, a phenomenon known as client drift. This problem has been discussed in previous work; see \cite{karimireddy2020scaffold}. Most of these studies have focused on synchronous federated learning methods, which have a similar update structure to FedAvg \cite{wang2020tackling, karimireddy2020scaffold, qu2021feddq, makarenko2022adaptive, mao2022communication, tyurin2022dasha}. However, synchronous methods can be disadvantageous because they require all clients to wait when one or more clients suffer from high network delays or have more data, and require a longer training time. This results in idle time and wasted computing resources.

Moreover, as the number of nodes in a system increases, it becomes infeasible for the central server to perform synchronous rounds among all participants, and synchrony can degrade the performance of distributed learning. A simple approach to mitigate this problem is node sampling, e.g. \cite{smith2017federated, bonawitz2019towards}, where the server only communicates with a subset of the nodes in a round. But if the number of stragglers is large, the overall training process still suffers from delays.

Synchronous FL methods are prone to stragglers. One important research direction is based on FedAsync \cite{xie2019asynchronous} and subsequent works. The core idea is to update the global model immediately when the central server receives a local model. However, when staleness is important, performance is similar to FedAvg, so it is suboptimal in practice. ASO-Fed \cite{chen2020asynchronous} proposes to overcome this problem and handles asynchronous FL with local streaming data by introducing memory-terms on the local client side. AsyncFedED \cite{wang2022asyncfeded} also relies on the FedAsync instantaneous update strategy and also proposes to dynamically adjust the learning rate and the number of local epochs to staleness. Only one local updated model is involved in FedAsync-like global model aggregations. As a result, a larger number of training epochs are required and the frequency of communication between the server and the workers increases greatly, resulting in massive bandwidth consumption. From a different perspective, QuAFL \cite{zakerinia2022quafl} develops a concurrent algorithm that is closer to the FedAvg strategy. QuAFL incorporates both asynchronous and compressed communication with convergence guarantees. Each client must compute $K$ local steps and can be interrupted by the central server at any time. The client updates its model with the (compressed) central version and its current private model. The central server randomly selects $s$ clients and updates the model with the (compressed) received local progress (since last contact) and the previous central model. QuAFL works with old variants of the model at each step, which slows convergence. However, when time, rather than the number of server rounds, is taken into account, QuAFL can provide a speedup because the asynchronous framework does not suffer from delays caused by stragglers. A concurrent and asynchronous approach aggregates local updates before updating the global model: FedBuff \cite{nguyen2022federated} addresses asynchrony using a buffer on the server side. Clients perform local iterations, and the base station updates the global model only after $Z$ different clients have completed and sent their local updates. The gradients computed on the client side may be stale. The main assumption is that the client computations completed at each step come from a uniform distribution across all clients. Fedbuff is asynchronous, but is also sensitive to stragglers (must wait until $Z$ different clients have done all local updates). Similarly, \cite{koloskova2022sharper} focus on Asynchronous SGD, and provide  guarantees depending on some $\tau_{max}$. Similar to \cite{nguyen2022federated} the algorithm is also impacted by stragglers, during the transitional regime at least. A recent work by \cite{fraboni2023general} extend the idea of \cite{koloskova2022sharper} by allowing multiple clients to contribute in one round. But this scheme also favors fast clients. \cite{liu2021adaptive} does not run on buffers, but develops an Adaptive Asynchronous Federated Learning (AAFL) mechanism to deal with speed differences between local devices. Similar to FedBuff, in \cite{liu2021adaptive}'s method, only a certain fraction of the locally updated models contribute to the global model update. Most convergence guarantees for asynchronous distributed methods depend on staleness or gradient delays \cite{nguyen2022federated, toghani2022unbounded, koloskova2022sharper}. Only \cite{mishchenko2022asynchronous} analyzes the asynchronous stochastic gradient descent (SGD) independently of the delays in the gradients. However, in the heterogeneous (non-IID) setting, convergence is proved up to an additive term that depends on the dissimilarity limit between the gradients of the local and global objective functions.

\section{Algorithm}
\label{sec:algo}
We consider optimization problems in which the components of the objective function (i.e., the data for machine learning problems) are distributed over $n$ clients, i.e.,
\begin{equation}
\label{eq:optimization-problem}
\min_{\param \in \rset^d} \objfunc(\param); \, \, \objfunc(\param)=\frac{1}{n} \sum_{i=1}^n \PE_{(x, y) \sim p_{\operatorname{data }}^i}[\ell(\mathrm{NN}(x, \param), y)],
\end{equation}
where $d$ is the number of parameters (network weights and biases), $n$ is the total number of clients, $\ell$ is the training loss (e.g., cross-entropy or quadratic loss), $\mathrm{NN}(x,\param)$ is the DNN prediction function, $p_{\operatorname{data}}^i$ is the training distribution on client $i$. In FL, the distributions $p_{\operatorname{data}}^i$ are allowed to differ between clients (statistical heterogeneity).

Each client maintains three key values in its local memory: the local model $w^i$, a counter $q^i$, and the value of the initial model with which it started the iterations $\param^i_{init}$. The counter $q^i$ is incremented for each SGD step the client performs locally until it reaches $K$, at which point the client stops updating its local model and waits for the server request. Upon the request to the client $i$, the local model and counter~$q^i$ are reset. If a server request occurs before the $K$ local steps are completed, the client simply pauses its current training process, reweights its gradient based on the number of local epochs (defined by $E^i_{t+1}$), and sends its current \emph{reweighted} model to the server.
 
In \cite{zakerinia2022quafl}, we identified the client update $\param^i_{} = \frac{1}{s+1}\param_{t-1} + \frac{s}{s+1} {\param}^i_{}$ as a major shortcoming. When the number of sampled clients $s$ is large enough, $\frac{s}{s+1}\param^i$ dominates the update and basically no server term are taken into consideration. This leads to a significant client drift. As a consequence, QuAFL does not perform well in the heterogeneous case (see \Cref{sec:experiments}). Second, one can note that the updates in QuAFL are biased in favor of fast clients. Indeed each client computes gradients at its own pace and can reach different numbers of epochs while being contacted by the central server. It is assumed that clients compute the \emph{same} number of local epochs in the analysis from \cite{zakerinia2022quafl}, but it is not the case in practice. As a consequence, we propose \Algo\ to deal with asynchronous updates without favoring fast clients. A first improvement is to update local weight directly with the received central model. Details can be found in \Cref{algo:flau}. Another idea to tackle gradient unbiasedness is to reweight the contributions from each of the $s$ selected clients: these can be done either by dividing by the (proper) number of locally computed epochs, or by the expected value of locally computed epochs. In practice, we define the reweight $\alpha^i = \PE[E^i_{t+1} \wedge K]$, or $\alpha^i = \mathbf{P}(E^i_{t+1}>0)(E^i_{t+1} \wedge K)$, where $\wedge$ stands for $\min$.
\begin{algorithm}[!t]
\SetKwInOut{Input}{Input}
\SetKwInOut{Output}{Output}
\SetKwBlock{Loop}{Loop}{end}
\SetKwBlock{Initialize}{Initialize}{end}
\SetKwFor{When}{When}{do}{end}
\SetKwFunction{Wait}{Wait}
\SetKwFunction{ClientMain}{ClientLocalTraining}
\SetAlgoLined
\SetInd{0.1em}{0.5em}
\Input{Number of steps $T$, LR $\eta$, Selection Size $s$, Maximum local steps $K$ \;}
\tcc{\textbf{At the Central Server}}
\Initialize{
Initialize parameters $w_0$\;
Server sends $w_0$ to all clients\;
}
\For{$t=1,\dots,T$}{
Generate set $\Set_t$ of $s$ clients uniformly at random\;
\For{all clients $i \in \Set_t$}{
Server receives $\textcolor{red}{{\param}^i_{unbiased}}$ from client $i$\;
}
Update central server model $\param_{t} \gets \frac{1}{s+1}\param_{t-1} + (\frac{1}{s+1}\sum_{i \in \Set_t}  \textcolor{red}{{\param}^i_{unbiased}})$\; \label{algo:line:server_update}
\For{all clients $i \in \Set_t$}{
Server sends $\param_t$ to client $i$\; \label{algo:line:server_multicast}
}
}

\tcc{\textbf{At Client $i$}}
\Initialize{
Client receives $w_0$ and $K$ from the Server\;
\textbf{Local variables} $\param^i = \param_0, q^i=0$\;
}
\Loop{
Run \ClientMain{} concurrently\;
\When{Contacted by the Server}{
Interrupt \ClientMain{}\;
Define $\textcolor{red}{\alpha^i}$ following the reweighting strategy \;
Send $w^i_{unbiased} := \param^i_{init} + \textcolor{red}{\frac{1}{\alpha^i}}(\param^i - \param^i_{init})$ to the server\;
Receive $w_t$ from the server\;
Update $w^i_{init} \gets w_t, \textcolor{red}{w^i \gets w_t}, q^i \gets 0$\;
Restart \ClientMain{} from zero with updated variables\;
}
}
\SetKwProg{Fn}{function}{:}{}
\Fn{\ClientMain{}}{
\While{$q^i < K$}{
Compute local stochastic gradient $\widetilde{g}^i$ at $w^i$\;
Update local model $w^i \gets w^i - \eta\widetilde{g}^i$\;
Update local counter $q^i \gets q^i + 1$\;
}
\Wait{};
}
\textbf{end function}\
\caption{\Algo\ over $T$ iterations. In \textcolor{red}{red} are highlighted the differences with QuAFL.}
\label{algo:flau}
\end{algorithm}
We assume that the server performs a number of training epochs $T \ge 1$. At each time step $t \in \{1,\dots,T\}$, the server has a model $w_t$. At initialization, the central server transmits identical parameters $\param_0$ to all devices. At each time step $t$, the central server selects a subset $\Set_t$ of $s$ clients uniformly at random and requests  their local models. Then, the requested clients submit their \emph{reweighted} local models back to the server. When all requested models arrive at the server, the server model is updated based on a simple average (see \Cref{algo:line:server_update}). Finally, the server multicasts the updated server model to all clients in $\Set_t$. In particular, all clients $i \notin \Set_t$ continue to run their individual processes without interruption.

\begin{remark}
In \Algo's setting, we assume that each client $i \in \{1,...,n\}$ keeps a full-precision local model $w^i$. In order to reduce the computational cost induced by the training process, \Algo\ can also be implemented with a quantization function $Q$. First, each client computes backpropagation with respect to its quantized weights $Q(w^i)$. That is, the stochastic gradients are unbiased estimates of $\nabla f_i\left(Q\left(w^i \right)\right)$. Moreover, the activations computed at forward propagation are quantized. Finally, the stochastic gradient obtained at backpropagation is quantized before the SGD update. In our supplementary experiments, we use the logarithmic unbiased quantization method of \cite{chmiel2021logarithmic}.
\end{remark}

\section{Analysis}
\label{sec:analysis}
In this section we provide complexity bounds for \Algo\ in a smooth nonconvex environment. We introduce an abstraction to model the stochastic optimization process and prove convergence guarantees for \Algo.
\textbf{Preliminaries.}
\label{sec:analysis-prel}
We abstract the optimization process to simplify the analysis. In the proposed algorithm, each client asynchronously computes its own local updates without taking into account the server time step $t$. Here in the analysis, we introduce a different, but statistically equivalent setting. At the beginning of each server timestep $t$, each client maintains a local model $w_{t-1}^i$. We then assume that all $n$ clients \emph{instantaneously} compute local steps from SGD. The update in local step $q$ for a client $i$ is given by:
\begin{equation}
\label{definition-h-tilda}
\widetilde{h}_{t,q}^i = \widetilde{g}^i\left(w_{t-1}^i - \sum_{s=1}^{q-1} \eta\widetilde{h}_{t,s}^i\right),
\end{equation}
where $\widetilde{g}^i$ represents the stochastic gradient that client $i$ computes for the function $f_i$. We also define $n$ independent random variables $E_t^1,\dots,E_t^n$ in $\nset$. Each random variable $E_t^i$ models the number of local steps the client $i$ could take before receiving the server request. We then introduce the following random variable: $\widetilde{h}_t^i = \sum_{q=1}^{E_t^i}\widetilde{h}_{t,q}^i$. Compared to \cite{zakerinia2022quafl}, we do not assume that clients performed the same number of local epochs. Instead, we reweight the sum of the gradients by weights $\alpha^i$, which can be either \emph{stochastic} or \emph{deterministic}:
\begin{equation}\label{eq:reweight}
\alpha^i = \begin{cases} \mathbf{P}(E^i_{t+1} > 0) (E^i_{t+1} \wedge K) \quad \text{\emph{stochastic} version,}\\
\PE[E^i_{t+1} \wedge K] \quad \text{\emph{deterministic} version}. \end{cases}
\end{equation}
And we can define the \emph{unbiased} gradient estimator: $    \check{h}_t^i = \frac{1}{\alpha^i} \sum_{q=1}^{E_t^i \wedge K}\widetilde{h}_{t,q}^i.$

Finally, a subset $\Set_t$ of $s$ clients is chosen uniformly at random. This subset corresponds to the clients that send their models to the server at time step $t$. In the current notation, each client $i \in \Set_t$ sends the value $w_{t-1}^i - \eta\check{h}_t^i$ to the server. We emphasise that in our abstraction, all clients compute $E^i_t$ local updates. However, only the clients in $\Set_t$ send their updates to the server, and each client $i \in \Set_t$ sends only the $K$ first updates. As a result, we introduce the following update equations:
\begin{equation}
\label{definition-updates}
\begin{cases}
\param_{t} = \frac{1}{s+1} \param_{t-1} + \frac{1}{s+1} \sum_{i \in \Set_t}(\param^i_{t-1}-\eta \frac{1}{\alpha^i} \sum_{s=1}^{E^i_t \wedge K}\widetilde{h}_{t,s}^i),\\
\param_{t}^i = \param_{t}, \quad  \text{for } i \in \Set_t,\\
\param_{t}^i = \param^i_{t-1}, \quad \text{for } i \notin \Set_t.\\
\end{cases}
\end{equation}

\begin{table*}
    \centering
    \caption{How long one has to wait to reach an $\epsilon$ accuracy for non-convex functions. For simplicity, we ignore all constant terms. Each constant $\textcolor{purple}{C_{\_}}$ depends on client speeds and represents the unit of time one has to wait in between two consecutive server steps. $L$ is the Lipschitz constant, and $F:=(f(\param_0) - f_*)$ is the initial conditions term. $a_i, b$ are constants depending on client speeds statistics, and defined in \Cref{theorem:cvgquant}.}
 \resizebox{\linewidth}{!}{
 \begin{tabular}{l|c}
\toprule
Method & Units of time\\
         \midrule
        FedAvg & $\left(\frac{FL\sigma^2+(1-\frac{s}{n})KG^2}{sK}\textcolor{blue}{\epsilon^{-2}} + FL^\frac{1}{2}G\textcolor{blue}{\epsilon^{-\frac{3}{2}}} + LFB^2\textcolor{blue}{\epsilon^{-1}}\right)\textcolor{purple}{C_{FedAvg}}$\\
        FedBuff  & $\left(FL(\sigma^2+G^2)\textcolor{blue}{\epsilon^{-2}}+ FL((\frac{\tau_{max}^2}{s^2}+1)(\sigma^2+nG^2))^\frac{1}{2}\textcolor{blue}{\epsilon^{-\frac{3}{2}}}+ FL\textcolor{blue}{\epsilon^{-1}}\right)\textcolor{purple}{C_{FedBuff}}$\\
        AsyncSGD & $\left(FL(3\sigma^2+4G^2)\textcolor{blue}{\epsilon^{-2}}+ FLG(s\tau_{avg})^\frac{1}{2} \textcolor{blue}{\epsilon^{-\frac{3}{2}}}+(s\tau_{max}F)^\frac{1}{2}\textcolor{blue}{\epsilon^{-1}}\right)\textcolor{purple}{C_{AsyncSGD}}$\\
        QuAFL & $\frac{1}{E^2}FLK(\sigma^2+ 2KG^2)\textcolor{blue}{\epsilon^{-2}}+\frac{n\sqrt{n}}{E\sqrt{Es}}FKL(\sigma^2+2KG^2)^\frac{1}{2}\textcolor{blue}{\epsilon^{-\frac{3}{2}}}+\frac{1}{E\sqrt{s}}n\sqrt{n}FBK^2L\textcolor{blue}{\epsilon^{-1}}$ \\
        \Algo\ & $FL(\sigma^2\sum_i^n\frac{a_i}{n}+ 8G^2b)\textcolor{blue}{\epsilon^{-2}}+\frac{n}{s}FL^2(K^2\sigma^2+L^2K^2G^2+s^2\sigma^2\sum_i^n\frac{a_i}{n}+s^2G^2b)^\frac{1}{2}\textcolor{blue}{\epsilon^{-\frac{3}{2}}}+nFB^2KLb\textcolor{blue}{\epsilon^{-1}}$ \\
        \bottomrule
    \end{tabular}
     }
    \label{tab:cvgcomparison}
\end{table*}\textbf{Assumptions and notations.}
\begin{assumption}
\label{assum:uniflowerbound}
    Uniform Lower Bound: There exists $f_* \in \mathbb{R}$ such that $f(x) \geq f_*$ for all $x \in \mathbb{R}^d$.
\end{assumption}
\begin{assumption}
\label{assum:smoothgradients}
    Smooth Gradients: For any client $i$, the gradient $\nabla f_i(x)$ is $L$-Lipschitz continuous for some $L>0$, i.e. for all $x, y \in \mathbb{R}^d$:
$
\left\|\nabla f_i(x)-\nabla f_i(y)\right\| \leq L\|x-y\| .
$
\end{assumption}
\begin{assumption}
\label{assum:boundedvariance}
    Bounded Variance: For any client $i$, the variance of the stochastic gradients is bounded by some $\sigma^2>0$, i.e. for all $x \in \mathbb{R}^d$:
$
\mathbb{E}[\left\|\widetilde{g}^i(x)-\nabla f_i(x)\right\|^2] \leq \sigma^2 .
$
\end{assumption}
\begin{assumption}
\label{assum:graddissim}
Bounded Gradient Dissimilarity: There exist constants $G^2 \geq 0$ and $B^2 \geq 1$, such that for all $x \in \mathbb{R}^d$:
$\sum_{i=1}^n \frac{\left\|\nabla f_i(x)\right\|^2}{n} \leq G^2+B^2\|\nabla f(x)\|^2.$
\end{assumption}

We define the notations required for the analysis. Consider a time step $t$, a client $i$, and a local step $q$. We define
\begin{equation}
\label{eq:definition-mu}
\mu_{t}=\left(\param_{t}+\sum_{i=1}^{n} \param^{i}_t\right) /(n+1)
\end{equation}
the average over all node models in the system at a given time $t$. The first step of the proof is to compute a preliminary upper bound on the divergence between the local models and their average. For this purpose, we introduce the Lyapunov function: $\Phi_{t}=\left\|\param_{t}-\mu_{t}\right\|^{2}+\sum_{i=1}^{n}\left\|\param^{i}_t-\mu_{t}\right\|^{2}.$

\textbf{Upper bounding the expected change in potential.}
A key result from our analysis is to upper bound the change (in expectation) of the aforementioned potential function $\Phi_t$:
\begin{lemma}
\label{lem:boundpotential}
For any time step $t>0$ we have:
\begin{equation}
    \mathbb{E}\left[\Phi_{t+1}\right] \leq\left(1-\kappa\right) \mathbb{E}\left[\Phi_{t}\right]+3 \frac{s^2}{n} \eta^{2} \sum_{i=1}^{n} \mathbb{E}\left\|\check{h}_{t+1}^i\right\|^{2},
\end{equation} with $\kappa = \frac{1}{n} \left(\frac{s(n-s)}{2(n+1)(s+1)}\right)$.
\end{lemma}
The intuition behind \Cref{lem:boundpotential} is that the potential function $\Phi_t$ remains concentrated around its mean, apart from deviations induced by the local gradient steps. The full analysis involves many steps and we refer the reader to \Cref{sec:app:proofs} for complete proofs. In particular, \Cref{lem:boundlocalgraddiffpoints,lem:boundscalarproduct} allow us to examine the scalar product between the expected node progress $\sum_{i=1}^{n} \check{h}_{t}^i$ and the true gradient evaluated on the mean model $\nabla f(\mu_{t})$. The next theorem allows us to compute an upper-bound on the averaged norm-squared of the gradient, a standard quantity studied in non-convex stochastic optimization.

\textbf{Convergence results.}
The following statement shows that \Algo\ algorithm converges towards a first-order stationary point, as $T$ the number of global epochs grows.
\begin{theorem}
\label{theorem:cvgquant}
Assume \Cref{assum:uniflowerbound} to \Cref{assum:graddissim} and assume that the learning rate $\eta$ satisfies $\eta \le \frac{1}{20B^2bKLs}$. Then \Algo\ converges at rate:
\small
\begin{align}
\frac{1}{T}\sum_{t=0}^{T-1} \mathbb{E}  \left\|\nabla f\left(\mu_{t}\right)\right\|^{2} & \le  \frac{2(n+1)F}{Ts\eta} + \frac{Ls}{n+1}(\frac{\sigma^2}{n} \sum_i^n a^i + 8G^2b)\eta\\
&+ L^2s^2(\frac{720\sigma^2}{n} \sum_i^n a^i + 5600bG^2)\eta^2,
\end{align}
with $F:=(f(\mu_0) - f_*)$, and 
\begin{equation}
    \begin{cases}
        a^i = \frac{1}{\mathbf{P}(E^i_{t+1}>0)^2}(\frac{\mathbf{P}(E^i_{t+1}>0)}{K^2} + \PE[\frac{\mathds{1}(E^i_{t+1}>0)}{E^i_{t+1} \wedge K}]),\\
        b = \max_i(\frac{1}{\mathbf{P}(E^i_{t+1}>0)}),
\end{cases}
\end{equation}for $\alpha^i= \mathbf{P}(E^i_{t+1}>0)(E^i_{t+1} \wedge K)$, or
\begin{equation}
    \begin{cases}
        a^i = \frac{1}{\PE[E^i_{t+1} \wedge K]} + \frac{\PE[(E^i_{+1} \wedge K)^2]}{K^2 \PE[E^i_{t+1} \wedge K]},\\
        b=\max_i(\frac{\PE[(E^i_{t+1} \wedge K)^2]}{\PE[E^i_{t+1} \wedge K]}),
    \end{cases}
\end{equation}for $\alpha^i= \PE[E^i_{t+1} \wedge K]$.
\normalsize
\end{theorem}
Note that the previous convergence result refers to the average model $\mu_t$. In practice, this does not pose much of a problem. After training is complete, the server can ask each client to submit its final model. It should be noted that each client communicates $\frac{sT}{n}$ times with the server during training. Therefore, an additional round of data exchange represents only a small increase in the total amount of data transmitted.

The bound in \Cref{theorem:cvgquant} contains 3 terms. The first term is standard for a general non-convex target and expresses how initialization affects convergence. The second and third terms depend on the statistical heterogeneity of the client distributions and the fluctuation of the minibatch gradients. \Cref{tab:cvgcomparison} compares complexity bounds along with synchronous and asynchronous methods.
One can note the importance of the ratio $\frac{s}{n}$. Compared to \cite{nguyen2022federated} or \cite{koloskova2022sharper}, \Algo\ can potentially suffer from delayed updates when $\frac{s}{n} \ll 1$, but \Algo\ does \emph{not} favor fast clients at all. In practice, it is not a major shortcoming, and \Algo\ is more robust to fast/slow clients distribution than FedBuff/AsyncSGD (see \Cref{fig:mnist_noniid_slowclients}). We emphasize both FedBuff and AsyncSGD rely on strong assumptions: neither the queuing process, nor the transitional regime are taken into account in their analysis. In practice, during the first iterations, only fast clients contribute. It induces a serious bias. Our experiments indicate that a huge amount of server iterations has to be accomplished to reach the stationary regime. Still, under this regime, slow clients are contributing with delayed information. \cite{nguyen2022federated,koloskova2022sharper} propose to uniformly bound this delay by some quantity $\tau_{max}$. We keep this notation while reporting complexity bounds in \Cref{tab:cvgcomparison}, but argue nothing guarantee $\tau_{max}$ is properly defined (i.e. finite). All analyses except that of \cite{mishchenko2022asynchronous} show that the number of updates required to achieve accuracy grows linearly with $\tau_{max}$, which can be very adverse. Specifically, suppose we have two parallel workers - a fast machine that takes only $1$ unit of time to compute a stochastic gradient, and a slow machine that takes $1000$ units of time. If we use these two machines to implement FedBuff/AsyncSGD, the gradient delay of the slow machine will be one thousand, because in the $1$ unit of time we wait for the slow machine, the fast machine will produce one thousand updates. As a result, the analysis based on $\tau_{max}$ deteriorates by a factor of $1000$.

In the literature, guarantees are most often expressed as a function of server steps. In the asynchronous case, this is \emph{inappropriate} because a single step can take very different amounts of time depending on the method. For example, with FedAvg or Scaffold \cite{karimireddy2020scaffold}, one must wait for the slowest client for each individual server step. Therefore, we introduce  in \Cref{tab:cvgcomparison} constants $\textcolor{purple}{C_{\_}}$ that depend on the client speed and represent the unit of time to wait between two consecutive server steps. Finally, optimizing the value of the learning rate $\eta$ with \Cref{lem:complexity} yields the following:
\begin{corollary}
\label{coro:cvg_quant}
Assume \Cref{assum:uniflowerbound} to \Cref{assum:graddissim}. We can optimize the learning rate by \Cref{lem:complexity} and \Algo\ reaches an $\epsilon$ precision for a number of server steps $T$ greater than (up to numerical constants):
\small
\begin{align}
&\frac{FL(\frac{\sigma^2}{n}\sum_i^n a^i + 8G^2b)}{\epsilon^2}\\
+&\frac{FL^2(K^2\sigma^2+L^2K^2G^2+\frac{s^2\sigma^2}{n}\sum_i^n a^i+s^2G^2b)^\frac{1}{2}}{\epsilon^\frac{3}{2}s/n}+\frac{FB^2KLb}{\epsilon/n},
\end{align} \normalsize where $F=(f(\mu_0) - f_*)$, and $(a^i,b)$ are defined in \Cref{theorem:cvgquant}.
\end{corollary}
The second term in \Cref{coro:cvg_quant} is better than the one from the QuAFL analysis \cite{zakerinia2022quafl}. Although this $(n+1)$ term can be suboptimal, note that it is only present at second order from $\epsilon$ and therefore becomes negligible when $\epsilon$ goes to $0$ \cite{lu2020moniqua, zakerinia2022quafl}.

\begin{remark}
Our analysis can be extended to the case of quantized neural networks. The derived complexity bounds also hold for the case when the quantization function $Q$ is biased. We make only a weak assumption about $Q$ (we assume that there is a constant $r_d$ such that for any $x \in \mathbb{R}^d$ $\|Q(x) - x\|^2 \le r_d$), which holds for standard quantization methods such as stochastic rounding and deterministic rounding. The only effect of quantization would be increased variance in the stochastic gradients. We need to add to the upper bound given in \Cref{theorem:cvgquant} an "error floor" of $12L^2r_d$, which remains independent of the number of server epochs. For stochastic or deterministic rounding, $r_d = \Theta(d\frac{1}{2^{2b}})$, where $b$ is the number of bits used. The error bound is the cost of using quantization as part of the optimization algorithm. Previous works with quantized models also include error bounds \cite{li2017training,li2019dimensionfree}.
\end{remark}

\section{Numerical Results}
\label{sec:experiments}
We test  \Algo\ on three image classification tasks: MNIST \cite{deng2012mnist}, CIFAR-10 \cite{Krizhevsky2009}, and TinyImageNet \cite{le2015tiny}. For the MNIST and CIFAR-10 datasets, two training sets are considered: an IID and a non-IIID split. In the first case, the training images are randomly distributed among the $n$ clients. In the second case, each client takes two classes (out of the ten possible) without replacement. This process leads to heterogeneity among the clients.

The standard evaluation measure for FL is the number of server rounds of communication to achieve target accuracy. However, the time spent between two consecutive server steps can be very different for asynchronous and synchronous methods. Therefore, we compare different synchronous and asynchronous methods w.r.t. \emph{total simulation time} (see below). We also measured the loss and accuracy of the model in terms of server steps and total local client steps (see \Cref{sec:app:detailedresults}).
In all experiments, we track the performance of each algorithm by evaluating the server model against an unseen validation dataset. We present the test accuracy and variance, defined as $\sum_{i=1}^n \Vert \param^i_t-\param_t \Vert^2$.

We decide to focus on non-uniform timing experiments as in \cite{nguyen2022federated}, and we base our simulation environment on QuAFL's code\footnote{https://github.com/ShayanTalaei/QuAFL}. After simulating $n$ clients, we randomly group them into fast or slow nodes. We assume that at each time step $t$ (for the central server), a set of $s$ clients is randomly selected without replacement. We assume that the clients have different computational speeds, and refer to \Cref{sec:app:simuruntime} for more details. We assume that only one-third of the clients are slow, unless otherwise noted. We compare \Algo\ with the classic synchronous approach FedAvg \cite{mcmahan2017communication} and two newer asynchronous metods QuAFL \cite{zakerinia2022quafl} and FedBuff \cite{nguyen2022federated}. Details on implementing other methods can be found in \Cref{sec:app:concu_works}.

We use the standard data augmentations and normalizations for all methods. \Algo\ is implemented in Pytorch, and experiments are performed on an NVIDIA Tesla-P100 GPU. Standard multiclass cross entropy loss is used for all experiments. All models are fine-tuned with $n=100$ clients, $K=20$ local epochs, and a batch of size $128$. Following the guidelines of \cite{nguyen2022federated}, the buffer size in FedBuff is set to $Z=10$. In FedAvg, the total simulated time depends on the maximum number of local steps $K$ and the slowest client runtime, so it is proportional to the number of local steps and the number of global steps. In QuAFL and \Algo\, on the other hand, each global step has a predefined duration that depends on the central server clock. Therefore, the global steps have similar durations and the total simulated time is the sum of the durations of the global steps. In FedBuff, a global step requires filling a buffer of size $Z$. Consequently, both the duration of a global step and the total simulated time depend on $Z$ and on the proportion of slow clients (see \Cref{sec:app:simuruntime} for a detailed discussion).

We first report the accuracy of a shallow neural network trained on MNIST. The learning rate is set to $0.5$ and the total simulated time is set to $5000$. We also compare the accuracy of a Resnet20 \cite{he2016deep} with the CIFAR-10 dataset \cite{Krizhevsky2009}, which consists of 50000 training images and 10000 test images (in 10 classes). For CIFAR-10, the learning rate is set to $0.005$ and the total simulation time is set to $10000$. 
\begin{figure}[h]
    \centering
    \includegraphics[scale=0.4]{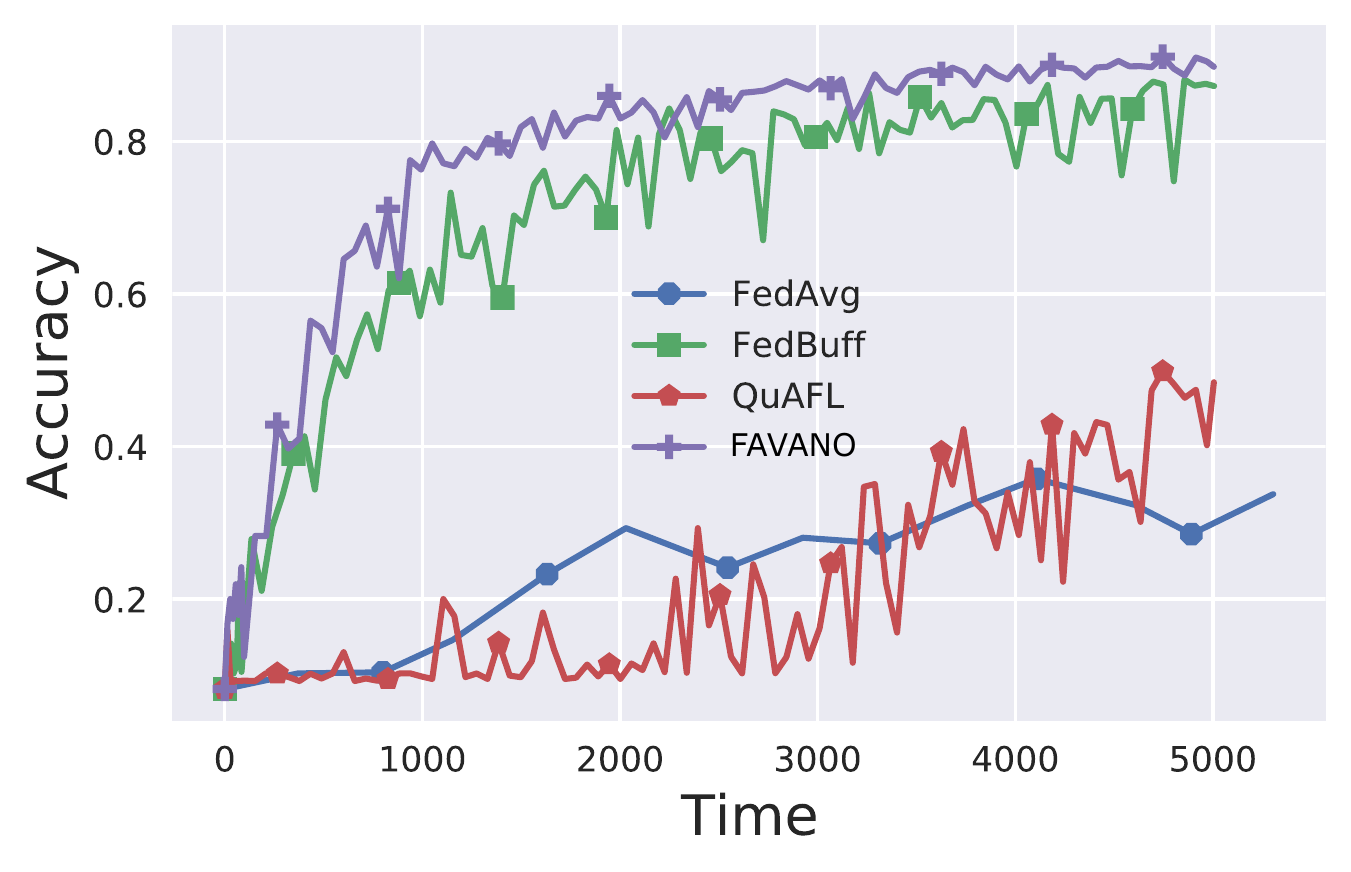}
    \caption{Test accuracy on the MNIST dataset with a non-IID split in between $n=100$ total nodes, $s=20$.}
    \label{fig:mnist_noniid}
\end{figure}
\begin{table}[]
    \caption{Final accuracy on the test set (average and standard deviation over 10 random experiments) for the MNIST classification task.}
    \label{tab:mnist_seeds}
    \begin{tabular}{lccc}
    \toprule
    Methods & IID split & \begin{tabular}{@{}c@{}}non-IID split \\ ($\frac{2}{3}$ fast clients)\end{tabular} & \begin{tabular}{@{}c@{}}non-IID split \\ ($\frac{1}{9}$ fast clients)\end{tabular} \\
         \midrule
         FedAvg & $93.4 \pm 0.3$ & $38.7 \pm 7.7$ & $44.8 \pm 6.9$\\
         QuAFL & $92.3 \pm 0.9$ & $40.7 \pm 6.7$ & $45.5 \pm 4.0$ \\
         FedBuff & $\textbf{96.0} \pm 0.1$ & $85.1 \pm 3.2$ & $67.3 \pm 5.5$ \\
         \Algo\ & $95.1 \pm 0.1$ & $\textbf{88.9} \pm 0.9$ & $\textbf{87.3} \pm 2.3$ \\
        \bottomrule
    \end{tabular}
\label{tab:mnistxp_seeds}
\end{table}
In \Cref{fig:mnist_noniid}, we show the test accuracy of \Algo\ and competing methods on the MNIST dataset. We find that \Algo\ and other asynchronous methods can offer a significant  advantage over FedAvg when time is taken into account. However, QuAFL does not appear to be adapted to the non-IID environment. We identified client-side updating as a major shortcoming. While this is not severe when each client optimizes (almost) the same function, the QuAFL mechanism suffers from significant client drift when there is greater heterogeneity between clients.
\begin{figure}[h]
    \centering
    \includegraphics[scale=0.475]{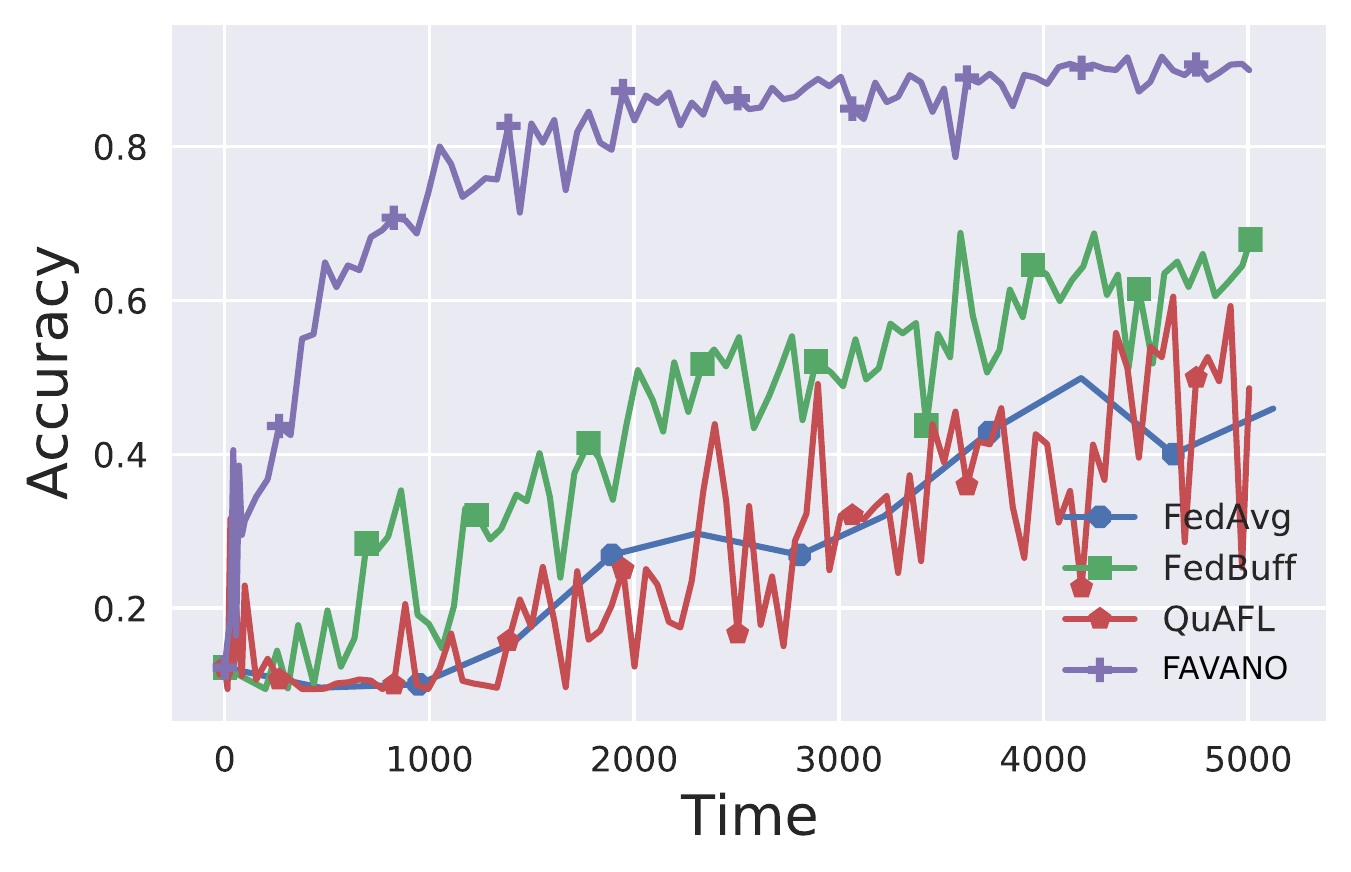}
    \includegraphics[scale=0.475]{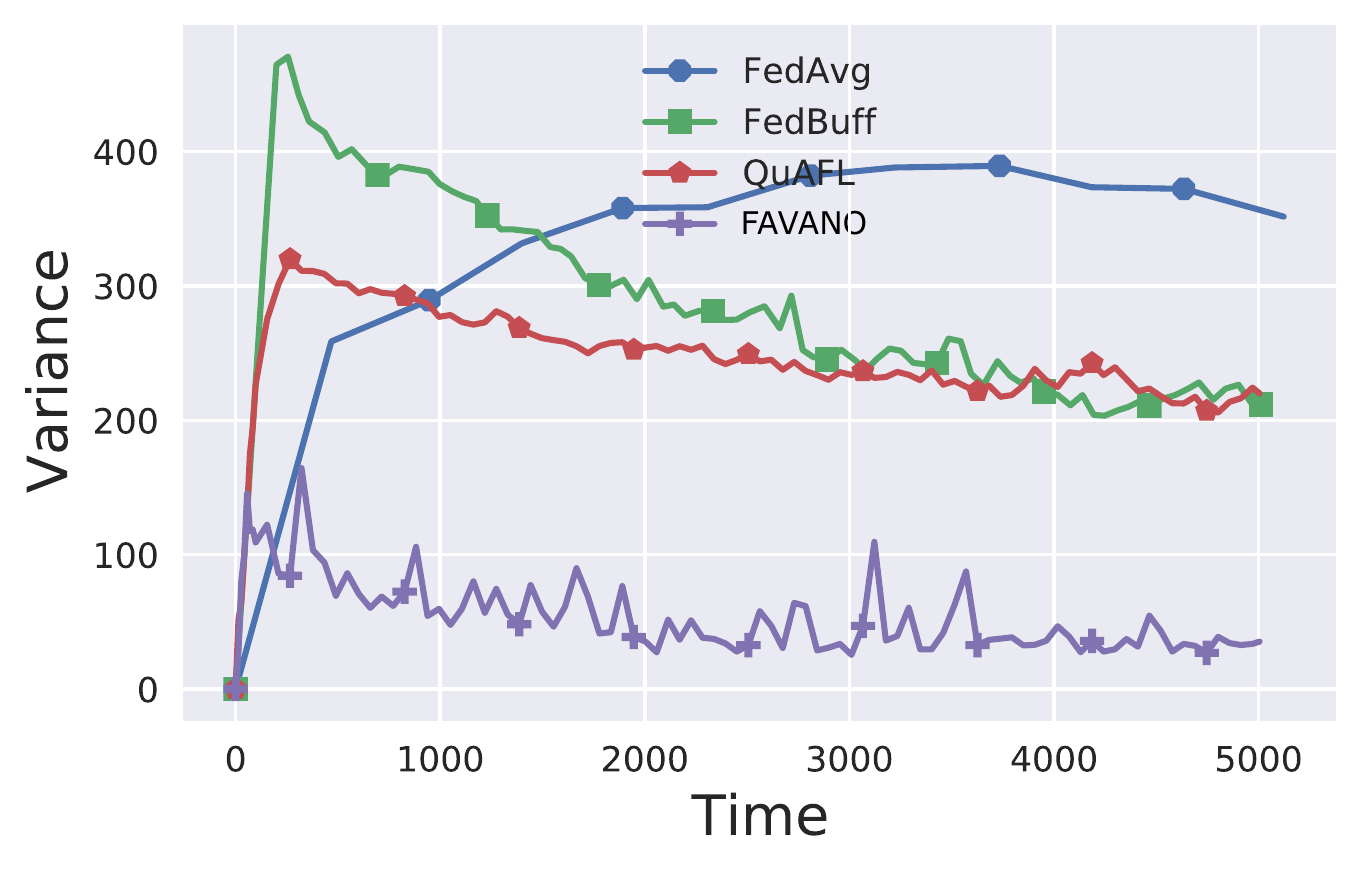}
    \caption{Test accuracy and variance on the MNIST dataset with a non-IID split between $n=100$ total nodes. In this particular experiment, one-ninth of the clients are defined as fast.}
    \label{fig:mnist_noniid_slowclients}
\end{figure}
FedBuff is efficient when the number of stragglers is negligible compared to $n$. However, FedBuff is sensitive to the fraction of slow clients and may get stuck if the majority of clients are classified as slow and a few are classified as fast. In fact, fast clients will mainly feed the buffer, so the central updates will be heavily biased towards fast clients, and little information from slow clients will be considered. \Cref{fig:mnist_noniid_slowclients} illustrates this phenomenon, where one-ninth of the clients are classified as fast. To provide a fair comparison, \Cref{tab:mnist_seeds} gives the average performance of 10 random experiments with the different methods on the test set.

\begin{figure}[h]
    \centering
    \begin{subfigure}[b]{0.47\textwidth}
         \centering         
         \includegraphics[scale=0.47]{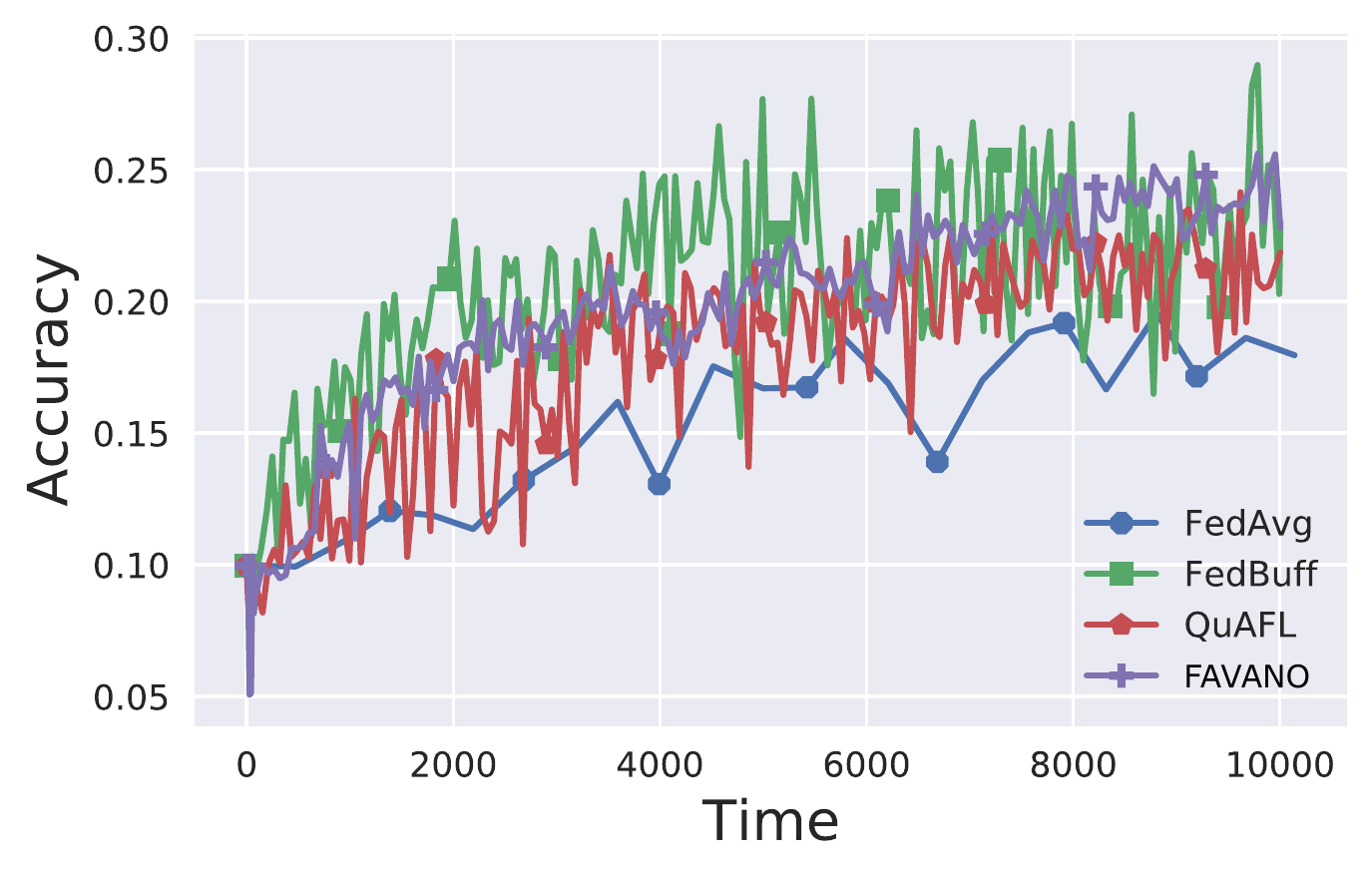}
        \caption{CIFAR-10 (non-IID)}
         \label{fig:cifar10_noniid}
     \end{subfigure}
     \hfill
     \begin{subfigure}[b]{0.47\textwidth}
         \centering
        \includegraphics[scale=0.47]{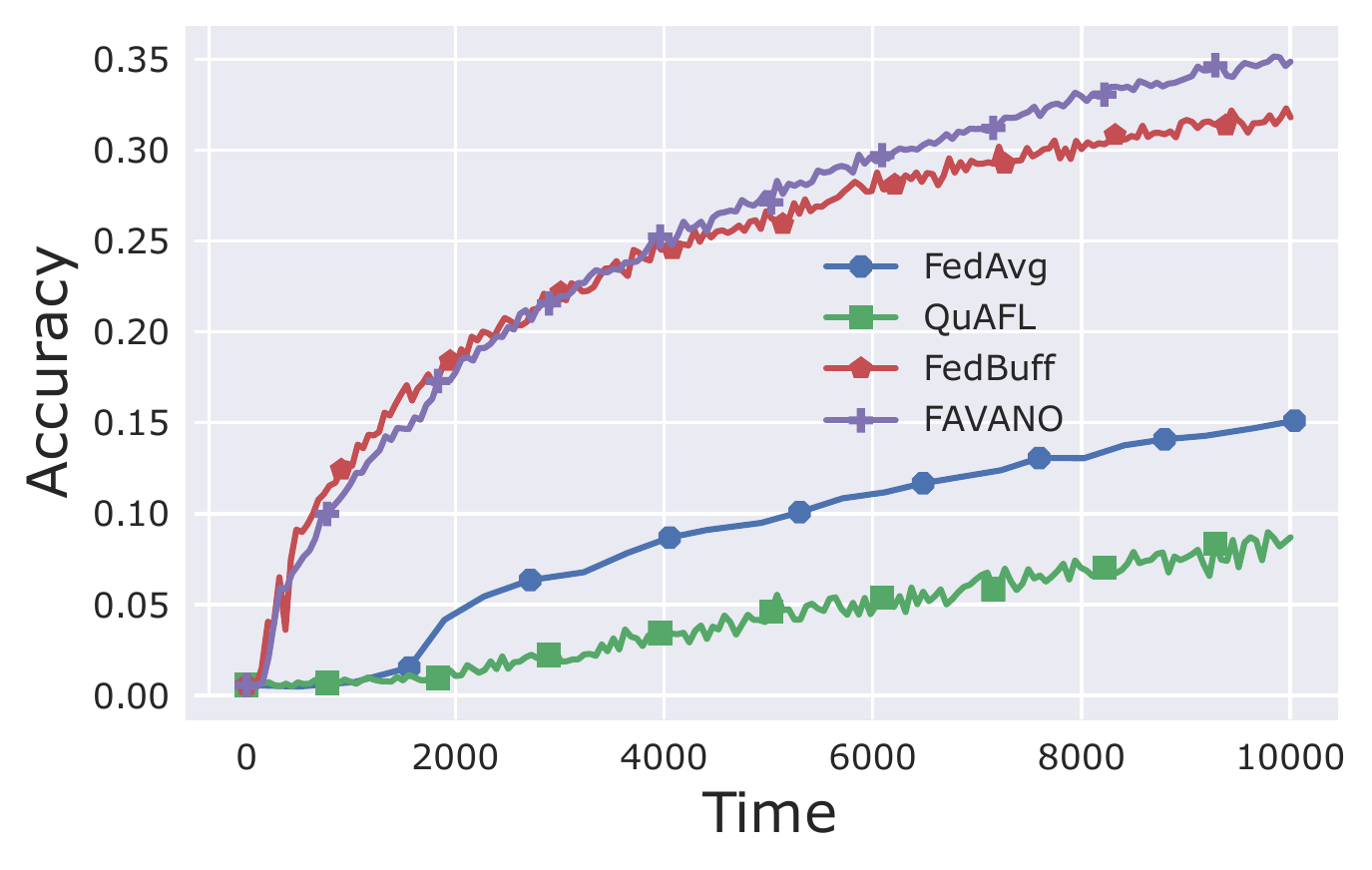}
         \caption{TinyImageNet (IID)}
         \label{fig:tinyimagenet}
     \end{subfigure}
        \caption{Test accuracy on CIFAR-10 and TinyImageNet datasets with $n=100$ total nodes. Central server selects $s=20$ clients at each round.}
    \label{fig:cifar10andtinyimagenet}
\end{figure}

In \Cref{fig:cifar10_noniid}, we report accuracy on a non-IID split of the CIFAR-10 dataset. FedBuff and \Algo\ both perform better than other approaches, but FedBuff suffers from greater variance. We explain this limitation by the bias FedBuff provides in favor of fast clients.
We also tested \Algo\ on the TinyImageNet dataset \cite{le2015tiny} with a ResNet18. TinyImageNet has 200 classes and each class has 500 (RGB) training images, 50 validation images and 50 test images. To train ResNet18, we follow the usual practices for training NNs: we resize the input images to $64 \times 64$ and then randomly flip them horizontally during training. During testing, we center-crop them to the appropriate size. The learning rate is set to $0.1$ and the total simulated time is set to $10000$. \Cref{fig:tinyimagenet} illustrates the performance of \Algo\ in this experimental setup. While the partitioning of the training dataset follows an IID strategy, TinyImageNet provides enough diversity to challenge federated learning algorithms. \Cref{fig:tinyimagenet} shows that \Algo\ scales much better on large image classification tasks than any of the methods we considered.

\begin{remark}
We also evaluated the performance of \Algo\ with and without quantization. We ran the code \footnote{https://openreview.net/forum?id=clwYez4n8e8} from LUQ \cite{chmiel2021logarithmic} and adapted it to our datasets and the FL framework. Even when the weights and activation functions are highly quantized, the results are close to their full precision counterpart (see \Cref{fig:quantiz} in \Cref{sec:app:exp}).
\end{remark}

\section{Conclusion}
We have presented \Algo\, the first (centralised) Federated Learning
method of federated averaging that accounts for asynchrony in resource-constrained environments. We established complexity bounds under verifiable assumptions with explicit dependence on all relevant constants. Empirical evaluation shows that \Algo\ is more efficient than synchronous and asynchronous state-of-the-art mechanisms in standard CNN training benchmarks for image classification.

\clearpage
\pagebreak


\bibliographystyle{IEEEbib}
\bibliography{biblio}

\begin{thebibliography}{10}

\bibitem{konevcny2015federated}
Jakub Kone{\v{c}}n{\`y}, Brendan McMahan, and Daniel Ramage,
\newblock ``Federated optimization: Distributed optimization beyond the datacenter,''
\newblock {\em arXiv preprint arXiv:1511.03575}, 2015.

\bibitem{mcmahan2017communication}
Brendan McMahan, Eider Moore, Daniel Ramage, Seth Hampson, and Blaise~Aguera y~Arcas,
\newblock ``Communication-efficient learning of deep networks from decentralized data,''
\newblock in {\em Artificial intelligence and statistics}. PMLR, 2017, pp. 1273--1282.

\bibitem{kairouz2021advances}
Peter Kairouz, H~Brendan McMahan, Brendan Avent, Aur{\'e}lien Bellet, Mehdi Bennis, Arjun~Nitin Bhagoji, Kallista Bonawitz, Zachary Charles, Graham Cormode, Rachel Cummings, et~al.,
\newblock ``Advances and open problems in federated learning,''
\newblock {\em Foundations and Trends{\textregistered} in Machine Learning}, vol. 14, no. 1--2, pp. 1--210, 2021.

\bibitem{zhao2018federated}
Yue Zhao, Meng Li, Liangzhen Lai, Naveen Suda, Damon Civin, and Vikas Chandra,
\newblock ``Federated learning with non-iid data,''
\newblock {\em arXiv preprint arXiv:1806.00582}, 2018.

\bibitem{horvath2022fedshuffle}
Samuel Horv{\'a}th, Maziar Sanjabi, Lin Xiao, Peter Richt{\'a}rik, and Michael Rabbat,
\newblock ``Fedshuffle: Recipes for better use of local work in federated learning,''
\newblock {\em arXiv preprint arXiv:2204.13169}, 2022.

\bibitem{wang2021field}
Jianyu Wang, Zachary Charles, Zheng Xu, Gauri Joshi, H~Brendan McMahan, Maruan Al-Shedivat, Galen Andrew, Salman Avestimehr, Katharine Daly, Deepesh Data, et~al.,
\newblock ``A field guide to federated optimization,''
\newblock {\em arXiv preprint arXiv:2107.06917}, 2021.

\bibitem{xie2019asynchronous}
Cong Xie, Sanmi Koyejo, and Indranil Gupta,
\newblock ``Asynchronous federated optimization,''
\newblock {\em arXiv preprint arXiv:1903.03934}, 2019.

\bibitem{chen2020asynchronous}
Yujing Chen, Yue Ning, Martin Slawski, and Huzefa Rangwala,
\newblock ``Asynchronous online federated learning for edge devices with non-iid data,''
\newblock in {\em 2020 IEEE International Conference on Big Data (Big Data)}. IEEE, 2020, pp. 15--24.

\bibitem{chen2021towards}
Zheyi Chen, Weixian Liao, Kun Hua, Chao Lu, and Wei Yu,
\newblock ``Towards asynchronous federated learning for heterogeneous edge-powered internet of things,''
\newblock {\em Digital Communications and Networks}, vol. 7, no. 3, pp. 317--326, 2021.

\bibitem{xu2021asynchronous}
Chenhao Xu, Youyang Qu, Yong Xiang, and Longxiang Gao,
\newblock ``Asynchronous federated learning on heterogeneous devices: A survey,''
\newblock {\em arXiv preprint arXiv:2109.04269}, 2021.

\bibitem{zakerinia2022quafl}
Hossein Zakerinia, Shayan Talaei, Giorgi Nadiradze, and Dan Alistarh,
\newblock ``Quafl: Federated averaging can be both asynchronous and communication-efficient,''
\newblock {\em arXiv preprint arXiv:2206.10032}, 2022.

\bibitem{le2015tiny}
Ya~Le and Xuan Yang,
\newblock ``Tiny imagenet visual recognition challenge,''
\newblock {\em CS 231N}, vol. 7, no. 7, pp. 3, 2015.

\bibitem{lin2019don}
Tao Lin, Sebastian~U Stich, Kumar~Kshitij Patel, and Martin Jaggi,
\newblock ``Don't use large mini-batches, use local sgd,''
\newblock in {\em International Conference on Learning Representations}, 2019.

\bibitem{karimireddy2020scaffold}
Sai~Praneeth Karimireddy, Satyen Kale, Mehryar Mohri, Sashank Reddi, Sebastian Stich, and Ananda~Theertha Suresh,
\newblock ``Scaffold: Stochastic controlled averaging for federated learning,''
\newblock in {\em International Conference on Machine Learning}. PMLR, 2020, pp. 5132--5143.

\bibitem{wang2020tackling}
Jianyu Wang, Qinghua Liu, Hao Liang, Gauri Joshi, and H~Vincent Poor,
\newblock ``Tackling the objective inconsistency problem in heterogeneous federated optimization,''
\newblock {\em Advances in neural information processing systems}, vol. 33, pp. 7611--7623, 2020.

\bibitem{qu2021feddq}
Linping Qu, Shenghui Song, and Chi-Ying Tsui,
\newblock ``Feddq: Communication-efficient federated learning with descending quantization,''
\newblock {\em arXiv preprint arXiv:2110.02291}, 2021.

\bibitem{makarenko2022adaptive}
Maksim Makarenko, Elnur Gasanov, Rustem Islamov, Abdurakhmon Sadiev, and Peter Richtarik,
\newblock ``Adaptive compression for communication-efficient distributed training,''
\newblock {\em arXiv preprint arXiv:2211.00188}, 2022.

\bibitem{mao2022communication}
Yuzhu Mao, Zihao Zhao, Guangfeng Yan, Yang Liu, Tian Lan, Linqi Song, and Wenbo Ding,
\newblock ``Communication-efficient federated learning with adaptive quantization,''
\newblock {\em ACM Transactions on Intelligent Systems and Technology (TIST)}, vol. 13, no. 4, pp. 1--26, 2022.

\bibitem{tyurin2022dasha}
Alexander Tyurin and Peter Richt{\'a}rik,
\newblock ``Dasha: Distributed nonconvex optimization with communication compression, optimal oracle complexity, and no client synchronization,''
\newblock {\em arXiv preprint arXiv:2202.01268}, 2022.

\bibitem{smith2017federated}
Virginia Smith, Chao-Kai Chiang, Maziar Sanjabi, and Ameet~S Talwalkar,
\newblock ``Federated multi-task learning,''
\newblock {\em Advances in neural information processing systems}, vol. 30, 2017.

\bibitem{bonawitz2019towards}
Keith Bonawitz, Hubert Eichner, Wolfgang Grieskamp, Dzmitry Huba, Alex Ingerman, Vladimir Ivanov, Chloe Kiddon, Jakub Kone{\v{c}}n{\`y}, Stefano Mazzocchi, Brendan McMahan, et~al.,
\newblock ``Towards federated learning at scale: System design,''
\newblock {\em Proceedings of Machine Learning and Systems}, vol. 1, pp. 374--388, 2019.

\bibitem{wang2022asyncfeded}
Qiyuan Wang, Qianqian Yang, Shibo He, Zhiguo Shui, and Jiming Chen,
\newblock ``Asyncfeded: Asynchronous federated learning with euclidean distance based adaptive weight aggregation,''
\newblock {\em arXiv preprint arXiv:2205.13797}, 2022.

\bibitem{nguyen2022federated}
John Nguyen, Kshitiz Malik, Hongyuan Zhan, Ashkan Yousefpour, Mike Rabbat, Mani Malek, and Dzmitry Huba,
\newblock ``Federated learning with buffered asynchronous aggregation,''
\newblock in {\em International Conference on Artificial Intelligence and Statistics}. PMLR, 2022, pp. 3581--3607.

\bibitem{koloskova2022sharper}
Anastasia Koloskova, Sebastian~U Stich, and Martin Jaggi,
\newblock ``Sharper convergence guarantees for asynchronous sgd for distributed and federated learning,''
\newblock {\em arXiv preprint arXiv:2206.08307}, 2022.

\bibitem{fraboni2023general}
Yann Fraboni, Richard Vidal, Laetitia Kameni, and Marco Lorenzi,
\newblock ``A general theory for federated optimization with asynchronous and heterogeneous clients updates,''
\newblock {\em Journal of Machine Learning Research}, vol. 24, no. 110, pp. 1--43, 2023.

\bibitem{liu2021adaptive}
Jianchun Liu, Hongli Xu, Lun Wang, Yang Xu, Chen Qian, Jinyang Huang, and He~Huang,
\newblock ``Adaptive asynchronous federated learning in resource-constrained edge computing,''
\newblock {\em IEEE Transactions on Mobile Computing}, 2021.

\bibitem{toghani2022unbounded}
Mohammad~Taha Toghani and C{\'e}sar~A Uribe,
\newblock ``Unbounded gradients in federated leaning with buffered asynchronous aggregation,''
\newblock {\em arXiv preprint arXiv:2210.01161}, 2022.

\bibitem{mishchenko2022asynchronous}
Konstantin Mishchenko, Francis Bach, Mathieu Even, and Blake Woodworth,
\newblock ``Asynchronous sgd beats minibatch sgd under arbitrary delays,''
\newblock {\em arXiv preprint arXiv:2206.07638}, 2022.

\bibitem{chmiel2021logarithmic}
Brian Chmiel, Ron Banner, Elad Hoffer, Hilla~Ben Yaacov, and Daniel Soudry,
\newblock ``Logarithmic unbiased quantization: Simple 4-bit training in deep learning,''
\newblock {\em arXiv preprint arXiv:2112.10769}, 2021.

\bibitem{lu2020moniqua}
Yucheng Lu and Christopher De~Sa,
\newblock ``Moniqua: Modulo quantized communication in decentralized sgd,''
\newblock in {\em International Conference on Machine Learning}. PMLR, 2020, pp. 6415--6425.

\bibitem{li2017training}
Hao Li, Soham De, Zheng Xu, Christoph Studer, Hanan Samet, and Tom Goldstein,
\newblock ``Training quantized nets: A deeper understanding,''
\newblock {\em Advances in Neural Information Processing Systems}, vol. 30, 2017.

\bibitem{li2019dimensionfree}
Zheng Li and Christopher~De Sa,
\newblock ``Dimension-free bounds for low-precision training,'' 2019.

\bibitem{deng2012mnist}
Li~Deng,
\newblock ``The mnist database of handwritten digit images for machine learning research [best of the web],''
\newblock {\em IEEE signal processing magazine}, vol. 29, no. 6, pp. 141--142, 2012.

\bibitem{Krizhevsky2009}
Alex Krizhevsky, Geoffrey Hinton, et~al.,
\newblock ``Learning multiple layers of features from tiny images,''
\newblock 2009.

\bibitem{he2016deep}
Kaiming He, Xiangyu Zhang, Shaoqing Ren, and Jian Sun,
\newblock ``Deep residual learning for image recognition,''
\newblock in {\em Proceedings of the IEEE conference on computer vision and pattern recognition}, 2016, pp. 770--778.

\bibitem{lannelongue2021green}
Lo{\"\i}c Lannelongue, Jason Grealey, and Michael Inouye,
\newblock ``Green algorithms: Quantifying the carbon footprint of computation,''
\newblock {\em Advanced Science}, vol. 8, no. 12, pp. 2100707, 2021.

\end{thebibliography}
\newpage
\appendix
\onecolumn
\section{Environmental footprint}
In the current context, we estimated the carbon footprint of our experiments to be about 10 kg CO2e (calculated using green-algorithms.org v2.1 \cite{lannelongue2021green}). This shed light on the crucial need to develop energy friendly NNs.
\section{Proofs}
\label{sec:app:proofs}
The complete analysis of \Cref{theorem:cvgquant} is fully provided in the following pages, and is heavily inspired by \cite{zakerinia2022quafl}'s analysis. We ask the reader to refer to \Cref{sec:analysis-prel} for a detailed definition of the random variables used in the analysis.

\subsection{Preliminaries}
\label{sec:app_prel}

Let $(\Omega, \mathcal{F}, \mathbb{P})$ be a probability space. We assume that all the random variables defined in this proof are defined on this space. Consider $I = \mathbb{N}^* \times \mathbb{N}$, together with the lexicographical ordering in $I$. We recall that $(a,b) \le (c,d)$ in lexicographical ordering if and only if $a < c$ or $a = c$ and $b \le d$. We define two families of $\sigma$-algebras $\left(\mathcal{F}_{(t,q)}\right)_{(t,q) \in I}$ and $\left(\mathcal{F}_{t}\right)_{t \in \mathbb{N}}$. These are defined by the relations $\mathcal{F}_0 = \{\emptyset, \Omega\}$ and, for $t \ge 1$ and $q \ge 0$,
\begin{align}
\label{eq:definition-F-t-q}
\mathcal{F}_{(t,q)} = \sigma\left(\mathcal{F}_{t-1} \cup \sigma\left(\widetilde{h}_{t, q'}^i : q' \le q, i\in[1,n]\right) \right) \\
\label{eq:definition-F-t}
\mathcal{F}_t = \sigma\left(\sigma\left(\Set_{t}, E_{t}^i : i \in [1,n]\right) \cup \bigcup_{q \ge 0} \mathcal{F}_{(t,q)}\right).
\end{align}
Therefore, $\mathcal{F}_t$ contains all information up to the end of time step $t$. Additionally, $\mathcal{F}_{(t,q)}$ contains all information up to the local step $q$ of time step $t$. Notice that at this point we do not have information about $\Set_t$ and $E_{t}^i$. We define $\PE_t$ to be the conditional expectation with respect to $\mathcal{F}_t$.

For all time steps $t$, local steps $q$, and client $i$, we define:
\begin{equation}
h_{t,q}^i = \nabla f_i \left(w_t^i - \eta\sum_{s=1}^{q-1} \widetilde{h}_{t,s}^i\right) = \PE \left[\widetilde{h}_{t,q}^i | \mathcal{F}_{(t,q-1)}\right]
\end{equation} where $\widetilde{h}^i_{t,s}$ is defined in \eqref{definition-h-tilda}.

Contrary to \cite{zakerinia2022quafl}, we do \emph{not} assume that all clients have computed the same number of epochs upon being contacted by the server.

We start by establishing a basic, yet useful, algebraic equality in the following Lemma.
\begin{lemma}
\label{lem:vectorequality2}
Let $s \in \mathbb{N}^*$,  $a_i, b \in \mathbb{R}^d$  for $i \in \left[|1, s|\right]$ be vectors. It holds that:
\begin{equation}
    \left\|\frac{\sum_{i =1}^s a_i+b}{s+1}\right\|^{2}-\frac{1}{s+1}\sum_{i=1}^s\left\|a_i\right\|^{2}-\frac{1}{s+1}\left\|b\right\|^{2} = \frac{-1}{(s+1)^2} \sum_{i =1}^s \Vert a_i-b\Vert^2 - \frac{1}{(s+1)^2} \sum_{i,j =1}^s \Vert a_i-a_j\Vert^2
\end{equation}
\begin{proof}
\begin{align}
I&=   \left\|\frac{\sum_{i =1}^s a_i+b}{s+1}\right\|^{2}-\frac{1}{s+1}\sum_{i=1}^s \left\|a_i\right\|^{2}-\frac{1}{s+1}\left\|b\right\|^{2} \\
&= (s+1)^{-2}\left( \left\langle \sum_{i =1}^s a_i + b, \sum_{i =1}^s a_i + b \right\rangle - (s+1)\sum_{i=1}^s \left\|a_i\right\|^{2} - (s+1)\left\|b\right\|^{2} \right).
\end{align}
We expand the inner product to obtain:
\begin{align}
&I = (s+1)^{-2}\left(-s\left\|b\right\|^{2} +2\sum_{i =1}^s \langle a_i,b\rangle +\sum_{i,j =1}^s \langle a_i, a_j \rangle - (s+1)\sum_{i=1}^s \left\|a_i\right\|^{2} \right)\\
&    = \frac{-1}{(s+1)^2} \left(\left(\sum_{i=1}^s \left\|a_i\right\|^{2} + s\left\|b\right\|^{2}-2\sum_{i =1}^s \langle a_i,b\rangle \right)+ \left(s\sum_{i=1}^s \left\|a_i\right\|^{2}-\sum_{i,j =1}^s \langle a_i, a_j \rangle \right)\right)\\
&    = \frac{-1}{(s+1)^2} \left(\left(\sum_{i=1}^s \Vert a_i-b\Vert^2 \right)+ \left(\sum_{i,j =1}^s \Vert a_i-a_j\Vert^2 \right)\right).
\end{align}
\end{proof}
\end{lemma}

\begin{lemma}
\label{lem:centerofmass}
Let $n,d$ be positive integers, $a_1,a_2,...,a_n,b \in \mathbb{R}^d$ be vectors, and $g = \frac{a_1+...+a_n}{n}$ be the center of mass of $a_1,...,a_n$. Then the following identity holds:
\begin{equation}
\sum_{i=1}^n \|b-a_i\|^2 = n\|b-g\|^2 + \sum_{i=1}^n \|g-a_i\|^2.
\end{equation}
\begin{proof}
We may compute:
\begin{align}
\sum_{i=1}^n \|b-a_i\|^2 = \sum_{i=1}^n \|b - g + g - a_i\|^2 = \sum_{i=1}^n \left(\|b - g\|^2 + \|g - a_i\|^2 + 2\left\langle b-g, g-a_i\right\rangle\right)
\end{align}
We can easily see that $\sum_{i=1}^n\left\langle b-g, g-a_i\right\rangle = 0$. The identity then follows.
\end{proof}
\end{lemma}

\begin{lemma}
\label{lem:expectation-sum}
Let $X_1,\dots,X_n$ be random variables. Moreover, let $\Set$ be a subset of $\{1,2,...,n\}$ containing $s$ elements chosen uniformly at random. Assume that $\Set$ is independent from $X_i$ for $i=1,...,n$. Then, we have:
\begin{align}
\PE\left[\sum\nolimits_{i \in \Set} X_i \right]= \sum_{i=1}^n \frac{s}{n} \PE [X_i].
\end{align}
\begin{proof}
We introduce indicator functions in the sum above and apply linearity of expectation:
\begin{align}
\PE\left[\sum\nolimits_{i \in \Set} X_i \right] = \PE \left[\sum\nolimits_{i=1}^n X_i \mathds{1}_{\Set}(i)\right] = \sum_{i=1}^n \PE [X_i \mathds{1}_{\Set}(i)].
\end{align}
Using that $\Set$ is independent of each $X_i$ we get:
\begin{align}
\PE\left[\sum\nolimits_{i \in \Set} X_i \right] = \sum_{i=1}^n \PE [X_i] \PE [\mathds{1}_{\Set}(i)] = \sum_{i=1}^n \frac{s}{n} \PE [X_i].
\end{align}
\end{proof}
\end{lemma}
The following preliminary lemmas allow one to recover unbiased gradient estimates in \Algo, and bound their variance.
\begin{lemma}
Let  $\{Y^q\}_{q>0}$  a collection of independent random variables such that  $\PE[Y_q]=\mu$. Let $S$ be a positive random variable independent from the collection $\{Y^q\}_{q>0}$, and with expected value $\PE[S]=m$. Consider $M_1=\PE[\frac{\mathds{1}[S>0]}{S\mathbf{P}(S>0)} \sum_{q=1}^S Y^q]$, and $M_2=\PE[\frac{\mathds{1}[S>0]}{\PE[S]} \sum_{q=1}^S Y^q]$. $M_1, M_2$ are unbiased estimate of $\mu$, i.e. $M=\mu$.
\label{lem:bias}
\begin{proof}
\textbf{$M_1:$} One can note this setting corresponds to $\alpha^i = \mathbf{P}(E^i_t>0)(E^i_t \wedge K)$. We have 
\begin{align}
M_1 =& \frac{1}{\mathbf{P}(S>0)}\PE[\PE[\frac{\mathds{1}[S>0]}{S} \sum_{q=1}^S Y^q|S]]\\
 =& \frac{1}{\mathbf{P}(S>0)} \PE[S \frac{\mathds{1}[S>0]}{S} \mu]\\
=& \frac{1}{\mathbf{P}(S>0)}\PE[\mathbf{1}[S>0] \mu]\\
=& \mu.
\end{align}
Thus when reweighting with the (random) number of additions, one need to also take into account the $\mathbf{P}(s>0)$ term to obtain an unbiased estimate.

\textbf{$M_2:$} This setting corresponds to $\alpha^i = \PE[E^i_t \wedge K]$. We have 
\begin{align}
M_2 =& \PE[\PE[\frac{\mathds{1}[S>0]}{\PE[S]} \sum_{q=1}^S Y^q|S]]\\
 =& \PE[S \frac{\mathds{1}[S>0]}{\PE[S]} \mu]\\
=& \frac{\PE[S\mathds{1}[S>0]]}{\PE[S]} \mu\\
=& \mu.
\end{align}
Thus reweighting the sum with $\PE[s]$ allows us to obtain an unbiased estimate $M_2=\mu$.
\end{proof}
\end{lemma}
\begin{lemma}
Let $\{Y^q\}_{q>0}$  a collection of independent random variables such that $\PE[Y_q]=\mu$, $\var(Y^q)=\var(Y)<\infty$. Let $S$ be a positive random variable independent from the collection $\{Y^q\}_{q>0}$, and with expected value $\PE[S]=m$.  Consider $M_1=\PE[\frac{\mathds{1}[S>0]}{S\mathbf{P}(S>0)} \sum_{q=1}^S Y^q]$, and $M_2=\PE[\frac{\mathds{1}[S>0]}{\PE[S]} \sum_{q=1}^S Y^q]$. We compute the variance :
\begin{equation} \begin{cases}
\var(M_1) = \frac{1}{\mathbf{P}(s>0)^2}(\mu^2 \mathbf{P}(S>0)(1-\mathbf{P}(S>0)) + \var(Y)\PE[\frac{\mathds{1}[S>0]}{S}])\\
\var(M_2) = \frac{\mu^2\var(S)}{\PE[S]^2} + \frac{\var(Y)}{\PE[S]} .
\end{cases}
\end{equation}
\label{lem:variance}
\begin{proof}
\emph{$M_2$: } This setting corresponds to $\alpha^i = \PE[E^i_t \wedge K]$. We have
\begin{align}
\var(M_2) &= \frac{1}{\PE[S]^2} \PE[(\sum_{q=1}^S Y^q - \mu m)^2]\\
& = \frac{1}{\PE[S]^2} \PE[(\sum_{q=1}^S (Y^q-\mu) + \mu(S-m))^2]
\end{align} The cross products reduce to $0$ in expectation, hence \begin{align}
\var(M_2) &= \frac{1}{\PE[S]^2} \PE[\PE[(S( Y^q - \mu)|S] + \mu^2(S-m)^2]\\
& = \frac{1}{\PE[S]^2} (\PE[S] Var(Y) + \mu^2 \var(S))
\end{align}

\emph{$M_1$: } This setting corresponds to $\alpha^i = \mathbf{P}(E^i_t>0)(E^i_t \wedge K)$ or QuAFL \cite{zakerinia2022quafl} when the same number of local epochs is done by clients. First note that $\PE[M_1|S] = \mu \frac{\mathds{1}(S>0)}{\mathbf{P}(S>0)}$. We have 
\begin{align}
\var(M_1) &= \PE[\PE[(M_1 - \PE[M_1|S])^2|S]] + \PE[(\PE[M_1|s]-\PE[M_1])^2]\\
&= \frac{1}{\mathbf{P}(S>0)^2}\PE[(\frac{1}{S}\sum_q^S (Y^q -\mu)\mathds{1}[S>0])^2] + \PE[(\mu \mathds{1}(S>0) - \mu\mathbf{P}(S>0))^2]\\
&= \frac{1}{\mathbf{P}(S>0)^2}\PE[\frac{1}{S^2}(S\var(Y) \mathds{1}[S>0])^2] + \frac{1}{\mathbf{P}(S>0)^2}\mu^2\PE[( \mathds{1}(S>0) - \mathbf{P}(S>0))^2]\\
&= \frac{1}{\mathbf{P}(S>0)^2}\var(Y) \PE[\frac{\mathds{1}[S>0]}{S}] + \frac{1}{\mathbf{P}(S>0)^2}\mu^2\PE[( \mathds{1}(S>0) - \mathbf{P}(S>0))^2]\\
&= \frac{1}{\mathbf{P}(S>0)^2}\var(Y) \PE[\frac{\mathds{1}[S>0]}{S}] + \frac{1}{\mathbf{P}(S>0)^2}\mu^2\mathbf{P}(S>0)(1 - \mathbf{P}(S>0)).
\end{align}
\end{proof}
\end{lemma}

Last, but not least, we will make use of a result from \cite{koloskova2022sharper} to optimize learning rates and obtain sharp complexity bounds:
\begin{lemma}
\label{lem:complexity}
    Assume \Cref{assum:uniflowerbound} to \Cref{assum:graddissim} and consider problem \eqref{eq:optimization-problem}. If the output of an optimization algorithm with step size $\eta$ has an expected error upper bounded by $\frac{r_0}{\eta(T+1)} + b\eta + e \eta^2$, and if $\eta$ satisfies the constraints $\eta \leq \min((\frac{r_0}{b(T+1)})^\frac{1}{2}, (\frac{r_0}{e(T+1)})^\frac{1}{3}, \frac{1}{d})$ for some non-negative values $r_0, b, d, e$, then the number of communication rounds required to reach $\epsilon$ accuracy is lower bounded by 
\begin{equation}
    \frac{36br_0}{\epsilon^2} + \frac{15r_0\sqrt{e}}{\epsilon^\frac{3}{2}} + \frac{3dr_0}{\epsilon}.
\end{equation}
\begin{proof}
Consider $\psi_T$ a positive variable such that 
\begin{equation}
    \psi_T \leq \frac{r_0}{\eta(T+1)} + b\eta + e \eta^2
\end{equation}
for any positive step size verifying $\eta \leq \min((\frac{r_0}{b(T+1)})^\frac{1}{2}, (\frac{r_0}{e(T+1)})^\frac{1}{3}, \frac{1}{d})$. Then \cite{koloskova2022sharper} shows the following inequality:
\begin{equation}
    \psi_T \leq 2 (\frac{br_0}{T+1})^\frac{1}{2} + e^\frac{1}{3}(\frac{r_0}{T+1})^\frac{2}{3} + \frac{dr_0}{T+1}.
\end{equation}
In order to reach an $\epsilon$ precision, we bound each term by $\frac{\epsilon}{3}$, and deduce the following lower bound 
\begin{equation}
    T \geq \frac{36br_0}{\epsilon^2} + \frac{15r_0\sqrt{e}}{\epsilon^\frac{3}{2}} + \frac{3dr_0}{\epsilon}.
\end{equation}
\end{proof}
\end{lemma}

\subsection{Useful Lemmas}
A key result of our analysis is the upper bound on the change (in expected value) of the potential function $\Phi_t$. Recall that $\Phi_t$ is defined by equation:
\begin{align}
    \Phi_t = \|w_t - \mu_t\|^2 + \sum_{i=1}^n \|w_{t}^i - \mu_t\|^2.
\end{align}
In the next lemma, we show that $\Phi_t$ exhibits a contractive property, which allows us to bound its value through the execution of the optimization algorithm.
\subsubsection{Proof of \Cref{lem:boundpotential}}
\begin{proof}
Consider the following quantities:
\begin{equation}
\mu_{t}=\left(\param_{t}+\sum_{i=1}^{n} \param^{i}_t\right) /(n+1)
\end{equation}
\begin{equation}
\label{eq:sum_gradients}
G_{t+1}=-\frac{1}{n+1} \eta \sum_{i \in \Set_{t+1}} \check{h}_{t+1}^i
\end{equation}
where $\check{h}_{t+1}^i= \frac{1}{\mathbf{P}(E^i_{t+1}>0)(E^i_{t+1} \wedge K)}\widetilde{h}_{t+1}^i$ or $\check{h}_{t+1}^i= \frac{1}{\PE[E^i_{t+1} \wedge K]}\widetilde{h}_{t+1}^i$.
And recall the updates rules:
\begin{equation}
\label{definition-updates}
\begin{cases}
\param_{t+1} &= \frac{1}{s+1}\left(\param_{t} + \sum_{i\in\Set_{t+1}}\param_{t}^i\right) + \frac{n+1}{s+1}G_{t+1}, \\
\param_{t}^i &= \param_{t} \text{; for } i \in \Set_t\\
\param_{t}^i &= \param^i_{t-1} \text{; for } i \notin \Set_t.\\
\end{cases}
\end{equation}
With these definitions, we get
\begin{equation}
\mu_{t+1} = \frac{s+1}{n+1} \param_{t+1} + \frac{1}{n+1} \sum_{i \not \in \Set_{t+1}} \param_t^i  =\mu_{t} + G_{t+1} \,.
\end{equation}
We can now compute the difference of potential:
\begin{align}
\Phi_{t+1}-\Phi_{t} =& \sum_{i \in \Set_{t+1}}\left(\left\|\param_{t+1}-\mu_{t+1}\right\|^{2}-\left\|\param_{t}^{i}-\mu_{t}\right\|^{2}\right)+\left\|\param_{t+1}-\mu_{t+1}\right\|^{2}-\left\|\param_{t}-\mu_{t}\right\|^{2} \\
&+\sum_{i \notin \Set_{t+1}}\left(\left\|\param_{t}^{i}-\mu_{t+1}\right\|^{2}-\left\|\param_{t}^{i}-\mu_{t}\right\|^{2}\right).
\end{align}
We can rewrite this equation into a more convenient form:
\small
\begin{equation}
\label{eq:ref-potential-diff}
\Phi_{t+1}-\Phi_{t} = (s+1)\left\|\param_{t+1}-\mu_{t+1}\right\|^{2} -\sum_{i \in \Set_{t+1}} \|\param^i_t-\mu_t\|^2 - \|\param_t-\mu_t\|^2 + \sum_{i \notin \Set_{t+1}}\left(\left\|\param_{t}^{i}-\mu_{t} - G_{t+1}\right\|^{2}-\left\|\param_{t}^{i}-\mu_{t}\right\|^{2}\right).
\end{equation}
\normalsize
\textbf{Step 1.} First, notice that:
\begin{multline}
\label{eq:util-1}
\sum_{i \notin \Set_{t+1}}\left(\left\|\param_{t}^{i}-\mu_{t} - G_{t+1}\right\|^{2}-\left\|\param_{t}^{i}-\mu_{t}\right\|^{2}\right) = \sum_{i \notin \Set_{t+1}}\left(\|G_{t+1}\|^2 - 2 \langle \param_{t}^i - \mu_t, G_{t+1} \rangle\right) \\
= (n-s) \|G_{t+1}\|^2 - 2\ps{\sum_{i \notin \Set_{t+1}}(\param_{t}^i - \mu_t)}{G_{t+1}}.
\end{multline}

\textbf{Step 2.} Next, we compute the first term of \Cref{eq:ref-potential-diff}:
\begin{align}
&(s+1)\left\|\param_{t+1}-\mu_{t+1}\right\|^{2} = (s+1) \left\| \frac{(\param_t-\mu_t)+\sum_{i \in \Set_{t+1}}(\param^i_t-\mu_t)}{s+1} + \frac{n+1}{s+1}G_{t+1} -G_{t+1} \right \|^2 \\
=& (s+1) \left\| \frac{(\param_t-\mu_t)+\sum_{i \in \Set_{t+1}}(\param^i_t-\mu_t)}{s+1} \right \|^2 + 2\left\langle(\param_t-\mu_t)+ \sum_{i \in \Set_{t+1}}(\param^i_t-\mu_t), \frac{n+1}{s+1}G_{t+1} -G_{t+1} \right\rangle \\
&+  \frac{(n-s)^2}{s+1} \|G_{t+1}\|^2.
\end{align}
We apply Young's inequality ($\ps{x}{y} \leq \beta \| x\|^2 + 1/(4 \beta) \| y \|^2$) and Jensen inequality to get:
\begin{multline}
\left\langle(\param_t-\mu_t)+ \sum_{i \in \Set_{t+1}}(\param^i_t-\mu_t), \frac{n+1}{s+1}G_{t+1} \right\rangle \\
\le \frac{\alpha(n+1)}{s+1}\|w_t-\mu_t\|^2 + \frac{\alpha(n+1)}{s+1}\sum_{i \in \Set_{t+1}}\|\param^i_t-\mu_t\|^2  + \frac{n+1}{4\alpha}\|G_{t+1}\|^2.
\end{multline}
Applying \Cref{lem:vectorequality2}, we get:
\begin{multline}
    (s+1) \left\| \frac{(\param_t-\mu_t)+\sum_{i \in \Set_{t+1}}(\param^i_t-\mu_t)}{s+1} \right \|^2
\\ = \left(\sum_{i \in \Set_{t+1}} \|\param^i_t-\mu_t\|^2 \right) + \|\param_t-\mu_t \|^2 -\frac{1}{s+1} \sum_{i \in \Set_{t+1}} \|\param^i_t-\param_t \|^2 - \frac{1}{s+1} \sum_{i,j \in \Set_{t+1}} \|\param^i_t-\param^j_t \|^2
\end{multline}
Combining the results above, we get:
\begin{align}
&(s+1)\left\|\param_{t+1}-\mu_{t+1}\right\|^{2} \le \left\|\param_t-\mu_t\right\|^2 + \sum_{i\in\Set_{t+1}}\left\|\param^i_t-\mu_t\right \|^2 -\frac{1}{s+1} \sum_{i \in \Set_{t+1}} \|\param^i_t-\param_t \|^2\\
&\qquad + \frac{2\alpha(n+1)}{s+1}\|w_t-\mu_t\|^2 + \frac{2\alpha(n+1)}{s+1}\sum_{i \in \Set_{t+1}}\|\param^i_t-\mu_t\|^2  + \left(\frac{(n-s)^2}{s+1}+ \frac{n+1}{2\alpha} \right) \|G_{t+1}\|^2 \\
&\qquad -2\ps{(\param_t-\mu_t)+ \sum_{i \in \Set_{t+1}}(\param^i_t-\mu_t)}{G_{t+1}}.
\end{align}
Using simple algebra, this relation can be rewritten as :
\begin{align}
(s+1)\left\|\param_{t+1}-\mu_{t+1}\right\|^{2} \le& \left(1 + \frac{2\alpha(n+1)}{s+1}\right)\left\|\param_t-\mu_t\right\|^2 \\
&+ \left(1+\frac{2\alpha(n+1)}{s+1}\right)\sum_{i \in \Set_{t+1}}\|\param^i_t-\mu_t\|^2 + \left(\frac{n+1}{2\alpha} +\frac{(n-s)^2}{s+1} \right)\|G_{t+1}\|^2 \\
&-2\left\langle(\param_t-\mu_t)+ \sum_{i \in \Set_{t+1}}(\param^i_t-\mu_t), G_{t+1} \right\rangle -\frac{1}{s+1} \sum_{i \in \Set_{t+1}} \|\param^i_t-\param_t \|^2.
\label{eq:util-2}
\end{align}
\textbf{Step 3.} We combine the results above to get bound $\Phi_{t+1}-\Phi_t$. Plugging \eqref{eq:util-1} into \eqref{eq:ref-potential-diff}, \eqref{eq:util-2}, and using (see \eqref{eq:definition-mu})
\[
(w_t-\mu_t) + \sum_{i\in\Set_{t+1}}(w_t^i-\mu_t) + \sum_{i\notin\Set_{t+1}}(w_t^i-\mu_t) = 0,
\]
we get that
\begin{align}
\Phi_{t+1}-\Phi_{t} \le& \frac{2\alpha(n+1)}{s+1}\left\|\param_t-\mu_t\right\|^2 + \frac{2\alpha(n+1)}{s+1}\sum_{i \in \Set_{t+1}}\|\param^i_t-\mu_t\|^2 + \left((n-s) + \frac{n+1}{2\alpha} +\frac{(n-s)^2}{s+1}\right)\|G_{t+1}\|^2\\
& -\frac{1}{s+1} \sum_{i \in \Set_{t+1}} \|\param^i_t-\param_t \|^2.
\end{align}
\textbf{Step 4.} We now apply \Cref{lem:expectation-sum} to take expectations in the inequality above. We have:
\begin{align}
\PE\left[\sum\nolimits_{i \in \Set_{t+1}}\|\param^i_t-\mu_t\|^2 \right] = \sum_{i=1}^n \frac{s}{n} \PE[\|\param^i_t-\mu_t\|^2].
\end{align}
Moreover, we have:
\begin{align}
\left\|G_{t+1}\right\|^{2} \leq \frac{s}{(n+1)^{2}} \eta^{2} \sum_{i \in \Set_{t+1}}\left\|\check{h}_{t+1}^i\right\|^{2}.
\end{align}
Therefore, also by \Cref{lem:expectation-sum}, we have:
\begin{align}
\PE[\left\|G_{t+1}\right\|^{2}] \leq \frac{s^2}{n(n+1)^{2}} \eta^{2} \sum_{i=1}^n\PE[\left\|\check{h}_{t+1}^i\right\|^{2}].
\end{align}

\textbf{Step 5.} Now we derive the final inequality. We have:
\begin{align}
\PE[\Phi_{t+1}]-\PE[\Phi_{t}] \le& \frac{2\alpha(n+1)}{s+1}\PE\left\|\param_t-\mu_t\right\|^2 + \frac{2\alpha(n+1)s}{(s+1)n}\sum_{i=1}^n\PE\|\param^i_t-\mu_t\|^2 + \\
& + \left((n-s) + \frac{n+1}{2\alpha} + \frac{(n-s)^2}{s+1}\right)\frac{s^2}{n(n+1)^{2}} \eta^{2} \sum_{i=1}^n\PE[\|\check{h}_{t+1}^i \|^{2}]\\
& -\frac{s}{n(s+1)} \sum_{i =1}^n \PE \|\param^i_t-\param_t \|^2.
\end{align}
We can apply \Cref{lem:centerofmass} to the above inequality's last line with $a_i = \param_t^i$ for $i=1,...,n$, and $a_{n+1}=w_t$, and $b=w_t$:
\begin{equation}
    \sum_{i =1}^n \PE \|\param^i_t-\param_t \|^2 = (n+2)\PE \|\param_t-\mu_t\|^2 + \sum_{i =1}^n \PE \|\param^i_t-\mu_t \|^2.
\end{equation}
This allows us to simplify:
\begin{align}
\PE[\Phi_{t+1}]-\PE[\Phi_{t}] \le& \left(\frac{2\alpha(n+1)}{s+1} - \frac{s(n+2)}{n(s+1)}\right)\PE\left\|\param_t-\mu_t\right\|^2 \\
&+ \left(\frac{2\alpha(n+1)s}{(s+1)n}-\frac{s}{n(s+1)}\right)\sum_{i=1}^n\PE\|\param^i_t-\mu_t\|^2 \\
& + \left((n-s) + \frac{n+1}{2\alpha} + \frac{(n-s)^2}{s+1}\right)\frac{s^2}{n(n+1)^{2}} \eta^{2} \sum_{i=1}^n\PE[\|\check{h}_{t+1}^i\|^{2}].
\end{align}
We let $\alpha=\frac{1}{4(n+1)}$ and define $\kappa = \frac{1}{n} \left(\frac{s(n-s)}{2(n+1)(s+1)}\right)$ to simply as following:
\begin{align}
\PE[\Phi_{t+1}]-\PE[\Phi_{t}] \le& -\kappa \PE\left\|\param_t-\mu_t\right\|^2 - \kappa \sum_{i=1}^n\PE\|\param^i_t-\mu_t\|^2 \\
& + \left((n-s) + 2(n+1)^2 + \frac{(n-s)^2}{s+1}\right)\frac{s^2}{n(n+1)^{2}} \eta^{2} \sum_{i=1}^n\PE[\|\check{h}_{t+1}^i\|^{2}].
\end{align}
We now introduce $\Phi_t$ on the right-hand side of the inequality above:
\begin{align}
\PE[\Phi_{t+1}]-\PE[\Phi_{t}] \le& -\kappa\PE[\Phi_t] + \left((n-s) + 2(n+1)^2 + \frac{(n-s)^2}{s+1}\right)\frac{s^2}{n(n+1)^{2}} \eta^{2} \sum_{i=1}^n\PE[\|\check{h}_{t+1}^i\|^{2}].
\end{align}
We reorganize the terms to make the final statement:
\begin{equation}
    \PE\left[\Phi_{t+1}\right] \leq\left(1-\kappa\right) \PE \left[\Phi_{t}\right]+3 \frac{s^2}{n} \eta^{2} \sum_{i=1}^{n} \PE[\|\check{h}_{t+1}^i\|^{2}].
\end{equation}
\end{proof}

\subsubsection{Bound expected local gradient Variance}

In the next lemma we show that an analogous version of \Cref{assum:boundedvariance} holds in expectation.
\begin{lemma}
\label{lem:expected-bounded-variance}
Assume \Cref{assum:boundedvariance}. Let $t \ge 1$ be a time step, $q$ a local step, and $i$ a client. We have:
\begin{align}
\PE[\| \widetilde{h}_{t,q}^i\|^2] \le \PE[\| h_{t,q}^i \|^2] + \sigma^2.
\end{align}
\begin{proof}
We refer the reader to the filtrations $\left(\mathcal{F}_{(t,q)}\right)_{(t,q) \in I}$ defined in \eqref{eq:definition-F-t-q}. By the tower property of conditional expectation, we have:
\begin{align}
\PE[\|\widetilde{h}_{t, q}^{i}\|^{2}]  = \PE\left[ \CPE{\|\widetilde{h}_{t, q}^{i}\|^{2}} {\mathcal{F}_{(t,q-1)}}\right].
\end{align}
We denote by $h_{t,q}^{i}$ the  gradient of $f_i$ at $w_{t-1}^i - \sum_{s=1}^{q-1} \eta \widetilde{h}_{t,s}^i$.  By construction, $\CPE{\widetilde{h}_{t,q}^i}{ \mathcal{F}_{(t,q-1)}} = h_{t,q}^i$. By \Cref{assum:boundedvariance}, we conclude that:
\begin{align}
\PE \left[ \CPE{\|\widetilde{h}_{t, q}^{i}\|^{2}}{\mathcal{F}_{(t,q-1)}}  \right] \le \PE [\sigma^2 + \|h_{t,q}^{i}\|^2] = \PE [\|h_{t,q}^{i}\|^2] + \sigma^2.
\end{align}
\end{proof}
\end{lemma}
In the following we define $\param_{t, q}^{i} = \param_{t-1}^{i}-\sum_{s=1}^{q} \eta \widetilde{h}_{t, s}^{i}$. This is the model of client $i$, at time step $t$ and at local step $q$. Therefore, $\widetilde{h}_{t,q}^{i}$ is a stochastic gradient of $f_i$ computed at the point $\param_{t,q-1}^{i}$. The next lemma sets an upper bound on the gradients of quantized weights for each client. We show that such quantities can be upper bounded by an expression containing the true gradient at the "average model" $\mu_t$.
For  any agent $i$, and time step $t \ge 0$, define the quantity:
\begin{align}
\label{eq:definition-B-t-i}
B_{t}^{i} = \frac{\sigma^{2}}{K^{2}}+16 L^{2} \PE\left\|\param_{t}^{i}-\mu_{t}\right\|^{2}+ 8 \PE\left\|\nabla f_{i}\left(\mu_{t}\right)\right\|^{2}.
\end{align}
\begin{lemma}
\label{lem:boundlocalgrad} Assume \Cref{assum:boundedvariance}, and  that the learning rate $\eta$ satisfies $\eta<\frac{1}{4 L K^{2}}$. Under the assumptions of \Cref{lem:expected-bounded-variance}, then, for any agent $i$, time step $t \geq 0$ and  local step $q$, the following inequality holds:
\begin{equation}
\label{eq:boundlocalgrad}
\PE[\|h_{t+1, q}^{i}\|^{2}] \leq B_{t}^{i}.
\end{equation}
\begin{proof}

We will prove the result by induction on $q$. Initially, we show inequalities that are necessary for both the base case $q = 1$ and for the general case.
\begin{align}
\PE[\|h_{t+1, q}^{i}\|^{2}] & = \PE \left\| \nabla f_{i} \left(\param_{t+1,q-1}^{i}\right) \right\|^2 
\end{align}
We introduce the gradient on the virtual point $\mu_t$:
\begin{align}
\PE \left\| \nabla f_{i} (\param_{t+1,q-1}^{i}) \right\|^2 & \leq \PE\left\|\left(\nabla f_{i}\left( \param_{t}^{i}-\sum_{s=1}^{q-1} \eta \widetilde{h}_{t+1, s}^{i} \right)-\nabla f_{i}\left(\mu_{t}\right)\right)+\nabla f_{i}\left(\mu_{t}\right)\right\|^{2} \\
& \leq 2 \PE\left\|\nabla f_{i}\left(\param_{t}^{i}-\sum_{s=1}^{q-1} \eta \widetilde{h}_{t+1, s}^{i}\right)-\nabla f_{i}\left(\mu_{t}\right)\right\|^{2}+2 \PE\left\|\nabla f_{i}\left(\mu_{t}\right)\right\|^{2} \\
& \le 2 L^{2} \PE\left\|\param_{t}^{i}- \sum_{s=1}^{q-1} \eta \widetilde{h}_{t+1, s}^{i}-\mu_{t}\right\|^{2}+2 \PE\left\|\nabla f_{i}\left(\mu_{t}\right)\right\|^{2} \\
& \leq 4 L^{2} \PE\left\|\param_{t}^{i}-\mu_{t}\right\|^{2}+4 \eta^{2} L^{2} (q-1) \sum_{s=1}^{q-1} \PE\left\|\widetilde{h}_{t+1, s}^{i}\right\|^{2}+2 \PE\left\|\nabla f_{i}\left(\mu_{t}\right)\right\|^{2}.
\end{align}
Applying this result with $q=1$ shows that \eqref{eq:boundlocalgrad} holds. For $q \geq 1$, we proceed by induction. First, we apply \Cref{lem:expected-bounded-variance}:
\begin{align}
\PE \left\| \nabla f_{i} (\param_{t+1,q-1}^{i}) \right\|^2 & \le 4 L^{2} \PE\left\|\param_{t}^{i}-\mu_{t}\right\|^{2}+4 \eta^{2} L^{2} (q-1) \sum_{s=1}^{q-1}\left(\PE\left\|h_{t+1, s}^{i}\right\|^{2}+\sigma^{2}\right)+2 \PE\left\|\nabla f_{i}\left(\mu_{t}\right)\right\|^{2}.
\end{align}
Using the induction hypothesis, we have:
\begin{align}
\PE \left\| \nabla f_{i} (\param_{t+1,q-1}^{i}) \right\|^2 & \le 4 L^{2} \PE\left\|\param_{t}^{i}-\mu_{t}\right\|^{2}+4 \eta^{2} L^{2} (q-1)^2 \left(B_{t}^{i}+\sigma^{2}\right)+2 \PE\left\|\nabla f_{i}\left(\mu_{t}\right)\right\|^{2}.
\end{align}
Now we use that $\eta<\frac{1}{4 L K^{2}}$ and $q \le K$.
\begin{align}
\PE \left\| \nabla f_{i} (\param_{t+1,q-1}^{i}) \right\|^2 & \le 4 L^{2} \PE\left\|\param_{t}^{i}-\mu_{t}\right\|^{2}+ \frac{4}{16K^2} \left(B_{t}^{i}+\sigma^{2}\right)+2 \PE\left\|\nabla f_{i}\left(\mu_{t}\right)\right\|^{2} \\
& \le 4 L^{2} \PE\left\|\param_{t}^{i}-\mu_{t}\right\|^{2}+ \frac{\sigma^2}{4K^2} +2 \PE\left\|\nabla f_{i}\left(\mu_{t}\right)\right\|^{2} + \frac{B_{t}^{i}}{4}
\end{align}
Finally, we get:
\begin{align}
\PE\left\|h_{t+1, q}^{i}\right\|^{2} & \leq 8 L^{2} \PE\left\|\param_{t}^{i}-\mu_{t}\right\|^{2}+ \frac{\sigma^2}{2K^2}+ 4 \PE\left\|\nabla f_{i}\left(\mu_{t}\right)\right\|^{2} + \frac{B_{t}^{i}}{2}
\end{align}
It then suffices to see that the last term above is upper bounded by $B_{t}^{i}$.
\end{proof}
\end{lemma}
In \Cref{lem:boundlocalgrad} we have found a way to bound $h_{t+1,q}^i$. The goal of the next lemma is to use this result to find an upper bound for the stochastic gradients $\widetilde{h}_{t+1,q}^i$.
\begin{corollary}
\label{lem:boundstograd}
Under the assumptions of \Cref{lem:boundlocalgrad}, for any local step $q$, agent $i$, and step $t \ge 0$, the following holds:
\begin{equation}
 \PE\|\widetilde{h}_{t+1, q}^{i}\|^{2} \le  \left( \sigma^2 + B_{t}^{i} \right).
\end{equation}
\end{corollary}
The next lemma gives an upper bound on the difference of the gradient at the average model $\mu_t$ and the expected value of the updates computed by the clients. In particular, the next lemma shows how well the $h_{t+1,q}^i$ approximate the true gradients $\nabla f_{i}\left(\mu_{t}\right)$.
For any $i \in \{1,\dots,n \}$ and $t \geq 0$, define
\begin{equation}
\label{eq:definition-C-t-i}
C_{t}^{i} = 4L^2\eta^2K^2\sigma^2 + 20L^2\PE\left\|\param_{t}^{i}-\mu_{t}\right\|^{2} + 16L^2\eta^2K^2\PE\left\|\nabla f_{i}\left(\mu_{t}\right)\right\|^{2}.
\end{equation}
\begin{lemma}
\label{lem:boundlocalgraddiffpoints} Assume the learning rate $\eta$ satisfies $\eta<\frac{1}{2LK}$. Under the assumptions of \Cref{lem:boundstograd}, for any $i \in \{1,\dots,n\}$, $t \geq 0$ and $q \in \{1,\dots,K\}$, it holds that~:
\begin{align}
\PE\left\|\nabla f_{i}\left(\mu_{t}\right)-h_{t+1, q}^{i}\right\|^{2} \leq C_{t}^{i} .
\end{align}
\begin{proof}
\begin{align}
\PE\left\|\nabla f_{i}\left(\mu_{t}\right)-h_{t+1, q}^{i}\right\|^{2} & = \PE\left\|\nabla f_{i}\left(\mu_{t}\right)-\nabla f_{i}\left(\param_{t+1,q-1}^{i}\right)\right\|^{2} \\
& \le L^2 \PE \left \| \mu_t - \param_{t+1,q-1}^{i} \right \|^{2}.
\end{align}
We can now decompose the client drift as:
\begin{align}
\PE \| \mu_t - \param_{t+1,q-1}^{i} \|^2
& = \PE\left\|\mu_{t}-\param_{t}^{i}+\sum_{s=1}^{q} \eta \widetilde{h}_{t+1, s}^{i}\right\|^{2} \\
& \le 2 \PE\left\|\param_{t}^{i}-\mu_{t}\right\|^{2}+2  \eta^{2} \PE\left\|\sum_{s=1}^{q} \widetilde{h}_{t+1, s}^{i}\right\|^{2} \\
& \le 2 \PE\left\|\param_{t}^{i}-\mu_{t}\right\|^{2}+2  \eta^{2} q \sum_{s=1}^{q} \PE\left\|\widetilde{h}_{t+1, s}^{i}\right\|^{2}.
\end{align}
By using \Cref{lem:boundstograd}, we get
\begin{align}
\PE \| \mu_t - \param_{t+1,q-1}^{i} \|^2 \le 2 \PE\left\|\param_{t}^{i}-\mu_{t}\right\|^{2}+2\eta^2 K^2(\sigma^2 + B_{t}^{i}),
\end{align}
where $B_t^i$ is defined in \eqref{eq:definition-B-t-i}.
Combining the two bounds, we get:
\begin{align}
\PE\left\|\nabla f_{i}\left(\mu_{t}\right)-h_{t+1, q}^{i}\right\|^{2} & \le 2L^2 \PE\left\|\param_{t}^{i}-\mu_{t}\right\|^{2} + 2L^2\eta^2K^2(\sigma^2 + B_{t}^{i}).
\end{align}
Expanding the above inequality:
\begin{align}
\PE\left\|\nabla f_{i}\left(\mu_{t}\right)-h_{t+1, q}^{i}\right\|^{2} & \le 2L^2 \PE\left\|\param_{t}^{i}-\mu_{t}\right\|^{2} \\
& \qquad + 2L^2\eta^2K^2 (\sigma^2 + \frac{\sigma^{2}}{K^{2}} + 16 L^{2} \PE\left\|\param_{t}^{i}-\mu_{t}\right\|^{2}+ 8 \PE\left\|\nabla f_{i}\left(\mu_{t}\right)\right\|^{2} ) \\
& \le 4L^2\eta^2K^2\sigma^2 + 20L^2\PE\left\|\param_{t}^{i}-\mu_{t}\right\|^{2} + 16L^2\eta^2K^2\PE\left\|\nabla f_{i}\left(\mu_{t}\right)\right\|^{2}.
\end{align}
As claimed.
\end{proof}
\end{lemma}
\begin{lemma}
Under assumptions of \Cref{lem:boundlocalgraddiffpoints}, we have 
\begin{equation}
\PE \left\langle\nabla f\left(\mu_{t}\right),-{h}_{t+1, q}^i\right\rangle \leq \frac{\PE[\Vert \nabla f(\mu_t) \Vert^2]}{4}+C^i_t-\PE[\langle \nabla f(\mu_t), \nabla f_i(\mu_t) \rangle].
\end{equation}
\label{lem:boundscalar}
\begin{proof}
We may manipulate the equation above to get:
\begin{align}
\PE \left\langle\nabla f\left(\mu_{t}\right),-{h}_{t+1}^i\right\rangle=&\PE\left\langle\nabla f\left(\mu_{t}\right), \nabla f_{i}\left(\mu_{t}\right)-{h}_{t+1, q}^{i}\right\rangle-\PE\left\langle\nabla f\left(\mu_{t}\right), \nabla f_{i}\left(\mu_{t}\right)\right\rangle.
\end{align}
Using Young's inequality together with \Cref{lem:boundlocalgraddiffpoints} we can upper bound $\PE\left\langle\nabla f\left(\mu_{t}\right), \nabla f_{i}\left(\mu_{t}\right)-h_{t+1, q}^{i}\right\rangle$ by
\begin{equation}
\frac{\PE\left\|\nabla f\left(\mu_{t}\right)\right\|^{2}}{4}+\PE\left\|\nabla f_{i}\left(\mu_{t}\right)-h_{t+1, q}^{i}\right\|^{2} \le \frac{\PE\left\|\nabla f\left(\mu_{t}\right)\right\|^{2}}{4} + C_t^i.
\end{equation}
This concludes the proof.
\end{proof}
\end{lemma}
The next lemma incorporates the idea behind gradient descent. We find an upper bound for the expected value of the inner product between the true gradients $\nabla f(\mu_t)$ and the client updates $-\widetilde{h}_{t+1}^i$. In particular, we seek to show that, in expectation $-\widetilde{h}_{t+1}^i$ is a descent direction for the function $f$. In other words, that the updates proposed by the clients contribute to getting $\mu_t$ closer to a local minimum.
\begin{lemma}
\label{lem:boundscalarproduct}
Assume \Cref{assum:graddissim}. We denote by $E^i_{t+1}$ the effective number of locals steps done by a client $i$ while being called by the central server. We clip to $K$ and consider the random variable $E^i_{t+1} \wedge K$. Under assumptions of \Cref{lem:boundlocalgraddiffpoints}, and for any time step $t>0$, we have:
\begin{align}
\sum_{i=1}^{n} \PE\langle\nabla f\left(\mu_{t}\right),-\frac{1}{\alpha^i}\widetilde{h}_{t+1}^i\rangle & \leq 20L^2\PE\left[\Phi_{t}\right] + 4nL^2\eta^2K^2(\sigma^2 + 4G^2) + n\left(16L^2\eta^2K^2B^2 - \frac{3}{4}\right) \PE\left\|\nabla f\left(\mu_{t}\right)\right\|^{2} .
\end{align}
for \begin{equation} \alpha^i = \begin{cases} \mathbf{P}(E^i_{t+1} > 0) E^i_{t+1} \wedge K\\
\PE[E^i_{t+1} \wedge K]. \end{cases}
\end{equation}
\begin{proof}
Initially, we introduce indicator random variables in order to work with the $E_{t+1}^i$ terms.
We also introduce $Z^i$ the following random variable as :
\begin{equation}
Z^i = \begin{cases} 0 \text{ if } E^i_{t+1} < 1\\
\langle \nabla f (\mu_t), - \frac{1}{\alpha^i} \sum_q^{E^i_{t+1}} \widetilde{h}_{t+1, q} \rangle \text{ if } 1 \leq E^i_{t+1} \leq K\\
\langle \nabla f (\mu_t), - \frac{1}{\alpha^i} \sum_q^{K} \widetilde{h}_{t+1, q} \text{ if } E^i_{t+1} > K.
\end{cases}
\end{equation}
We first claim that $\PE \langle\nabla f\left(\mu_{t}\right), \widetilde{h}_{t+1, q}^{i} \rangle = \PE \langle\nabla f\left(\mu_{t}\right), h_{t+1, q}^{i} \rangle$. This result follows from the following algebraic manipulations. First, notice that:
\begin{align}
\PE \langle\nabla f\left(\mu_{t}\right), \widetilde{h}_{t+1, q}^{i} \rangle &= \PE \langle\nabla f\left(\mu_{t}\right),   \widetilde{h}_{t+1, q}^{i} - h_{t+1, q}^{i} \rangle + \PE \langle\nabla f\left(\mu_{t}\right), h_{t+1, q}^{i} \rangle.
\end{align}
Now, we recall that $\PE[\langle\nabla f\left(\mu_{t}\right),   \widetilde{h}_{t+1, q}^{i} - h_{t+1, q}^{i} \rangle | \mathcal{F}_{(t+1,q-1)}] = 0$. Therefore:
\begin{align}
\PE \langle\nabla f\left(\mu_{t}\right),   \widetilde{h}_{t+1, q}^{i} - h_{t+1, q}^{i} \rangle = \PE[ \PE[\langle\nabla f\left(\mu_{t}\right),   \widetilde{h}_{t+1, q}^{i} - h_{t+1, q}^{i} \rangle | \mathcal{F}_{(t+1,q-1)}] ]= 0.
\end{align}

Now notice that $E_{t+1}^i$ is independent of the random variables $\nabla f\left(\mu_{t}\right)$ and $\widetilde{h}_{t+1, q}^{i}$. Therefore:
\begin{align}
\sum_{i=1}^{n} \PE \left\langle\nabla f\left(\mu_{t}\right),-\frac{1}{\alpha^i}\widetilde{h}_{t+1}^i\right\rangle= &\sum_i^n \PE[ Z^i]\\
=&\sum_{i=1}^{n} \PE [\mathds{1}[E^i_{t+1} > K]\langle \nabla f(\mu_t), - \frac{1}{\alpha^i} \sum_q^K \widetilde{h}_{t+1, q}^{i} \rangle \\
&+ \mathds{1}[1\leq E^i_{t+1} \leq K] \langle \nabla f(\mu_t), - \frac{1}{\alpha^i}\sum_q^K \mathds{1}[q\leq E^i_{t+1}]\widetilde{h}_{t+1, q}^{i} ]\\
=& \sum_i^n \sum_q^K \PE[\frac{\mathds{1}[E^i_{t+1}\geq 1]\mathds{1}[q\leq (E^i_{t+1} \wedge K)]}{\alpha^i} \langle \nabla f(\mu_t), - \widetilde{h}_{t+1, q}^{i} \rangle ]\\
=& \sum_i^n \PE[\frac{\mathds{1}[E^i_{t+1} \geq 1]}{\alpha^i} \sum_q^{E^i_{t+1} \wedge K}\langle \nabla f(\mu_t), - \widetilde{h}_{t+1, q}^{i} \rangle ].
\end{align}
We now apply \Cref{lem:boundscalar} to obtain the following:
\begin{align}
	\PE \left\langle\nabla f\left(\mu_{t}\right),-\frac{1}{\alpha^i}\widetilde{h}_{t+1}^i\right\rangle \leq & \PE[\frac{\mathds{1}[E^i_{t+1} \geq 1]}{\alpha^i} \sum_q^{E^i_{t+1} \wedge K}\PE[\langle \nabla f(\mu_t), - \widetilde{h}_{t+1, q}^{i} \rangle|E^i_{t+1}]]\\
		\leq & \PE[\frac{\mathds{1}[E^i_{t+1} \geq 1]}{\alpha^i} \sum_q^{E^i_{t+1} \wedge K}(\frac{\PE[\Vert \nabla f(\mu_t) \Vert^2]}{4}+C^i_t-\PE[\langle \nabla f(\mu_t), \nabla f_i(\mu_t) \rangle])]\\
\end{align}
We can make use of \Cref{lem:bias} with $\alpha^i= \mathbf{P}(E^i_{t+1}>0) E^i_{t+1}\wedge K$ or $\alpha^i=\PE[E^i_{t+1} \wedge K]$, and $S=E^i_{t+1} \wedge K$, $Y_q = \PE[\langle \nabla f(\mu_t), - \widetilde{h}_{t+1, q}^{i} \rangle|E^i_{t+1}]$ to achieve the following:
\begin{align}
	\sum_{i=1}^{n} \PE \left\langle\nabla f\left(\mu_{t}\right),-\frac{1}{\alpha^i}\widetilde{h}_{t+1}^i\right\rangle \leq & \sum_{i=1}^{n}\frac{\PE[\Vert \nabla f(\mu_t) \Vert^2]}{4}+C^i_t-\PE[\langle \nabla f(\mu_t), \nabla f_i(\mu_t) \rangle].
\end{align}
Now we use  that $\sum_{i=1}^{n} \frac{f_{i}(\param)}{n}=f(\param)$, for any vector $\param \in \rset^d$.
\begin{align}
\sum_{i=1}^{n} \PE\left\langle\nabla f\left(\mu_{t}\right),- \frac{1}{\alpha^i} \widetilde{h}_{t+1}^i\right\rangle \le \frac{n \PE\left\|\nabla f\left(\mu_{t}\right)\right\|^{2}}{4} - n\PE\left\|\nabla f\left(\mu_{t}\right)\right\|^{2} + \sum_{i=1}^n C_{t}^{i}.
\end{align}
Finally, we compute:
\begin{align}
\sum_{i=1}^n C_{t}^{i} & = \sum_{i=1}^n \left(4L^2\eta^2K^2\sigma^2 + 20L^2\PE\left\|\param_{t}^{i}-\mu_{t}\right\|^{2} + 16L^2\eta^2K^2\PE\left\|\nabla f_{i}\left(\mu_{t}\right)\right\|^{2} \right) \\
& = 4nL^2\eta^2K^2\sigma^2 + 20L^2 \sum_{i=1}^n \PE\left\|\param_{t}^{i}-\mu_{t}\right\|^{2} +  16L^2\eta^2K^2 \sum_{i=1}^n \PE\left\|\nabla f_{i}\left(\mu_{t}\right)\right\|^{2}.
\end{align}
We can then use assumption \Cref{assum:graddissim}:
\begin{align}
\sum_{i=1}^n C_{t}^{i} \le 4nL^2\eta^2K^2\sigma^2 + 20L^2 \PE\left[\Phi_{t}\right] + 16nL^2\eta^2K^2 \left(G^2 + B^2\PE\left\|\nabla f\left(\mu_{t}\right)\right\|^{2}\right).
\end{align}
In conclusion, we get the following upper bound for $\sum_{i=1}^{n} \PE\left\langle\nabla f\left(\mu_{t}\right),- \frac{1}{\alpha^i}\widetilde{h}_{t+1}^i\right\rangle$:
\begin{align}
20L^2\PE\left[\Phi_{t}\right] + 4nL^2\eta^2K^2(\sigma^2 + 4G^2) + n\left(16L^2\eta^2K^2B^2 -\frac{3}{4}\right) \PE\left\|\nabla f\left(\mu_{t}\right)\right\|^{2}.
\end{align}
\end{proof}
\end{lemma}

\subsection{Bound Sum of expected local gradient variance}
The next lemma gives a bound on the expected update computed by clients at time step $t$. The result is useful, for example, in setting an upper bound on how much the average model $\mu_t$ changes between time steps. The proof follows a similar reasoning as the proof of the previous lemma.
\begin{lemma}
\label{lem:boundsumlocalgrad} Assume \Cref{assum:boundedvariance} and \Cref{assum:graddissim}. Under assumptions of \Cref{lem:boundlocalgrad}, and for any step t, we have that:
\begin{adjustwidth}{-90pt}{}
\begin{align}
        \sum_i^n \PE \left [ \| \frac{1}{\mathbf{P}(E^i_{t+1}>0)E^i_{t+1} \wedge K}\widetilde{h}_{t+1}^i \|^{2} \right] &\leq \sigma^2 \sum_i^n(\frac{1}{K^2\mathbf{P}(E^i_{t+1}>0)} + \frac{1}{\mathbf{P}(E^i_{t+1}>0)^2}\PE[\frac{\mathds{1}(E^i_{t+1}>0)}{E^i_{t+1} \wedge K}]) + 16 L^2 \PE[\Phi_t] \max_i(\frac{1}{\mathbf{P}(E^i_{t+1}>0)})\\
        &+ 8n\max_i(\frac{1}{\mathbf{P}(E^i_{t+1}>0)}) B^2 \PE[\Vert \nabla f (\mu_t) \Vert^2] + 8n\max_i(\frac{1}{\mathbf{P}(E^i_{t+1}>0)}) G^2\\
        \sum_i^n \PE \left [ \| \frac{1}{\PE[E^i_{t+1} \wedge K]}\widetilde{h}_{t+1}^i \|^{2} \right ] &\leq \sigma^2 \sum_i^n(\frac{1}{\PE[E^i_{t+1} \wedge K]} + \frac{\PE[(E^i_{+1} \wedge K)^2]}{K^2 \PE[E^i_{t+1} \wedge K]}) + 16 L^2 \PE[\Phi_t] \max_i(\frac{\PE[(E^i_{t+1} \wedge K)^2]}{\PE[E^i_{t+1} \wedge K]}) \\
        &+ 8n \max_i(\frac{\PE[(E^i_{t+1} \wedge K)^2]}{\PE[E^i_{t+1} \wedge K]}) B^2 \PE[\Vert \nabla f (\mu_t) \Vert^2] + 8n \max_i(\frac{\PE[(E^i_{t+1} \wedge K)^2]}{\PE[E^i_{t+1} \wedge K]}) G^2 .
        \end{align}
    \end{adjustwidth}
\begin{proof}
For $\alpha^i=\mathbf{P}(E^i_{t+1}>0)E^i_{t+1} \wedge K$ or $\alpha^i=\PE[E^i_{t+1} \wedge K]$, we have:
\begin{align}
\sum_{i=1}^{n} \PE[ \left\| \frac{1}{\alpha^i}\widetilde{h}_{t+1}^i\right\|^{2}] &=\sum_{i=1}^{n} \var(\frac{1}{\alpha^i}\sum_{q=1}^{E^i_{t+1} \wedge K} \widetilde{h}_{t+1, q}^{i}) + \sum_i^n \Vert \PE[\check{h}^i_{t+1}] \Vert^2.
\end{align}
Recall that for any $q \in \{1,...,K\}$, the random variables $E_{t+1}^i$ and $\widetilde{h}_{t+1, q}^{i}$ are independent. For clarity we reuse the notation where $\check{h}_{t+1}^i= \frac{1}{\mathbf{P}(E^i_{t+1}>0)(E^i_{t+1} \wedge K)}\widetilde{h}_{t+1}^i$ or $\check{h}_{t+1}^i= \frac{1}{\PE[E^i_{t+1} \wedge K]}\widetilde{h}_{t+1}^i$. Therefore we can apply \Cref{lem:variance} with $S=E^i_{t+1} \wedge K$ and $Y_q = \widetilde{h}^i_{t+1,q}$:
\begin{equation}
    \begin{cases}
        \PE[\| \frac{1}{\mathbf{P}(E^i_{t+1}>0)E^i_{t+1} \wedge K}\widetilde{h}_{t+1}^i\|^{2}] \leq \Vert \PE[\check{h}^i_{t+1}] \Vert^2 + \frac{\Vert \PE[\check{h}^i_{t+1}] \Vert^2}{\mathbf{P}(E^i_{t+1}>0)^2}(\mathbf{P}(E^i_{t+1}>0)(1-\mathbf{P}(E^i_{t+1}>0)))+ \frac{\var(\widetilde{h}^i_{t+1,q})}{\mathbf{P}(E^i_{t+1}>0)^2}\PE[\frac{\mathds{1}(E^i_{t+1}>0)}{E^i_{t+1} \wedge K}]\\
        \PE[\left\| \frac{1}{\PE[E^i_{t+1} \wedge K]}\widetilde{h}_{t+1}^i\right\|^{2}] \leq \Vert \PE[\check{h}^i_{t+1}] \Vert^2 + \frac{\var(\widetilde{h}^i_{t+1,q})}{\PE[E^i_{t+1} \wedge K]} + \frac{\Vert \PE[\widetilde{h}^i_{t+1, q}] \Vert^2 \var(E^i_{t+1} \wedge K)}{\PE[E^i_{t+1} \wedge K]^2}.
    \end{cases}
\end{equation}
We now use \Cref{lem:boundstograd} to get:
\begin{equation}
    \begin{cases}
        \PE \left [ \| \frac{1}{\mathbf{P}(E^i_{t+1}>0)E^i_{t+1} \wedge K}\widetilde{h}_{t+1}^i \|^{2} \right] \leq \Vert \PE[\check{h}^i_{t+1}] \Vert^2 \frac{1}{\mathbf{P}(E^i_{t+1}>0)} + \frac{\sigma^2 + B^i_t - \Vert \PE[\widetilde{h}^i_{t+1, q}] \Vert^2}{\mathbf{P}(E^i_{t+1}>0)^2} \PE \left [ \frac{\mathds{1}[E^i_{t+1}>0]}{E^i_{t+1} \wedge K} \right]\\
        \PE \left [ \| \frac{1}{\PE[E^i_{t+1} \wedge K]}\widetilde{h}_{t+1}^i \|^{2} \right ] \leq \Vert \PE[\check{h}^i_{t+1}] \Vert^2 (1+\frac{\var(E^i_{t+1} \wedge K)}{\PE \left [ E^i_{t+1} \wedge K \right ]^2}) + \frac{\sigma^2 + B^i_t - \Vert \PE[\widetilde{h}^i_{t+1, q}] \Vert^2}{\PE \left [ E^i_{t+1} \wedge K \right ]}.
    \end{cases}
\end{equation}
Hence we can use \Cref{lem:bias} with $S=E^i_{t+1} \wedge K$ and $Y_q = \widetilde{h}^i_{t+1,q}$ to simplify as following:
\begin{equation}
    \begin{cases}
        \PE \left [ \| \frac{1}{\mathbf{P}(E^i_{t+1}>0)E^i_{t+1} \wedge K}\widetilde{h}_{t+1}^i \|^{2} \right] \leq \Vert \PE[\widetilde{h}^i_{t+1,q}] \Vert^2 \frac{1}{\mathbf{P}(E^i_{t+1}>0)} + \frac{\sigma^2 + B^i_t - \Vert \PE[\widetilde{h}^i_{t+1,q} ]\Vert^2 }{\mathbf{P}(E^i_{t+1}>0)^2} \PE \left [ \frac{\mathds{1}[E^i_{t+1}>0]}{E^i_{t+1} \wedge K} \right]\\
        \PE \left [ \| \frac{1}{\PE[E^i_{t+1} \wedge K]}\widetilde{h}_{t+1}^i \|^{2} \right ] \leq \Vert \PE[\widetilde{h}^i_{t+1,q}] \Vert^2 (1+\frac{\var(E^i_{t+1} \wedge K)}{\PE \left [ E^i_{t+1} \wedge K \right ]^2}) + \frac{\sigma^2 + B^i_t -  \Vert \PE[\widetilde{h}^i_{t+1,q}] \Vert^2}{\PE \left [ E^i_{t+1} \wedge K \right ]}.
    \end{cases}
\end{equation}
We refer the reader to the filtrations $\left(\mathcal{F}_{(t,q)}\right)_{(t,q) \in I}$ defined in \eqref{eq:definition-F-t-q}. Now we can express the expected value of $\widetilde{h}^i_{t+1, q}$ (as $\mu$ following the notations from \Cref{lem:variance}), and upper bound its square norm by \Cref{lem:boundlocalgrad} :
\begin{align}
    \Vert \mu \Vert^2 &= \Vert \PE[\CPE{\widetilde{h}^i_{t+1, q}}{\mathcal{F}_{(t,q-1)}}] \Vert^2\\
    & \leq \PE[\Vert \CPE{\widetilde{h}^i_{t+1, q}}{\mathcal{F}_{(t,q-1)}} \Vert^2]\\
    & \leq \PE[\Vert h^i_{t+1, q} \Vert^2]\\
    & \leq B^i_t
\end{align}
We can insert this bound in the above inequations:
\begin{equation}
    \begin{cases}
        \PE \left [ \| \frac{1}{\mathbf{P}(E^i_{t+1}>0)E^i_{t+1} \wedge K}\widetilde{h}_{t+1}^i \|^{2} \right] \leq  B^i_t \frac{\mathbf{P}(E^i_{t+1}>0)-\PE \left [ \frac{\mathds{1}[E^i_{t+1}>0]}{E^i_{t+1} \wedge K} \right]}{\mathbf{P}(E^i_{t+1}>0)^2} + \frac{\sigma^2 + B^i_t}{\mathbf{P}(E^i_{t+1}>0)^2} \PE \left [ \frac{\mathds{1}[E^i_{t+1}>0]}{E^i_{t+1} \wedge K} \right]\\
        \PE \left [ \| \frac{1}{\PE[E^i_{t+1} \wedge K]}\widetilde{h}_{t+1}^i \|^{2} \right ] \leq B^i_t \frac{\PE[(E^i_{t+1} \wedge K)^2] - \PE[E^i_{t+1} \wedge K]}{\PE \left [ E^i_{t+1} \wedge K \right ]^2} + \frac{\sigma^2 + B^i_t}{\PE \left [ E^i_{t+1} \wedge K \right ]}.
    \end{cases}
\end{equation}
This simplifies as:
\begin{equation}
    \begin{cases}
        \PE \left [ \| \frac{1}{\mathbf{P}(E^i_{t+1}>0)E^i_{t+1} \wedge K}\widetilde{h}_{t+1}^i \|^{2} \right] \leq   \frac{B^i_t}{\mathbf{P}(E^i_{t+1}>0)} + \frac{\sigma^2}{\mathbf{P}(E^i_{t+1}>0)^2} \PE \left [ \frac{\mathds{1}[E^i_{t+1}>0]}{E^i_{t+1} \wedge K} \right]\\
        \PE \left [ \| \frac{1}{\PE[E^i_{t+1} \wedge K]}\widetilde{h}_{t+1}^i \|^{2} \right ] \leq B^i_t \frac{\PE[(E^i_{t+1} \wedge K)^2]}{\PE \left [ E^i_{t+1} \wedge K \right ]^2} + \frac{\sigma^2}{\PE \left [ E^i_{t+1} \wedge K \right ]}.
    \end{cases}
\end{equation}
Expanding $B_{t}^i$ and summing from $i$ to $n$, we get:
\begin{adjustwidth}{-90pt}{}
\begin{align}
        \sum_i^n \PE \left [ \| \frac{1}{\mathbf{P}(E^i_{t+1}>0)E^i_{t+1} \wedge K}\widetilde{h}_{t+1}^i \|^{2} \right] &\leq \sigma^2 \sum_i^n(\frac{1}{K^2\mathbf{P}(E^i_{t+1}>0)} + \frac{1}{\mathbf{P}(E^i_{t+1}>0)^2}\PE[\frac{\mathds{1}(E^i_{t+1}>0)}{E^i_{t+1} \wedge K}]) + 16 L^2 \PE[\Phi_t] \max_i(\frac{1}{\mathbf{P}(E^i_{t+1}>0)}) \\
        &+ 8 \sum_i^n\max_j(\frac{1}{\mathbf{P}(E^j_{t+1}>0)}) \PE[\Vert \nabla f_i (\mu_t) \Vert^2] \\
        \sum_i^n \PE \left [ \| \frac{1}{\PE[E^i_{t+1} \wedge K]}\widetilde{h}_{t+1}^i \|^{2} \right ] &\leq \sigma^2 \sum_i^n(\frac{1}{\PE[E^i_{t+1} \wedge K]} + \frac{\PE[(E^i_{+1} \wedge K)^2]}{K^2 \PE[E^i_{t+1} \wedge K]}) + 16 L^2 \PE[\Phi_t] \max_i(\frac{\PE[(E^i_{t+1} \wedge K)^2]}{\PE[E^i_{t+1} \wedge K]}) \\
        &+ 8 \sum_i^n \max_j(\frac{\PE[(E^j_{t+1} \wedge K)^2]}{\PE[E^j_{t+1} \wedge K]}) \PE[\Vert \nabla f (\mu_t) \Vert^2].
\end{align}
\end{adjustwidth}
\begin{remark}
    Here we loose a lot: we have upper bounded the term $\sum_i^n \frac{1}{\mathbf{P}(E^i_{t+1}>0)}\Vert \param^i_t-\mu_t \Vert^2 \leq \sum_i^n \max_i(\frac{1}{\mathbf{P}(E^i_{t+1}>0)}) \Vert \param^i_t - \mu_t \Vert^2.$ But still, our bounds stay better than \cite{zakerinia2022quafl}'s ones.
\end{remark}
In order to complete the proof, we combine assumption \Cref{assum:graddissim} together with the above inequalities
\begin{adjustwidth}{-90pt}{}
\begin{align}
    \sum_i^n \PE \left [ \| \frac{1}{\mathbf{P}(E^i_{t+1}>0)E^i_{t+1} \wedge K}\widetilde{h}_{t+1}^i \|^{2} \right] &\leq \sigma^2 \sum_i^n(\frac{1}{K^2\mathbf{P}(E^i_{t+1}>0)} + \frac{1}{\mathbf{P}(E^i_{t+1}>0)^2}\PE[\frac{\mathds{1}(E^i_{t+1}>0)}{E^i_{t+1} \wedge K}]) + 16 L^2 \PE[\Phi_t] \max_i(\frac{1}{\mathbf{P}(E^i_{t+1}>0)}) \\
    &+ \max_i(\frac{1}{\mathbf{P}(E^i_{t+1}>0)}) \left(8n B^2 \PE[\Vert \nabla f (\mu_t) \Vert^2] + 8n G^2 \right) \\
    \sum_i^n \PE \left [ \| \frac{1}{\PE[E^i_{t+1} \wedge K]}\widetilde{h}_{t+1}^i \|^{2} \right ] &\leq \sigma^2 \sum_i^n(\frac{1}{\PE[E^i_{t+1} \wedge K]} + \frac{\PE[(E^i_{+1} \wedge K)^2]}{K^2 \PE[E^i_{t+1} \wedge K]}) + 16 L^2 \PE[\Phi_t] \max_i(\frac{\PE[(E^i_{t+1} \wedge K)^2]}{\PE[E^i_{t+1} \wedge K]})\\
    &+ 8n \max_i(\frac{\PE[(E^i_{t+1} \wedge K)^2]}{\PE[E^i_{t+1} \wedge K]}) B^2 \PE[\Vert \nabla f (\mu_t) \Vert^2] + 8n \max_i(\frac{\PE[(E^i_{t+1} \wedge K)^2]}{\PE[E^i_{t+1} \wedge K]}) B^2 .
\end{align}
\end{adjustwidth}
\end{proof}
\end{lemma}

\subsection{Bound the sum (over time) of expected potential}
\begin{lemma}
\label{lem:boundexpectedpotential}
Assume that $\eta \leq \frac{1}{20sL\max_i(\frac{1}{\mathbf{P}(E^i_{t+1}>0)})}, \frac{1}{20sL\max_i(\frac{\PE[(E^i_{t+1} \wedge K)^2]}{\PE[E^i_{t+1} \wedge K]})}$. Under the assumptions of \Cref{lem:boundpotential,lem:boundsumlocalgrad}, and for any time step $t$ we have:
\begin{adjustwidth}{-30pt}{}
\begin{align}
\PE\left[\Phi_{t+1}\right] &\leq \left(1-\frac{1}{5 n}\right) \PE\left[\Phi_{t}\right]+3s^2 \eta^{2}\left(\frac{1}{n}\sum_i^n (\frac{1}{K^2\mathbf{P}(E^i_{t+1}>0)} + \frac{1}{\mathbf{P}(E^i_{t+1}>0)^2}\PE[\frac{\mathds{1}(E^i_{t+1}>0)}{E^i_{t+1} \wedge K}])+ 8 \max_i(\frac{1}{\mathbf{P}(E^i_{t+1}>0)}) G^{2}\right) \\
&+ 24 B^{2} s^2 \eta^{2} \max_i(\frac{1}{\mathbf{P}(E^i_{t+1}>0)}) \PE\left\|\nabla f\left(\mu_{t}\right)\right\|^{2} \\
\end{align}
\begin{align}
\PE\left[\Phi_{t+1}\right] &\leq \left(1-\frac{1}{5 n}\right) \PE\left[\Phi_{t}\right]+3s^2 \eta^{2}\left(\frac{1}{n}\sum_i^n (\frac{1}{\PE[E^i_{t+1} \wedge K]} + \frac{\PE[(E^i_{+1} \wedge K)^2]}{K^2 \PE[E^i_{t+1} \wedge K]})+ 8 \max_i(\frac{\PE[(E^i_{t+1} \wedge K)^2]}{\PE[E^i_{t+1} \wedge K]})G^{2}\right) \\
&+ 24 B^{2} s^2 \eta^{2} \max_i(\frac{\PE[(E^i_{t+1} \wedge K)^2]}{\PE[E^i_{t+1} \wedge K]}) \PE\left\|\nabla f\left(\mu_{t}\right)\right\|^{2} \\
\end{align}
\end{adjustwidth}
\begin{proof}
We first use \Cref{lem:boundpotential}:
\begin{align}
\PE\left[\Phi_{t+1}\right] \leq & \left(1-\kappa\right) \PE\left[\Phi_{t}\right]+3 \frac{s^2}{n} \eta^{2} \sum_{i=1}^{n} \PE[\frac{1}{{\alpha^i}^2} \left\|\widetilde{h}_{t+1}^i\right\|^{2}].
\end{align}, with $\alpha^i=\mathbf{P}(E^{i}_{t+1}>0)(E^i_{t+1} \wedge K)$ or $\alpha^i=\PE[E^i_{t+1} \wedge K]$.
Now we expand the quantity above using the inequality in \Cref{lem:boundsumlocalgrad}:
\begin{align}
\PE\left[\Phi_{t+1}\right] & \le \left(1-\kappa\right) \PE\left[\Phi_{t}\right]+ 3\frac{s^2}{n} \eta^{2}\left(\sigma^2\sum_i^n a^i + b (16L^2\PE[\Phi_t] + 8nB^2 \PE[\Vert \nabla f(\mu_t) \Vert^2]) + 8nG^2\right) \\
& \leq \left(1-\frac{1}{n} \left(\frac{s(n-s)}{2(n+1)(s+1)}\right)+48 \frac{s^2}{n} L^{2} b \eta^{2}\right) \PE\left[\Phi_{t}\right] + 3\frac{s^2}{n}(\sigma^2\sum_i^n a^i + 8nbG^2) \eta^{2} + 24 B^{2} s^2  b \eta^{2} \PE\left\|\nabla f\left(\mu_{t}\right)\right\|^{2}.
\end{align}
With \begin{equation}
    \begin{cases}
        a^i, b = \frac{1}{K^2\mathbf{P}(E^i_{t+1}>0)} + \frac{1}{\mathbf{P}(E^i_{t+1}>0)^2}\PE[\frac{\mathds{1}(E^i_{t+1}>0)}{E^i_{t+1} \wedge K}], \max_i(\frac{1}{\mathbf{P}(E^i_{t+1}>0)})\\
        a^i, b = \frac{1}{\PE[E^i_{t+1} \wedge K]} + \frac{\PE[(E^i_{+1} \wedge K)^2]}{K^2 \PE[E^i_{t+1} \wedge K]}, \max_i(\frac{\PE[(E^i_{t+1} \wedge K)^2]}{\PE[E^i_{t+1} \wedge K]}).
    \end{cases}
\end{equation}
To complete, we use $\eta \leq \frac{1}{20sLb}$:
\begin{align}
\PE\left[\Phi_{t+1}\right] \le \left(1-\frac{1}{5 n}\right) \PE\left[\Phi_{t}\right]+3s^2 \eta^{2}\left(\sigma^2\frac{1}{n}\sum_i^n a^i+ 8b G^{2}\right) + 24 B^{2} s^2 \eta^{2} b \PE\left\|\nabla f\left(\mu_{t}\right)\right\|^{2}.
\end{align}
\end{proof}
\end{lemma}
In the next lemma, we bound the cumulative sum of potential functions.
\begin{lemma}
\label{lem:boundsumexpectedpotential}
Let $T$ be a positive integer. Under the assumptions of \Cref{lem:boundexpectedpotential}, the following inequality holds:
\begin{align}
\sum_{t=0}^{T} \PE\left[\Phi_{t}\right] & \le 120n B^{2} s^2 \max_i(\frac{1}{\mathbf{P}(E^i_{t+1}>0)}) \eta^{2}  \sum_{t=0}^{T-1} \PE\left\|\nabla f\left(\mu_{t}\right)\right\|^{2}\\
&+ 15Ts^2\eta^2 \left(\sigma^2\sum_i^n(\frac{1}{K^2\mathbf{P}(E^i_{t+1}>0)} + \frac{1}{\mathbf{P}(E^i_{t+1}>0)^2}\PE[\frac{\mathds{1}(E^i_{t+1}>0)}{E^i_{t+1} \wedge K}])+8n\max_i(\frac{1}{\mathbf{P}(E^i_{t+1}>0)})G^2 \right).
\end{align}
\begin{align}
\sum_{t=0}^{T} \PE\left[\Phi_{t}\right] & \le 120n B^{2} s^2 \max_i(\frac{\PE[(E^i_{t+1} \wedge K)^2]}{\PE[E^i_{t+1} \wedge K]}) \eta^{2}  \sum_{t=0}^{T-1} \PE\left\|\nabla f\left(\mu_{t}\right)\right\|^{2}\\
&+ 15Ts^2\eta^2 \left(\sigma^2\sum_i^n(\frac{1}{\PE[E^i_{t+1} \wedge K]} + \frac{\PE[(E^i_{+1} \wedge K)^2]}{K^2 \PE[E^i_{t+1} \wedge K]})+8n\max_i(\frac{\PE[(E^i_{t+1} \wedge K)^2]}{\PE[E^i_{t+1} \wedge K]})G^2 \right).
\end{align}
\begin{proof}
From \Cref{lem:boundexpectedpotential}, we get that there exist $\alpha, \beta$ not depending on $t$ such that:
\begin{align}
\PE\left[\Phi_{t+1}\right] \leq\left(1-\frac{1}{5 n}\right) \PE\left[\Phi_{t}\right]+\alpha \PE\left\|\nabla f\left(\mu_{t}\right)\right\|^{2} + \beta.
\end{align}
Therefore:
\begin{align}
\sum_{t=0}^{T-1} \PE\left[\Phi_{t+1}\right]  \leq & \sum_{t=0}^{T-1}\left(\left(1-\frac{1}{5 n}\right) \PE\left[\Phi_{t}\right]+ \alpha \PE\left\|\nabla f\left(\mu_{t}\right)\right\|^{2} + \beta\right) \\
\leq & \left(1-\frac{1}{5 n}\right) \sum_{t=0}^{T-1} \PE\left[\Phi_{t}\right]+T\beta + \alpha \sum_{t=0}^{T-1} \PE\left\|\nabla f\left(\mu_{t}\right)\right\|^{2}.
\end{align}
Rearranging the terms in the sum we obtain the following:
\begin{align}
\left(1 - \frac{1}{5n}\right) \PE\left[\Phi_{0}\right] + \frac{1}{5n}\sum_{t=1}^{T-1} \PE\left[\Phi_{t}\right] + \PE\left[\Phi_{T}\right] \le T\beta + \alpha \sum_{t=0}^{T-1} \PE\left\|\nabla f\left(\mu_{t}\right)\right\|^{2}.
\end{align}
From this inequality, we get:
\begin{align}
\sum_{t=0}^{T} \PE\left[\Phi_{t}\right] \leq & 5 n\left(T\beta + \alpha \sum_{t=0}^{T-1} \PE\left\|\nabla f\left(\mu_{t}\right)\right\|^{2}\right).
\end{align}
Expanding on the values of $\alpha, \beta$ obtained by \Cref{lem:boundexpectedpotential}, we get the desired result:
\begin{align}
\sum_{t=0}^{T} \PE\left[\Phi_{t}\right] & \le 5nT\frac{3s^2}{n}\eta^2 \left(\sigma^2\sum_i^na^i+8nbG^2 \right) + 120n B^{2} s^2 b \eta^{2}  \sum_{t=0}^{T-1} \PE\left\|\nabla f\left(\mu_{t}\right)\right\|^{2}.
\end{align}
\end{proof}
\end{lemma}

\subsection{Bound the change in the average model}
The next lemma upper bounds the expected change in the average model $\mu_t$.
\begin{lemma}
\label{lem:boundmeannode}
For any time step $t \ge 0$,
\begin{equation}
\PE\left\|\mu_{t+1}-\mu_{t}\right\|^{2} \leq \frac{s^{2} \eta^{2}}{n(n+1)^{2}} \sum_{i=1}^n \PE\left\|\check{h}_{t+1}^i\right\|^{2}.
\end{equation}
\begin{proof}
Recall that
\begin{align}
\mu_{t+1}-\mu_t = - \frac{\eta}{n+1}\sum_{i \in \Set_{t+1}} \check{h}_{t+1}^i.
\end{align}
Therefore we may compute an upper bound:
\begin{align}
\|\mu_{t+1}-\mu_t\|^2 &= \frac{\eta^2}{(n+1)^2}\left\|\sum_{i \in \Set_{t+1}} \check{h}_{t+1}^i\right\|^2 \\
& \le \frac{s\eta^2}{(n+1)^2}\sum_{i \in \Set_{t+1}} \left\|\check{h}_{t+1}^i\right\|^2
\end{align}
We may then apply \Cref{lem:expectation-sum} to get the desired result:
\begin{equation}
\PE\left\|\mu_{t+1}-\mu_{t}\right\|^{2} \leq \frac{s^{2} \eta^{2}}{n(n+1)^{2}} \sum_{i=1}^n \PE\left\|\check{h}_{t+1}^i\right\|^{2}.
\end{equation}
\end{proof}
\end{lemma}
We now give another upper bound on how the average model $\mu_t$ changes at time step $t$.
\begin{lemma}
\label{lem:boundmeannodetruegrad}
Under the assumptions of \Cref{lem:boundmeannode,lem:boundsumlocalgrad}, and for any step $t$:
\begin{adjustwidth}{-30pt}{}
\begin{align}
        \PE \| \mu_{t+1}- \mu_{t}\|^{2} &\leq \frac{s^{2} \eta^{2}\sigma^2}{n(n+1)^{2}} \sum_i^n(\frac{1}{K^2\mathbf{P}(E^i_{t+1}>0)} + \frac{1}{\mathbf{P}(E^i_{t+1}>0)^2}\PE[\frac{\mathds{1}(E^i_{t+1}>0)}{E^i_{t+1} \wedge K}]) + \frac{16L^2s^{2} \eta^{2}}{n(n+1)^{2}} \PE[\Phi_t] \max_i(\frac{1}{\mathbf{P}(E^i_{t+1}>0)})\\
        &+ \frac{8s^{2}B^2 \eta^{2}}{(n+1)^{2}}\max_i(\frac{1}{\mathbf{P}(E^i_{t+1}>0)}) \PE[\Vert \nabla f (\mu_t) \Vert^2] + \frac{8s^{2} G^2\eta^{2}}{(n+1)^{2}}\max_i(\frac{1}{\mathbf{P}(E^i_{t+1}>0)})\\
        \PE \| \mu_{t+1}- \mu_{t}\|^{2} &\leq \frac{s^{2} \eta^{2}\sigma^2}{n(n+1)^{2}} \sum_i^n(\frac{1}{\PE[E^i_{t+1} \wedge K]} + \frac{\PE[(E^i_{+1} \wedge K)^2]}{K^2 \PE[E^i_{t+1} \wedge K]}) + \frac{16L^2s^{2} \eta^{2}}{n(n+1)^{2}} \PE[\Phi_t] \max_i(\frac{\PE[(E^i_{t+1} \wedge K)^2]}{\PE[E^i_{t+1} \wedge K]}) \\
        &+ \frac{8s^{2}B^2 \eta^{2}}{(n+1)^{2}} \max_i(\frac{\PE[(E^i_{t+1} \wedge K)^2]}{\PE[E^i_{t+1} \wedge K]}) \PE[\Vert \nabla f (\mu_t) \Vert^2] + \frac{8s^{2} G^2\eta^{2}}{(n+1)^{2}}\max_i(\frac{\PE[(E^i_{t+1} \wedge K)^2]}{\PE[E^i_{t+1} \wedge K]}) .
\end{align}
\end{adjustwidth}
\begin{proof}
For this proof, we will combine the inequality obtained from \Cref{lem:boundmeannode} to the one from \Cref{lem:boundsumlocalgrad}. This will be enough to obtain the desired result.
\begin{align}
&\PE \| \mu_{t+1}- \mu_{t}\|^{2} \le  \sum_{i=1}^n \frac{s^{2} \eta^{2}}{n(n+1)^{2}} \PE \| \check{h}_{t+1}^i \|^{2}.
\end{align}
Simplifying the above quantity, we get the desired inequality:
\begin{adjustwidth}{-30pt}{}
\begin{align}
        \PE \| \mu_{t+1}- \mu_{t}\|^{2} &\leq \frac{s^{2} \eta^{2}\sigma^2}{n(n+1)^{2}} \sum_i^n(\frac{1}{K^2\mathbf{P}(E^i_{t+1}>0)} + \frac{1}{\mathbf{P}(E^i_{t+1}>0)^2}\PE[\frac{\mathds{1}(E^i_{t+1}>0)}{E^i_{t+1} \wedge K}]) + \frac{16L^2s^{2} \eta^{2}}{n(n+1)^{2}} \PE[\Phi_t] \max_i(\frac{1}{\mathbf{P}(E^i_{t+1}>0)})\\
        &+ \frac{8s^{2}B^2 \eta^{2}}{(n+1)^{2}}\max_i(\frac{1}{\mathbf{P}(E^i_{t+1}>0)}) \PE[\Vert \nabla f (\mu_t) \Vert^2] + \frac{8s^{2} G^2\eta^{2}}{(n+1)^{2}}\max_i(\frac{1}{\mathbf{P}(E^i_{t+1}>0)}) G^2\\
        \PE \| \mu_{t+1}- \mu_{t}\|^{2} &\leq \frac{s^{2} \eta^{2}\sigma^2}{n(n+1)^{2}} \sum_i^n(\frac{1}{\PE[E^i_{t+1} \wedge K]} + \frac{\PE[(E^i_{+1} \wedge K)^2]}{K^2 \PE[E^i_{t+1} \wedge K]}) + \frac{16L^2s^{2} \eta^{2}}{n(n+1)^{2}} \PE[\Phi_t] \max_i(\frac{\PE[(E^i_{t+1} \wedge K)^2]}{\PE[E^i_{t+1} \wedge K]}) \\
        &+ \frac{8s^{2}B^2 \eta^{2}}{(n+1)^{2}} \max_i(\frac{\PE[(E^i_{t+1} \wedge K)^2]}{\PE[E^i_{t+1} \wedge K]}) \PE[\Vert \nabla f (\mu_t) \Vert^2] + \frac{8s^{2} G^2\eta^{2}}{(n+1)^{2}}\max_i(\frac{\PE[(E^i_{t+1} \wedge K)^2]}{\PE[E^i_{t+1} \wedge K]}) G^2 .
        \end{align}
        \end{adjustwidth}
\end{proof}
\end{lemma}

\subsection{Convergence result}

In this section, we use the lemmas proved so far to demonstrate \Cref{theorem:cvgquant}. Following the proof, we establish the learning rate $\eta$ that results in the best overall rate of convergence.

\begin{proof}
Using $L$-smoothness, we have:
\begin{equation}
\label{eq:smoothness}
f\left(\mu_{t+1}\right) \leq f\left(\mu_{t}\right)+\left\langle\nabla f\left(\mu_{t}\right), \mu_{t+1}-\mu_{t}\right\rangle+\frac{L}{2} \left\|\mu_{t+1}-\mu_{t}\right\|^{2}.
\end{equation}

First we look at the term $\left\langle\nabla f\left(\mu_{t}\right), \mu_{t+1}-\mu_{t}\right\rangle$. Recall that:
\begin{align}
\mu_{t+1}-\mu_t = - \frac{\eta}{n+1}\sum_{i \in \Set_{t+1}} \check{h}_{t+1}^i
\end{align}
by \Cref{lem:expectation-sum}, we have:
\begin{align}
    \PE_t[\mu_{t+1}-\mu_t] = -\frac{s\eta}{n(n+1)}\sum_{i=1}^n \check{h}_{t+1}^i
\end{align}
and subsequently
$$
\PE_{t}\left\langle\nabla f\left(\mu_{t}\right), \mu_{t+1}-\mu_{t}\right\rangle=\sum_{i=1}^{n} \frac{s \eta}{n(n+1)} \PE_{t}\left\langle\nabla f\left(\mu_{t}\right),-\check{h}_{t+1}^i\right\rangle .
$$
Hence, we can rewrite \eqref{eq:smoothness} as:
\begin{equation}
\PE_{t}\left[f\left(\mu_{t+1}\right)\right] \leq f\left(\mu_{t}\right)+\sum_{i=1}^{n} \frac{s \eta}{n(n+1)} \PE_{t}\left\langle\nabla f\left(\mu_{t}\right),-\check{h}_{t+1}^i\right\rangle+\frac{L}{2} \PE_{t}\left\|\mu_{t+1}-\mu_{t}\right\|^{2}.
\end{equation}
Next, we remove the conditioning with the tower law of expectation:
\begin{align}
\PE\left[f\left(\mu_{t+1}\right)\right] \leq \PE\left[f\left(\mu_{t}\right)\right] +\sum_{i=1}^{n} \frac{s \eta}{n(n+1)} \PE\left\langle\nabla f\left(\mu_{t}\right),-\check{h}_{t+1}^i\right\rangle +\frac{L}{2} \PE\left\|\mu_{t+1}-\mu_{t}\right\|^{2} .
\end{align}
We now define some notation to simplify the computations. By \Cref{lem:boundscalarproduct}, there exist $a_1, a_2, a_3$ not depending on $t$ such that
\begin{align}
\sum_{i=1}^{n} \PE\left\langle\nabla f\left(\mu_{t}\right),- \check{h}_{t+1}^i\right\rangle & \leq a_1 \PE\left[\Phi_{t}\right]  + a_2 \PE\left\|\nabla f\left(\mu_{t}\right)\right\|^{2} + a_3.
\end{align}
Similarly, by \Cref{lem:boundmeannodetruegrad}, there exist $b_1, b_2, b_3$ not depending on $t$ such that:
\begin{align}
\PE\left\|\mu_{t+1}-\mu_{t}\right\|^{2} \leq b_1 \PE\left[\Phi_{t}\right] + b_2 \PE\left\|\nabla f\left(\mu_{t}\right)\right\|^{2} + b_3.
\end{align}
Defining $c_i = a_i \frac{s \eta}{n(n+1)} + b_i\frac{L}{2}$, we have
\begin{align}
\PE\left[f\left(\mu_{t+1}\right)\right]-\PE\left[f\left(\mu_{t}\right)\right] \leq & c_1 \PE\left[\Phi_{t}\right] + c_2 \PE\left\|\nabla f\left(\mu_{t}\right)\right\|^{2} + c_3.
\end{align}
Summing the above inequality for $t=0, 1, ..., T-1$ we get that:
\begin{align}
\PE\left[f\left(\mu_{T}\right)\right]- f(\mu_0) \le c_1 \sum_{t=0}^{T-1} \PE\left[\Phi_{t}\right] + c_2 \sum_{t=0}^{T-1} \PE\left\|\nabla f\left(\mu_{t}\right)\right\|^{2} + c_3T.
\end{align}
By \Cref{lem:boundsumexpectedpotential} there exist $d_1, d_2$ independent of $T$ such that:
\begin{align}
\sum_{t=0}^{T} \PE\left[\Phi_{t}\right] \leq  d_1  \sum_{t=0}^{T-1} \PE\left\|\nabla f\left(\mu_{t}\right)\right\|^{2} + Td_2.
\end{align}
We then get:
\begin{align}
\PE\left[f\left(\mu_{T}\right)\right]- f(\mu_0) \le (c_1d_1 + c_2) \sum_{t=0}^{T-1} \PE\left\|\nabla f\left(\mu_{t}\right)\right\|^{2} + T(c_1d_2 + c_3).
\end{align}
We now assume that $c_1d_1 + c_2 < 0$. Later in the proof, we will show that this is true for small enough $\eta$. Using the fact that $f(\mu_T) \ge f_*$ and rearranging the terms, we get:
\begin{align}
\frac{1}{T}\sum_{t=0}^{T-1} \PE\left\|\nabla f\left(\mu_{t}\right)\right\|^{2} \le \frac{f(\mu_0) - f_* }{T(-c_1d_1 - c_2)} + \frac{c_1d_2 + c_3}{-c_1d_1 - c_2}.
\end{align}
Of course, now we expand each of these terms. Refer to \Cref{lem:boundscalarproduct}, \Cref{lem:boundmeannodetruegrad}, and \Cref{lem:boundsumexpectedpotential} for the specific values of the defined quantities $a_i$, $b_i$, and $d_i$. We have:
\begin{align}
c_1 & = \frac{s\eta}{n(n+1)} a_1 + \frac{L}{2} b_1 \\
& = 20L^2 \frac{s\eta}{n(n+1)} + \frac{16s^2\eta^2L^2b}{n(n+1)^2} \frac{L}{2} \\
& = \frac{4L^2s\eta}{n(n+1)^2}\left(5(n+1) + 2s\eta Lb \right).
\end{align}
We recall here the definition from \Cref{lem:boundexpectedpotential}
\begin{equation}
    \begin{cases}
        a^i, b = \frac{1}{K^2\mathbf{P}(E^i_{t+1}>0)} + \frac{1}{\mathbf{P}(E^i_{t+1}>0)^2}\PE[\frac{\mathds{1}(E^i_{t+1}>0)}{E^i_{t+1} \wedge K}], \max_i(\frac{1}{\mathbf{P}(E^i_{t+1}>0)})\\
        a^i, b = \frac{1}{\PE[E^i_{t+1} \wedge K]} + \frac{\PE[(E^i_{+1} \wedge K)^2]}{K^2 \PE[E^i_{t+1} \wedge K]}, \max_i(\frac{\PE[(E^i_{t+1} \wedge K)^2]}{\PE[E^i_{t+1} \wedge K]}).
    \end{cases}
\end{equation}
By using $\eta \le \frac{n+1}{2sLb}$ we get:
\begin{align}
c_1 \le \frac{24L^2s\eta}{n(n+1)}.
\end{align}
Therefore:
\begin{align}
c_1d_1 & \le \frac{24L^2s\eta}{n(n+1)} 120nB^2s^2b\eta^2 \\
& \le \frac{2880L^2s^3\eta^3B^2b}{n+1}.
\end{align}
Moreover:
\begin{align}
c_2 & = \frac{s\eta}{n(n+1)} a_2 + \frac{L}{2} b_2 \\
& = \frac{s\eta}{n(n+1)}\left( n\left(16L^2\eta^2K^2B^2 -\frac{3}{4}\right) \right) + \frac{L}{2} \left( \frac{8s^2\eta^2bB^2}{(n+1)^2}\right) \\
& = \frac{s\eta}{(n+1)}\left(16L^2\eta^2K^2B^2  - \frac{3}{4} \right) + \frac{4Ls^2\eta^2bB^2}{(n+1)^2}.
\end{align}
By using $\eta \le \frac{1}{20B^2bKLs}$ we get:
\begin{align}
c_1d_1 \le \frac{2880}{8000(n+1)LB^4K^3b^2}
\end{align}
\begin{align}
c_2 & \le \frac{s\eta}{n+1}\left( \frac{16}{400s^2b^2B^2} - \frac{3}{4} \right) + \frac{4}{400(n+1)^2 LbB^2K^2} \\
& \le \frac{-s\eta}{2(n+1)} - c_1d_1 .
\end{align}
Therefore we conclude that $-c_2 - c_1d_1 \ge \frac{s\eta}{2(n+1)}$, which is greater than $0$. Now we compute:
\begin{align}
c_1d_2 & \le \frac{24L^2s\eta}{n(n+1)}\left(15s^2\eta^2 \left(\sigma^2 \sum_i^n a^i + 8nb G^2 \right) \right) \\
& \le \frac{360L^2s^3\eta^3}{(n+1)} \left(\sigma^2 \frac{1}{n} \sum_i^n a^i + 8bG^2\right).
\end{align}
And additionally:
\begin{align}
c_3 & = \frac{s\eta}{n(n+1)}a_3 + \frac{L}{2}b_3 \\
& = \frac{s\eta}{n(n+1)} \left(4nL^2\eta^2K^2(\sigma^2+4G^2)\right) + \frac{L}{2}\left( \frac{s^2\eta^2\sigma^2}{n(n+1)^2} \sum_i^n a^i + \frac{8s^2G^2\eta^2}{(n+1)^2}b \right).
\end{align}
Thus
\begin{align}
c_3 + c_1d_2 \le & \left( \frac{4L^2\eta^2K^2s\eta}{(n+1)} + \frac{Ls^2\eta^2}{2n(n+1)^2} \sum_i^n a^i + \frac{360L^2s^3\eta^3}{n(n+1)} \sum_i^n a^i \right) \sigma^2 \\
& + \left( \frac{16L^2\eta^2K^2s\eta}{(n+1)} + \frac{4Ls^2\eta^2}{(n+1)^2}b + \frac{2880bL^2s^3\eta^3}{(n+1)}\right) G^2.
\end{align}
And therefore:
\begin{align}
\frac{c_3 + c_1d_2}{-c_1d_1-c_2} \le & \left( 8L^2\eta^2K^2 + \frac{Ls\eta}{n(n+1)} \sum_i^n a^i + \frac{720L^2s^2\eta^2}{n} \sum_i^n a^i \right) \sigma^2 \\
& + \left( 32L^2\eta^2K^2 + \frac{8Ls\eta}{(n+1)}b + 5600bL^2s^2\eta^2\right) G^2.
\end{align}
Finally:
\begin{align}
\frac{1}{T}\sum_{t=0}^{T-1} \PE\left\|\nabla f\left(\mu_{t}\right)\right\|^{2} \le & 2(n+1)\frac{f(\mu_0) - f_*}{Ts\eta} \\
& + \left( 8L^2\eta^2K^2 + \frac{Ls\eta}{n(n+1)} \sum_i^n a^i + \frac{720L^2s^2\eta^2}{n} \sum_i^n a^i \right) \sigma^2 \\
& + \left( 32L^2\eta^2K^2 + \frac{8Ls\eta}{(n+1)}b + 5600bL^2s^2\eta^2\right) G^2.
\end{align}
\end{proof}
In particular for stochastic reweighting:
\begin{adjustwidth}{-70pt}{}
\begin{align}
\frac{1}{T}&\sum_{t=0}^{T-1} \PE\left\|\nabla f\left(\mu_{t}\right)\right\|^{2} \le 2(n+1)\frac{f(\mu_0) - f_*}{Ts\eta} \\
& + \left( 8L^2\eta^2K^2 + \frac{Ls\eta}{n(n+1)} \sum_i^n (\frac{1}{K^2\mathbf{P}(E^i_{t+1}>0)} + \frac{1}{\mathbf{P}(E^i_{t+1}>0)^2}\PE[\frac{\mathds{1}(E^i_{t+1}>0)}{E^i_{t+1} \wedge K}]) \right)\sigma^2\\
& + \left( \frac{720L^2s^2\eta^2}{n} \sum_i^n (\frac{1}{K^2\mathbf{P}(E^i_{t+1}>0)} + \frac{1}{\mathbf{P}(E^i_{t+1}>0)^2}\PE[\frac{\mathds{1}(E^i_{t+1}>0)}{E^i_{t+1} \wedge K}]) \right) \sigma^2 \\
& + \left( 32L^2\eta^2K^2 + \frac{8Ls\eta}{(n+1)}\max_i(\frac{1}{\mathbf{P}(E^i_{t+1}>0)}) + 5600L^2s^2\eta^2 \max_i(\frac{1}{\mathbf{P}(E^i_{t+1}>0)})\right) G^2.
\end{align}
\end{adjustwidth}
And particular for expectation reweighting:
\begin{adjustwidth}{-70pt}{}
\begin{align}
\frac{1}{T}&\sum_{t=0}^{T-1} \PE\left\|\nabla f\left(\mu_{t}\right)\right\|^{2} \le  2(n+1)\frac{f(\mu_0) - f_*}{Ts\eta} \\
& + \left( 8L^2\eta^2K^2 + \frac{Ls\eta}{n(n+1)} \sum_i^n (\frac{1}{\PE[E^i_{t+1} \wedge K]} + \frac{\PE[(E^i_{+1} \wedge K)^2]}{K^2 \PE[E^i_{t+1} \wedge K]}) + \frac{720L^2s^2\eta^2}{n} \sum_i^n (\frac{1}{\PE[E^i_{t+1} \wedge K]} + \frac{\PE[(E^i_{+1} \wedge K)^2]}{K^2 \PE[E^i_{t+1} \wedge K]}) \right) \sigma^2 \\
& + \left( 32L^2\eta^2K^2 + \frac{8Ls\eta}{(n+1)}\max_i(\frac{\PE[(E^i_{t+1} \wedge K)^2]}{\PE[E^i_{t+1} \wedge K]}) + 5600L^2s^2\eta^2 \max_i(\frac{\PE[(E^i_{t+1} \wedge K)^2]}{\PE[E^i_{t+1} \wedge K]}) \right) G^2.
\end{align}
\end{adjustwidth}

\clearpage
\newpage
\section{Detailed simulation environment}
\begin{figure}[H]
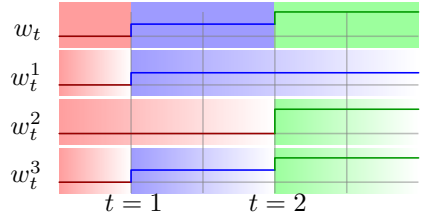

    \centering
%

\definecolor{bgblue}{rgb}{0.61961,0.60784,1.0}%
\definecolor{bgred}{rgb}{1,0.61569,0.61569}%
\definecolor{bggreen}{rgb}{0.61,1,0.61569}%
\definecolor{fgblue}{rgb}{0,0,0.6}%
\definecolor{fgred}{rgb}{0.6,0,0}%
\definecolor{fggreen}{rgb}{0, 0.6,0}%
\definecolor{fgblack}{rgb}{1, 1,1}%
\begin{tikztimingtable}[
    timing/slope=0,         
    timing/coldist=2pt,     
    xscale=3.25,yscale=1.1, 
    semithick               
  ]
  $\param_t$     & [fgred]L; N(A); ZZ; [fggreen]HH       \\
  $\param^1_t$    & [fgred]L; N(B); ZZZZ    \\
  $\param^2_t$    &    [fgred]LLL; [fggreen]HH  \\
  $\param^3_t$     &  [fgred]L;ZZ;[fggreen]HH      \\
\extracode
 \makeatletter
 \begin{pgfonlayer}{background}
  \shade [right color=white,left color=bgred]
     (1,-6.45) rectangle (0,-4.6);
  \shade [right color=white,left color=bgblue]
     (3,-6.45) rectangle (1,-4.6);
  \shade [right color=white,left color=bggreen]
     (5,-6.45) rectangle (3,-4.6);
  \shade [right color=white,left color=bgred]
     (3,-4.45) rectangle (0,-2.6);
  \shade [right color=white,left color=bggreen]
     (5,-4.45) rectangle (3,-2.6);
  \shade [right color=white,left color=bgred]
     (1,-2.45) rectangle (0,-0.6);
  \shade [right color=white,left color=bgblue]
     (5,-2.45) rectangle (1,-0.6);
  \shade [right color=bgred,left color=bgred]
     (1,-0.45) rectangle (0,1.4);
  \shade [right color=bgblue,left color=bgblue]
     (3,-0.45) rectangle (1,1.4);
  \shade [right color=bggreen,left color=bggreen]
     (5,-0.45) rectangle (3,1.4);
  \begin{scope}[gray,semitransparent,semithick]
    \horlines{}
    \foreach \x in {1,...,4}
      \draw (\x,1) -- (\x,-6.45);
  \end{scope}
  \node [anchor=south east,inner sep=0pt]
    at (1.45,-7.2) {$t=1$};
  \node [anchor=south east,inner sep=0pt]
    at (3.45,-7.2) {$t=2$};
 \end{pgfonlayer}
\end{tikztimingtable}%

    \caption{Example of asynchronous updates with $n=3$ nodes and selection size $s=2$. At $t=0$, all clients are initialized withe the same value. At time $t=1$, clients $\{1,3\}$ are selected, and at time $t=2$, clients $\{2,3\}$. At time $t=2$, client $2$ is reporting updates computed on outdated parameter.}
    \label{fig:modeltime}
\end{figure}

From \Cref{algo:flau}, one must note that local weights are reset with the central model only when being contacted by the central server. Hence initially we have $\param^i_0 = \param_0$, but at time $t$ we may have $\param^i_t \neq \param_t$, see \Cref{fig:modeltime}.
\label{sec:app:exp}
\subsection{Implementation of concurrent works}
\label{sec:app:concu_works} In \Cref{sec:experiments} we have simulated experiments and run the code for the concurrent approaches FedAvg, QuAFL, and FedBuff. FedAvg is a standard synchronous method. At the beginning of each round, the central node $s$ selects clients uniformly at random and broadcast its current model. Each of these clients take the central server value and then performs exactly $K$ local steps, and then sends the resulting model progress back to the server. The server then computes the average of the $s$ received models and updates its model. In this synchronous structure, the server must wait in each round for the slowest client to complete its update.
QuAFL is an asynchronous method that randomly selects $s$ clients at each server invocation. The server then replaces its model with a convex combination of the received models and its current model. Also, the $s$ receiving clients replace their local model with a convex combination between their current model and the model of the receiving server.
In FedBuff, clients compute local training asynchronously as well, with the help of a buffer. Once the buffer is filled with $Z$ different client updates, the server averages the buffer updates and performs a gradient step on the computed average. Then the buffer is reset to zero and the available clients get the server model as a new starting point.
\subsection{Discussion on simulated runtime}
\label{sec:app:simuruntime}
We based our simulations mainly on the code developed by \cite{nguyen2022federated}: we assume a server and $n$ clients, each of which initially has a model copy. We assume that, at each time step $t$ (for the central server), a batch of $s$ clients are sampled at random without replacement. For the client $i$, the inter-arrival time of two successive requests are therefore independent and distributed according to a geometric distributions of parameter $s/n$. The time elapsed from the last renewal is distributed according to the stationary distribution of the age process (assuming that the renewal is stationary), which is also distributed according to a geometric random variable with the same parameter $s/n$.  We assume that the clients have different computational speeds. For this purpose $E_t^i$ is distributed according to a geometrical distribution of parameter $\lambda^i$: $E^i_t \sim$ $\Geom(\lambda^i)$. The parameter $\lambda^i$ is $1 / 2$ for fast clients and $1 / 16$ for slow clients; the expected running time $\PE[E^i_t]$ is 2 and 16, respectively. The training dataset is distributed among the clients so that each of them has access to an equal portion of the training data (whether it is IID or non-IIID). We track the performance of each algorithm by evaluating the server's model against an unseen validation dataset. We measure the loss and accuracy of the model in terms of simulation time, server steps, and total local steps taken by clients.

To adequately capture the time spent on the server side for computations and orchestration of centralized learning, two quantities are implemented: the server waiting time (the time the server waits between two consecutive calls ) and the server interaction time (the time the server takes to send and receive the required data). In all experiments, they are set to $4$ and $3$, respectively.

For each global step, the FedAvg runtime is the sum of the server interaction time (see above) and the local step runtime of the slowest selected client times the number of maximum epochs $K$ (we wait until all clients have computed all their local epochs in this synchronous setting). For QuAFL and \Algo, the duration of a global step is simply the sum of the server interaction time and the server waiting time. For FedBuff, the runtime is the sum of the server interaction time and the time spent feeding the buffer of size $Z$. The waiting time for feeding the buffer depends on the respective local runtimes of the slow and fast clients, as well as on the ratio between slow and fast clients: in the code, we reset a counter at the beginning of each global step and read the runtime when the $Z^{th}$ local update arrives.

\subsection{Detailed results}
\label{sec:app:detailedresults}
Below we provide further insight into the experiments described in \Cref{sec:experiments}. We present figures for loss, variance ($\sum_{i=1}^n \Vert \param^i_t-\param_t \Vert^2$), but also for accuracy (evaluated on the held-out test set on the server side) in terms of time, but also in terms of total local steps and total server steps.

We find that \Algo\ and other asynchronous methods, when time rather than the number of server steps \Algo\ - and more generally asynchronous methods - can achieve significant speedups on these metrics compared to FedAvg. This is due to asynchronous communication allowing rounds to complete faster without always having to wait for slower nodes to complete their local computations. Although this behaviour is simulated, we believe it reflects the practical potential of \Algo.
\begin{figure}[H]
    \centering
    \includegraphics[scale=0.5]{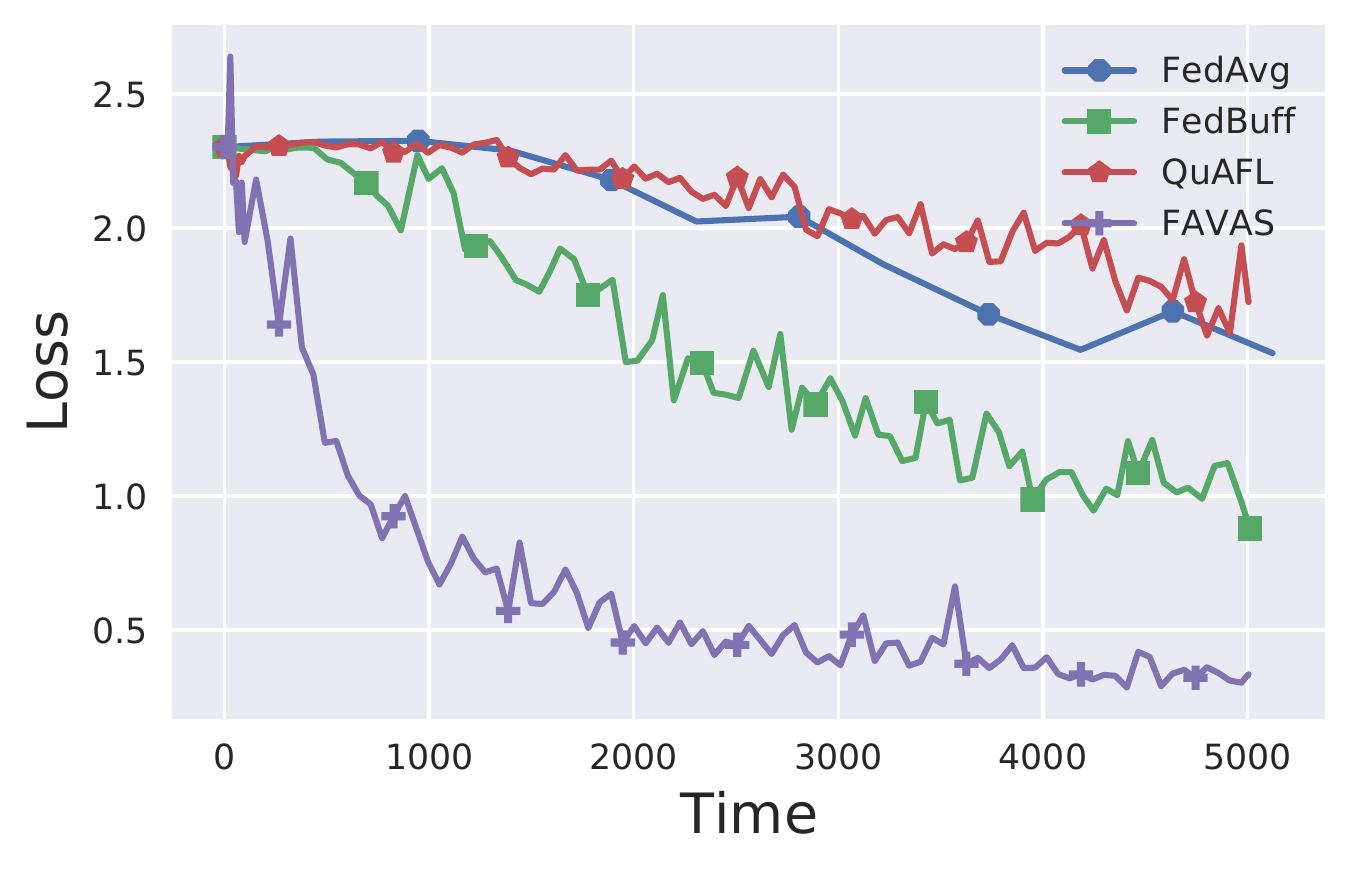}
    \includegraphics[scale=0.5]{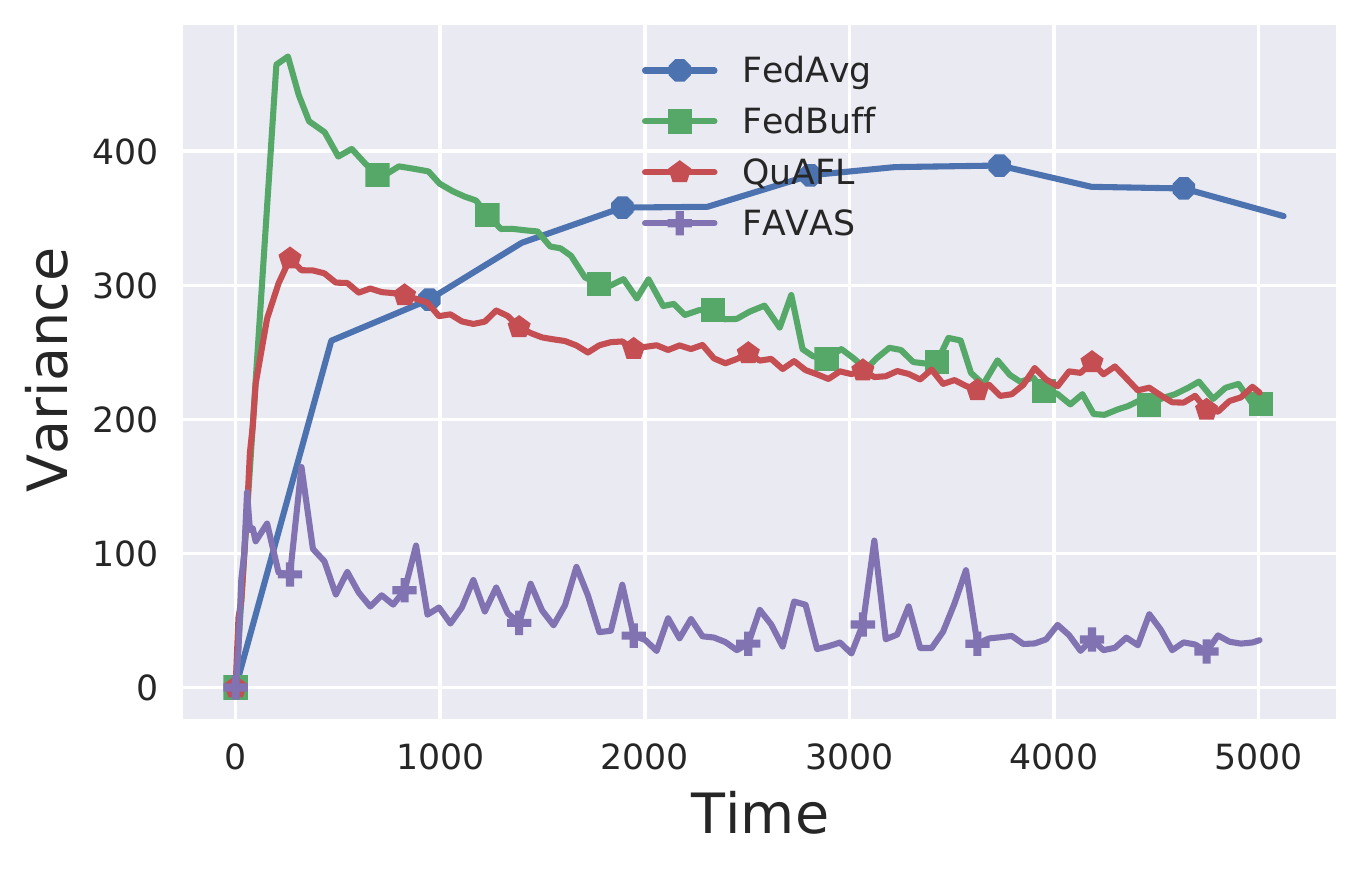}
    \includegraphics[scale=0.5]{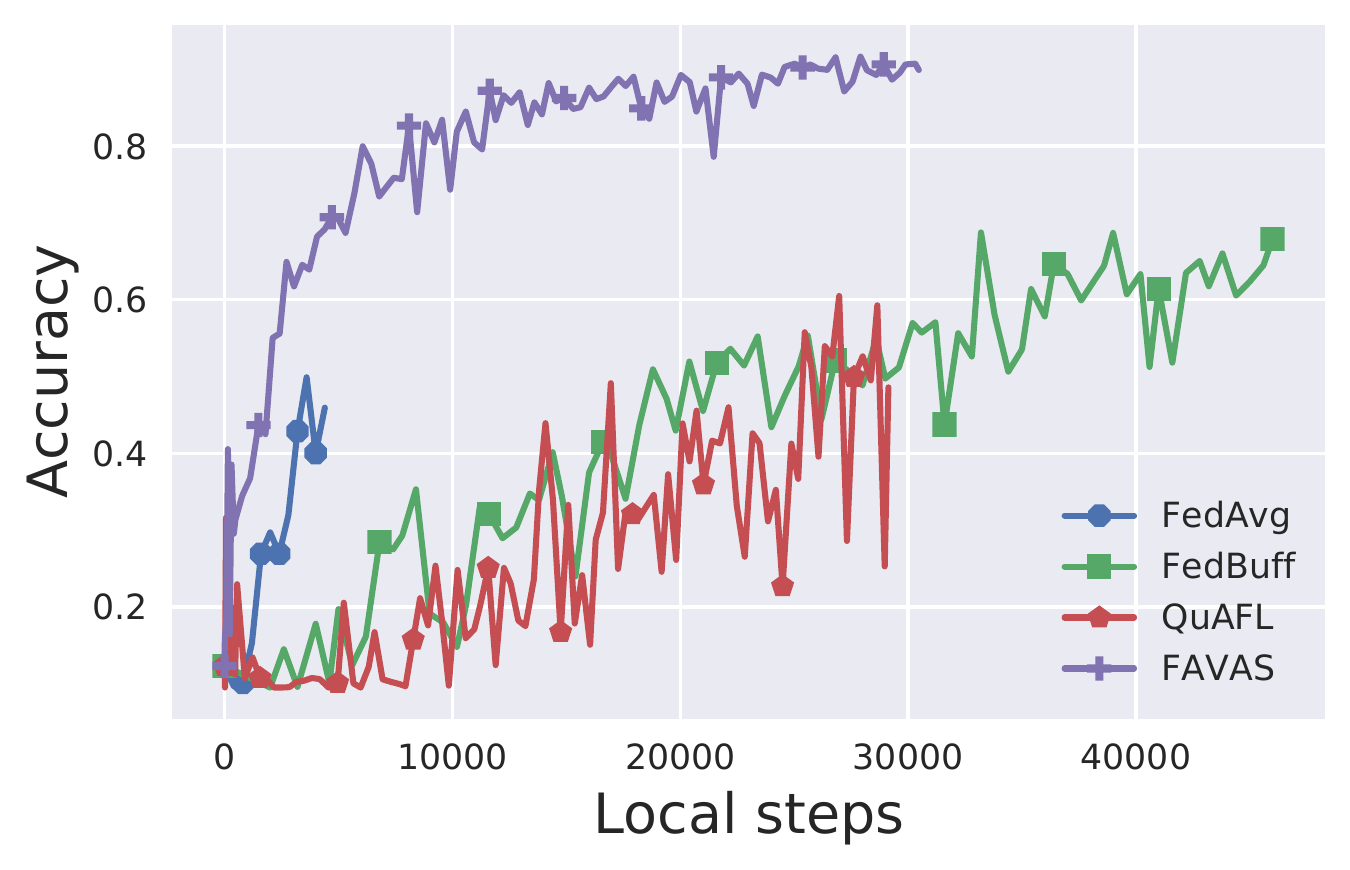}
    \includegraphics[scale=0.5]{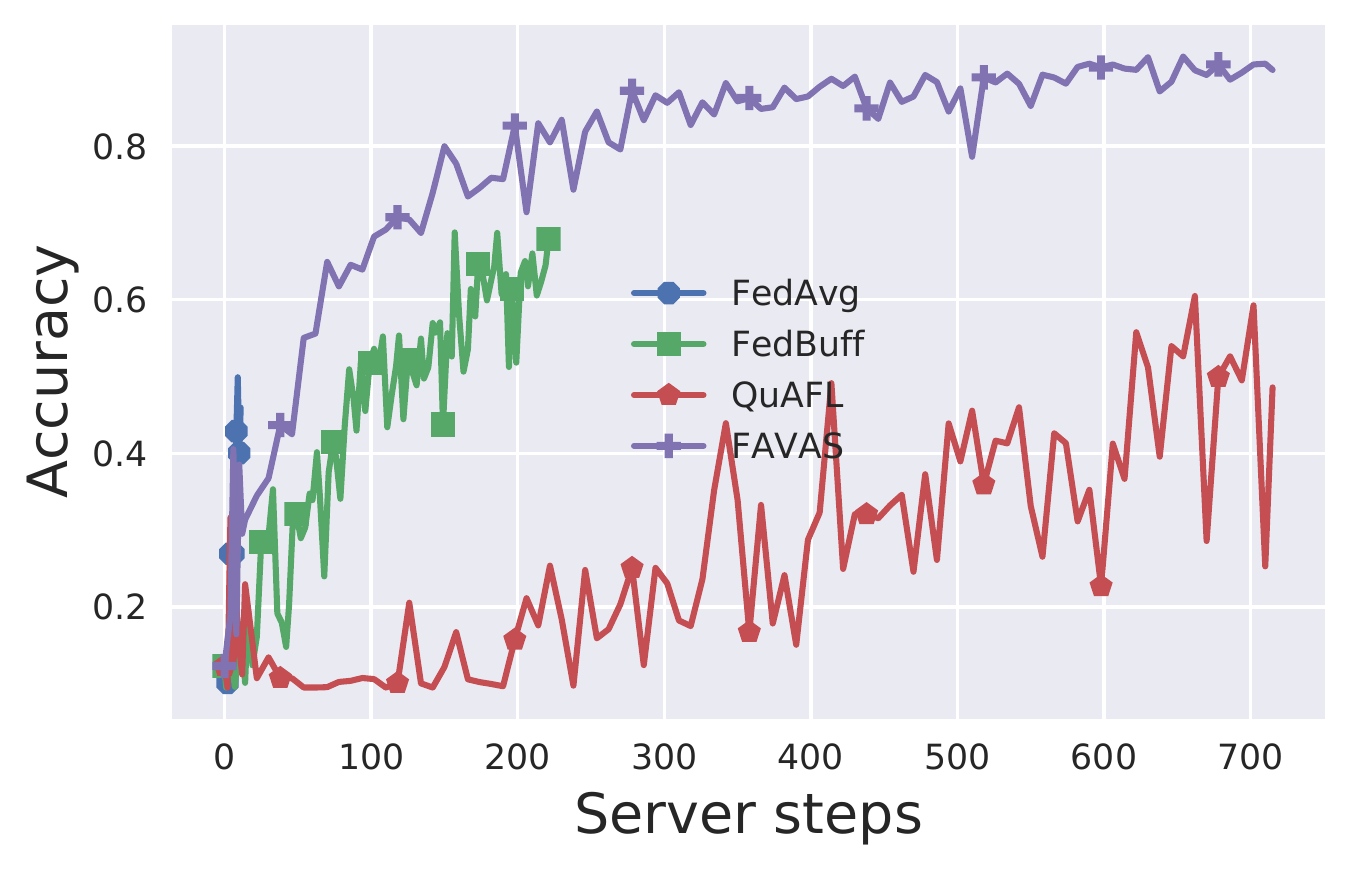}
    \includegraphics[scale=0.5]{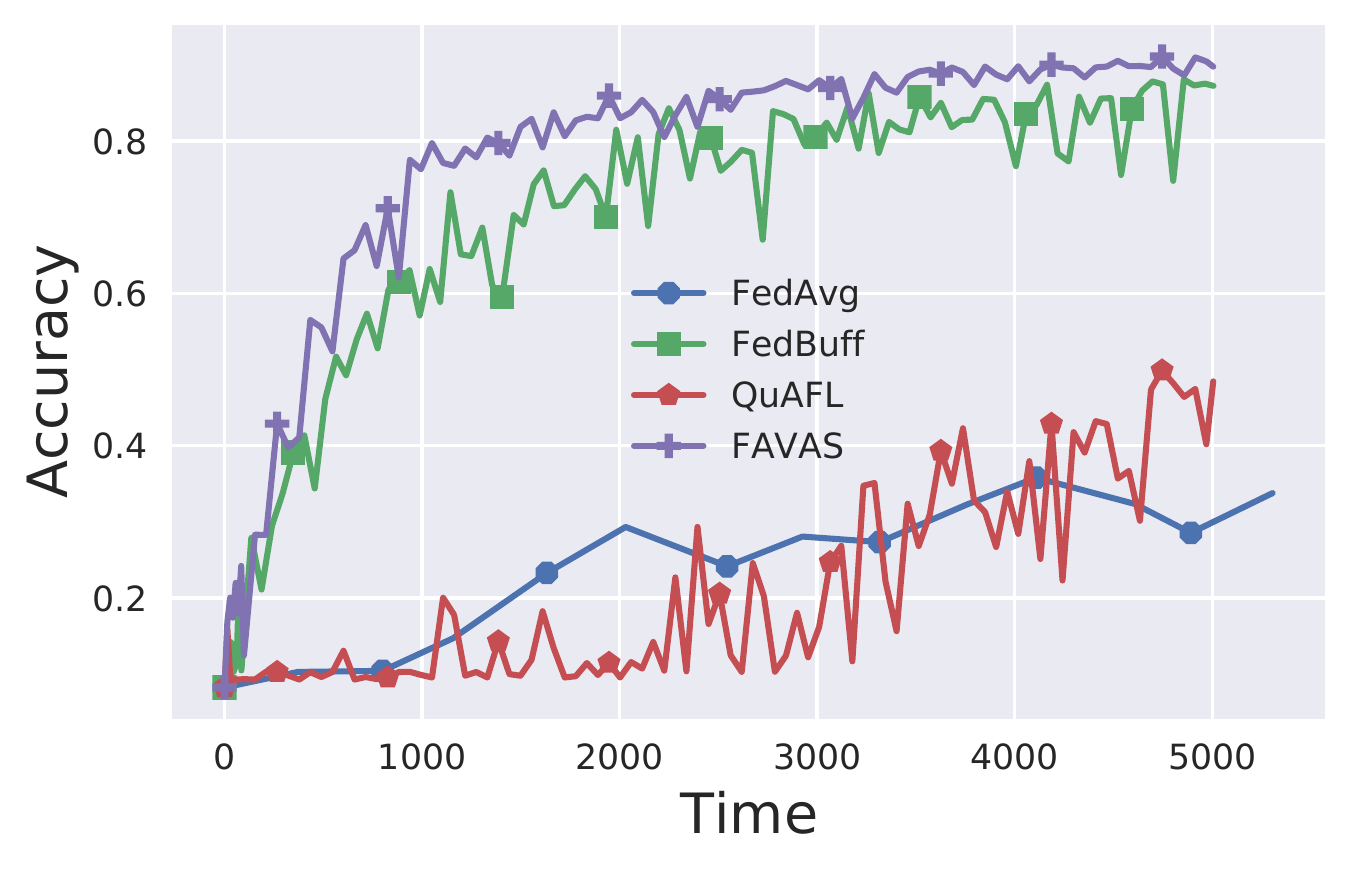}
    \includegraphics[scale=0.5]{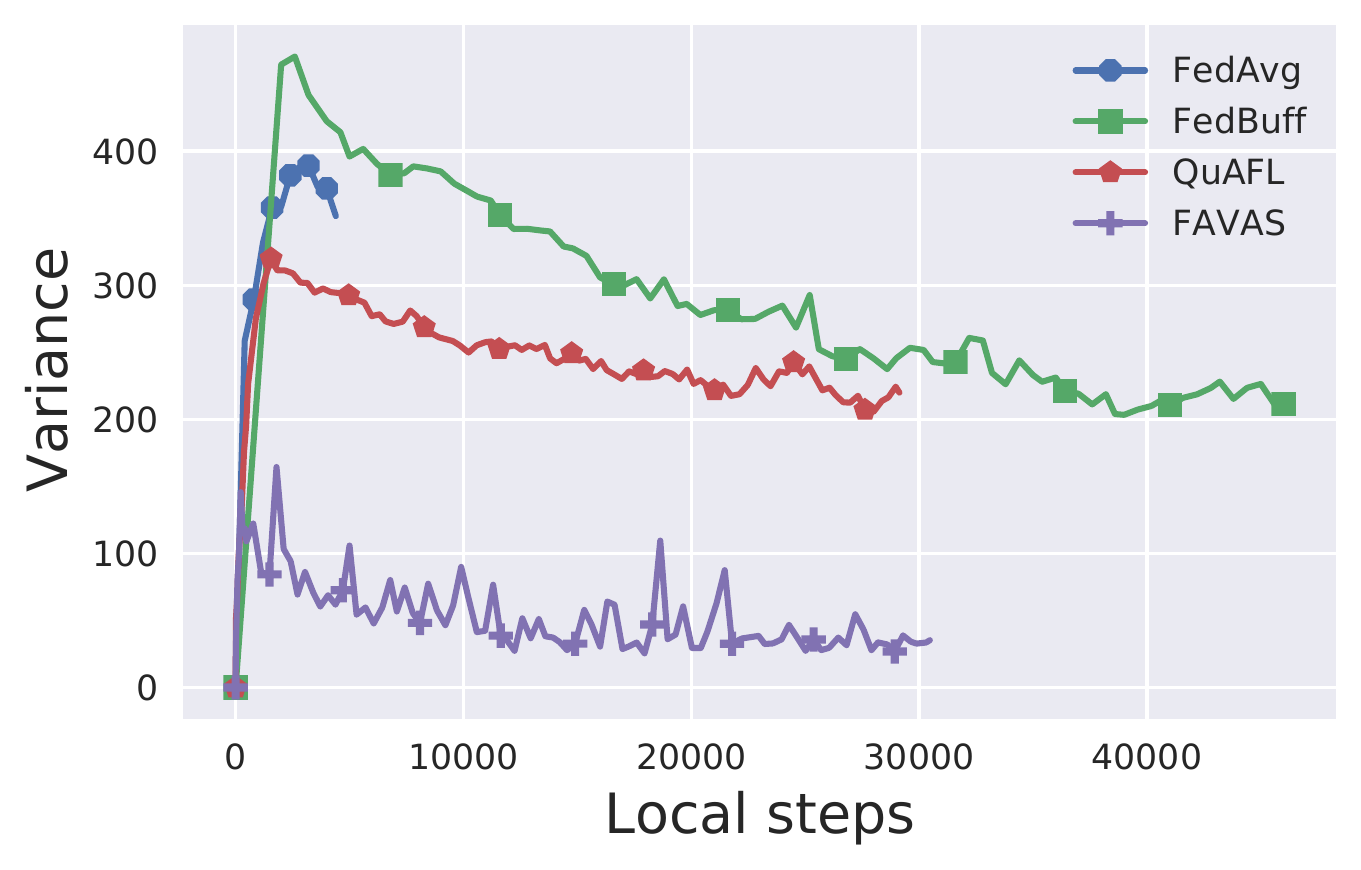}
    \caption{Validation loss/accuracy and variance on the MNIST dataset with a non-iid split in between $n=100$ total nodes. In this particular experiment, one ninth of the clients are defined as fast.}
    \label{fig:appmnist_noniid_slowclients}
\end{figure}
\begin{figure}[H]
    \centering
    \includegraphics[scale=0.47]{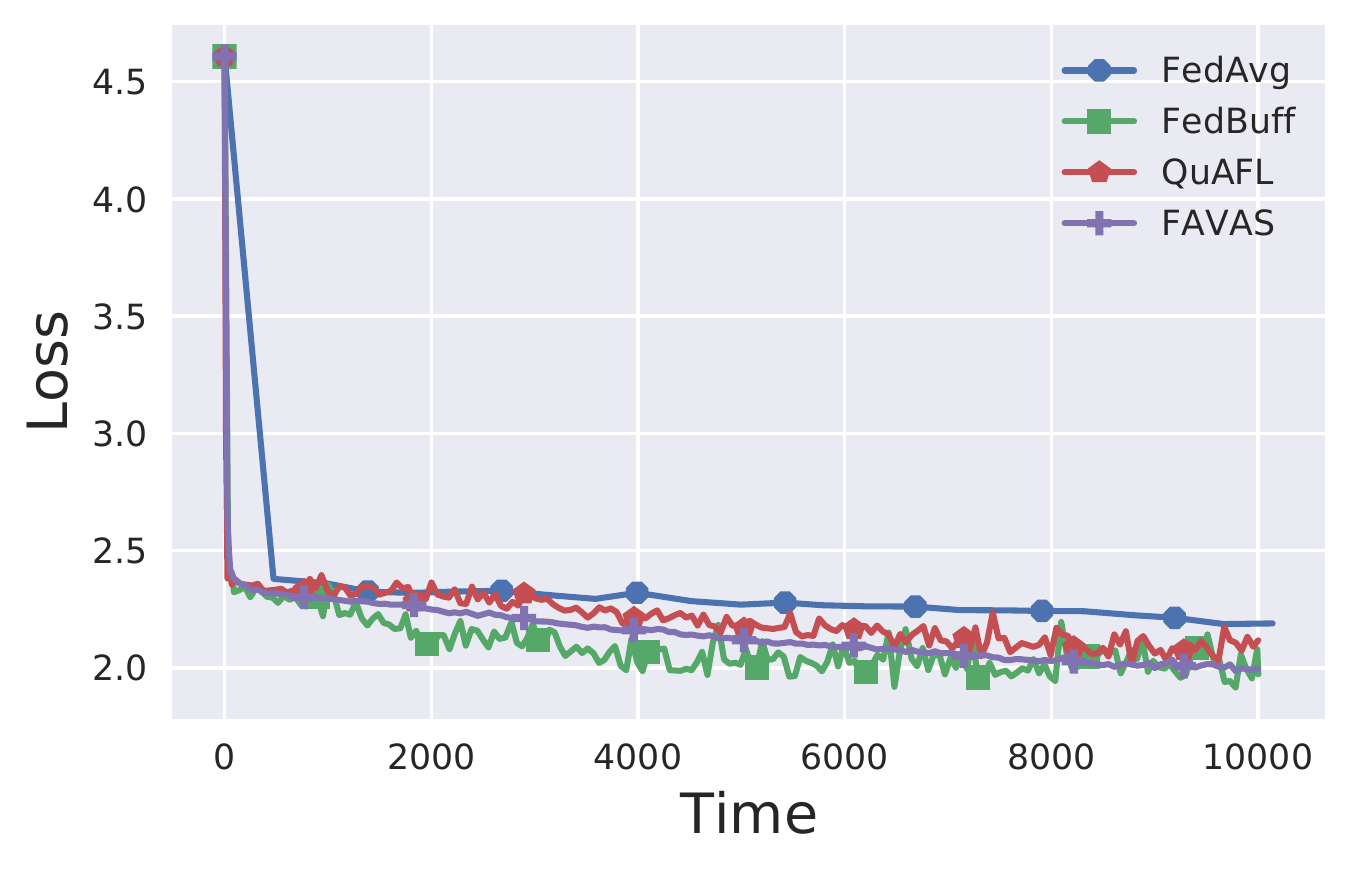}
    \includegraphics[scale=0.47]{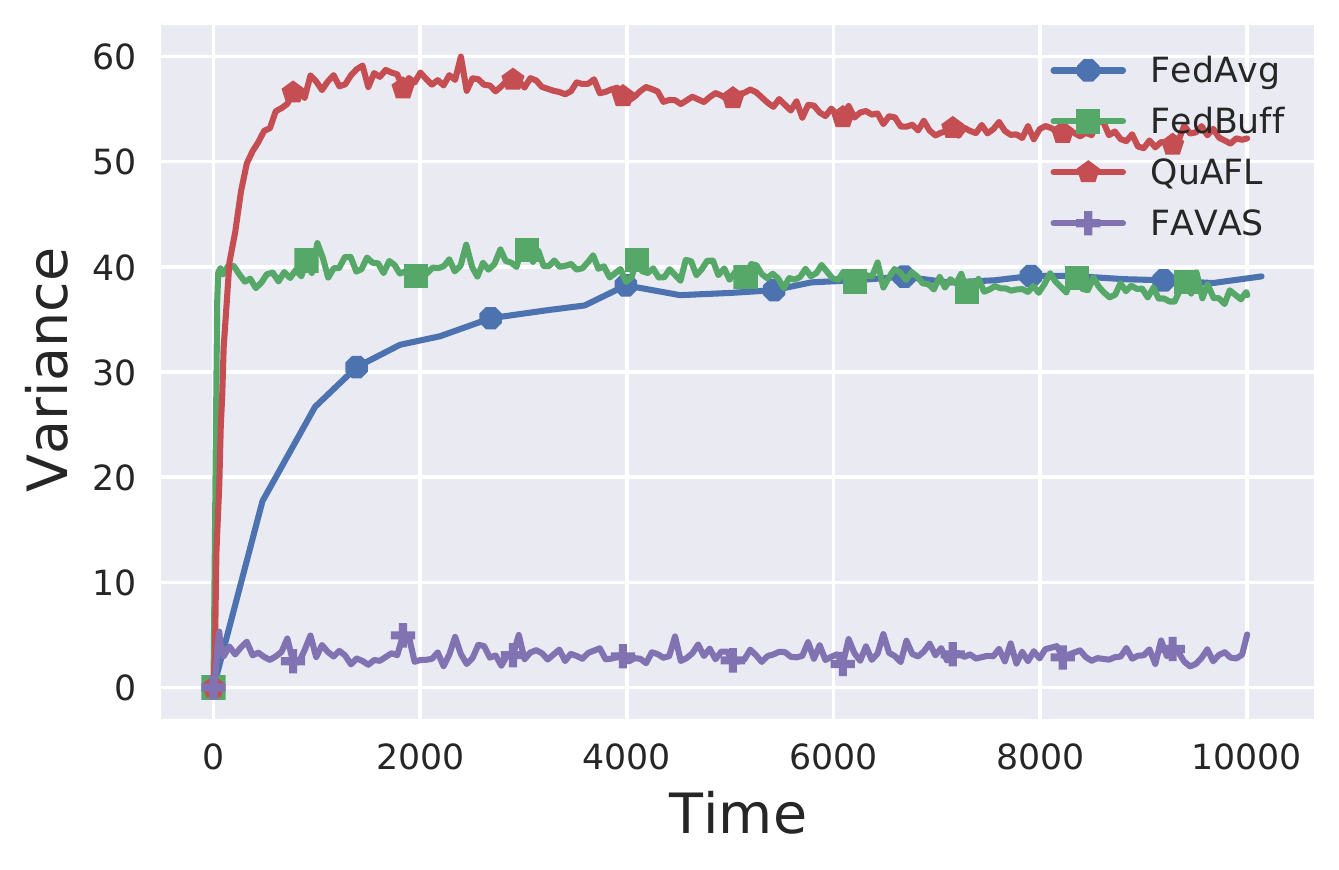}
    \includegraphics[scale=0.47]{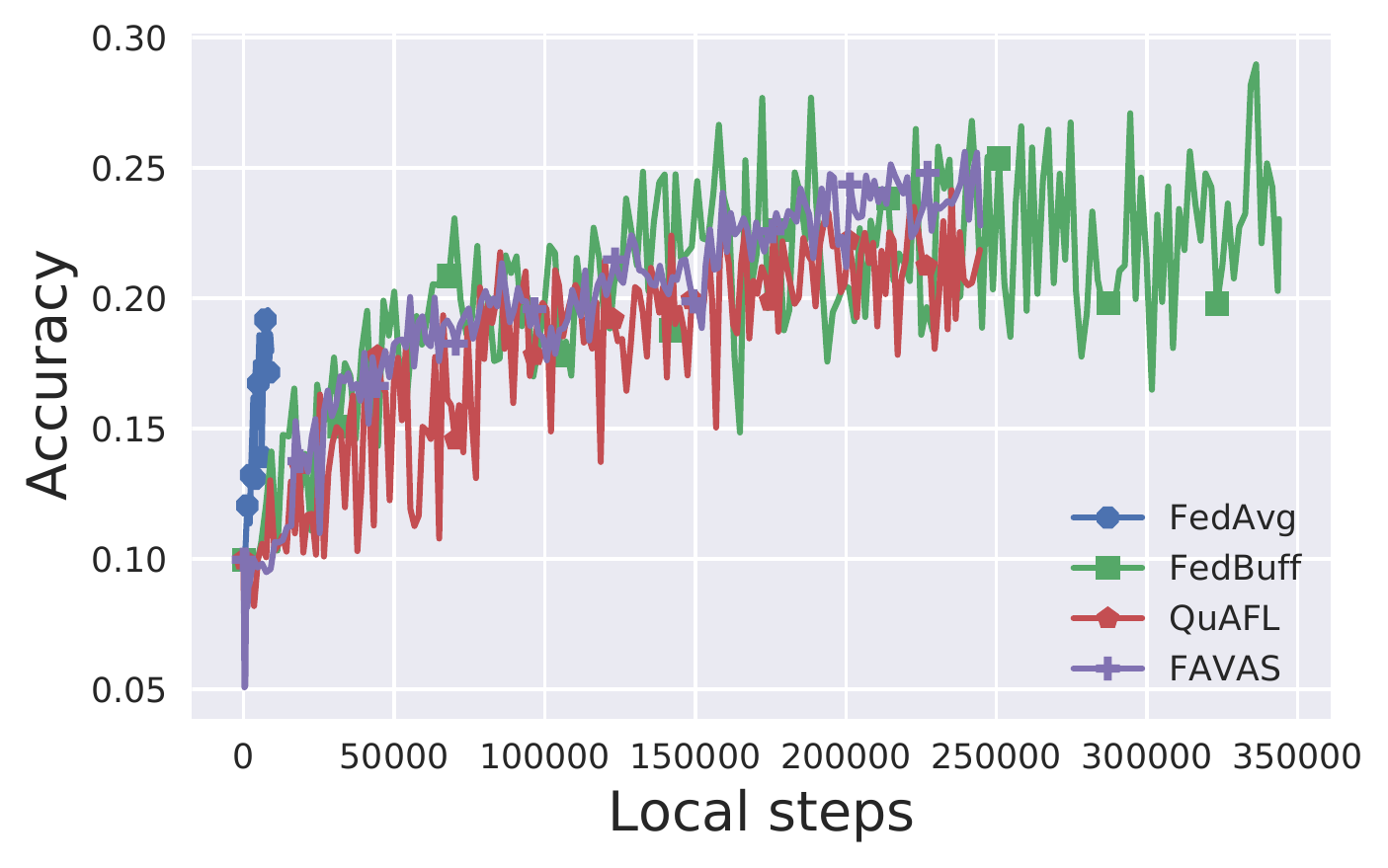}
    \includegraphics[scale=0.47]{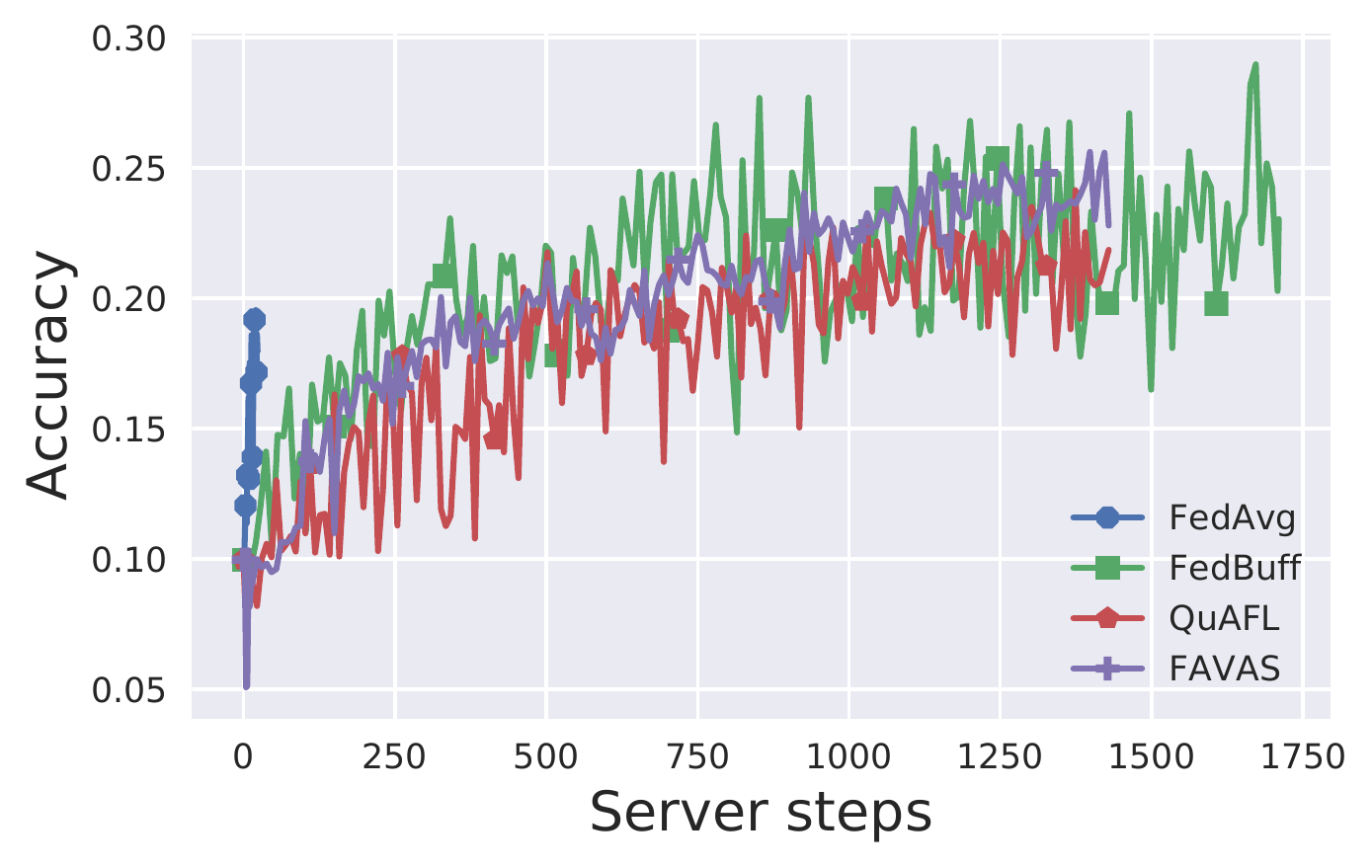}
    \includegraphics[scale=0.47]{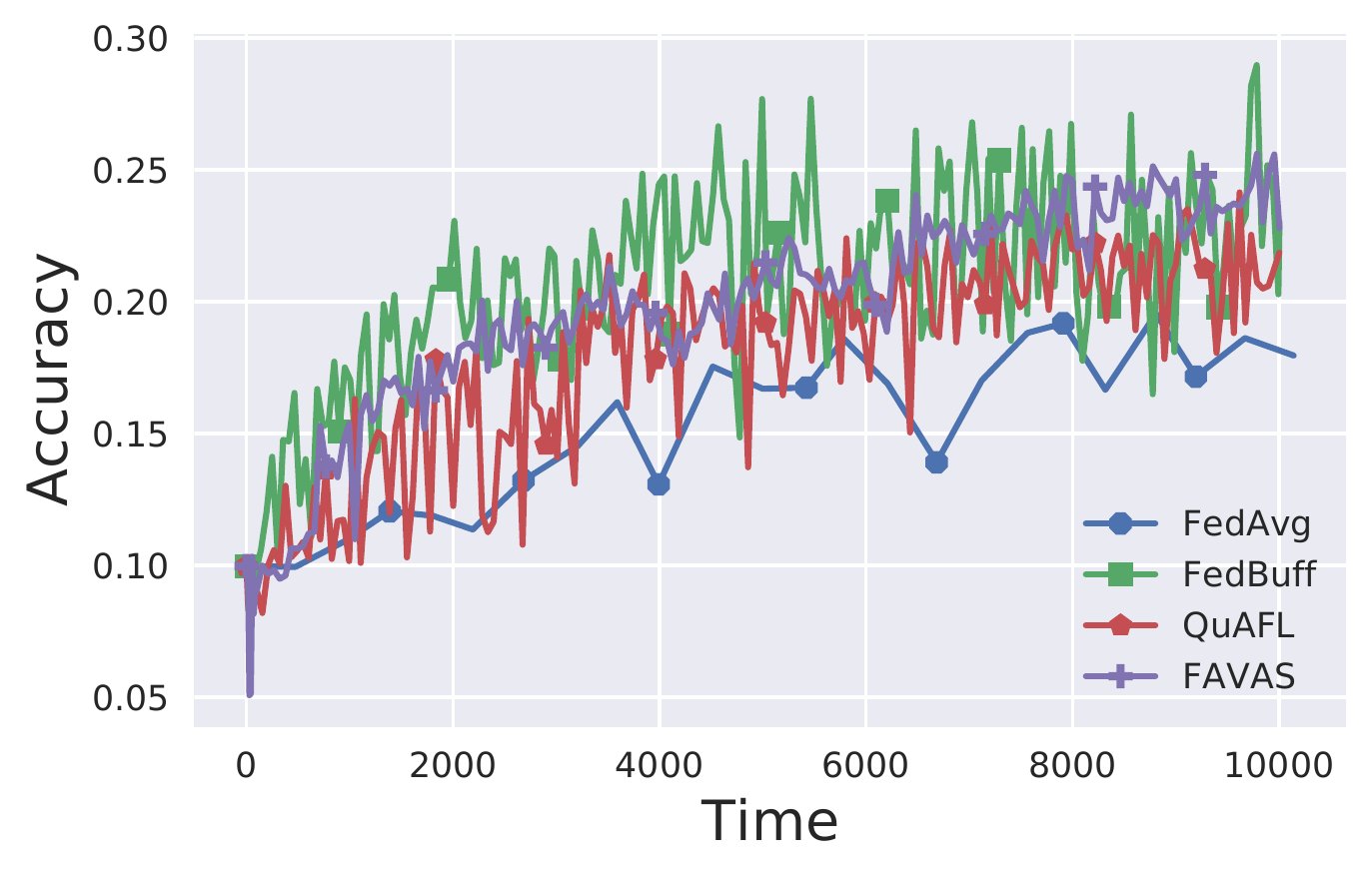}
    \includegraphics[scale=0.47]{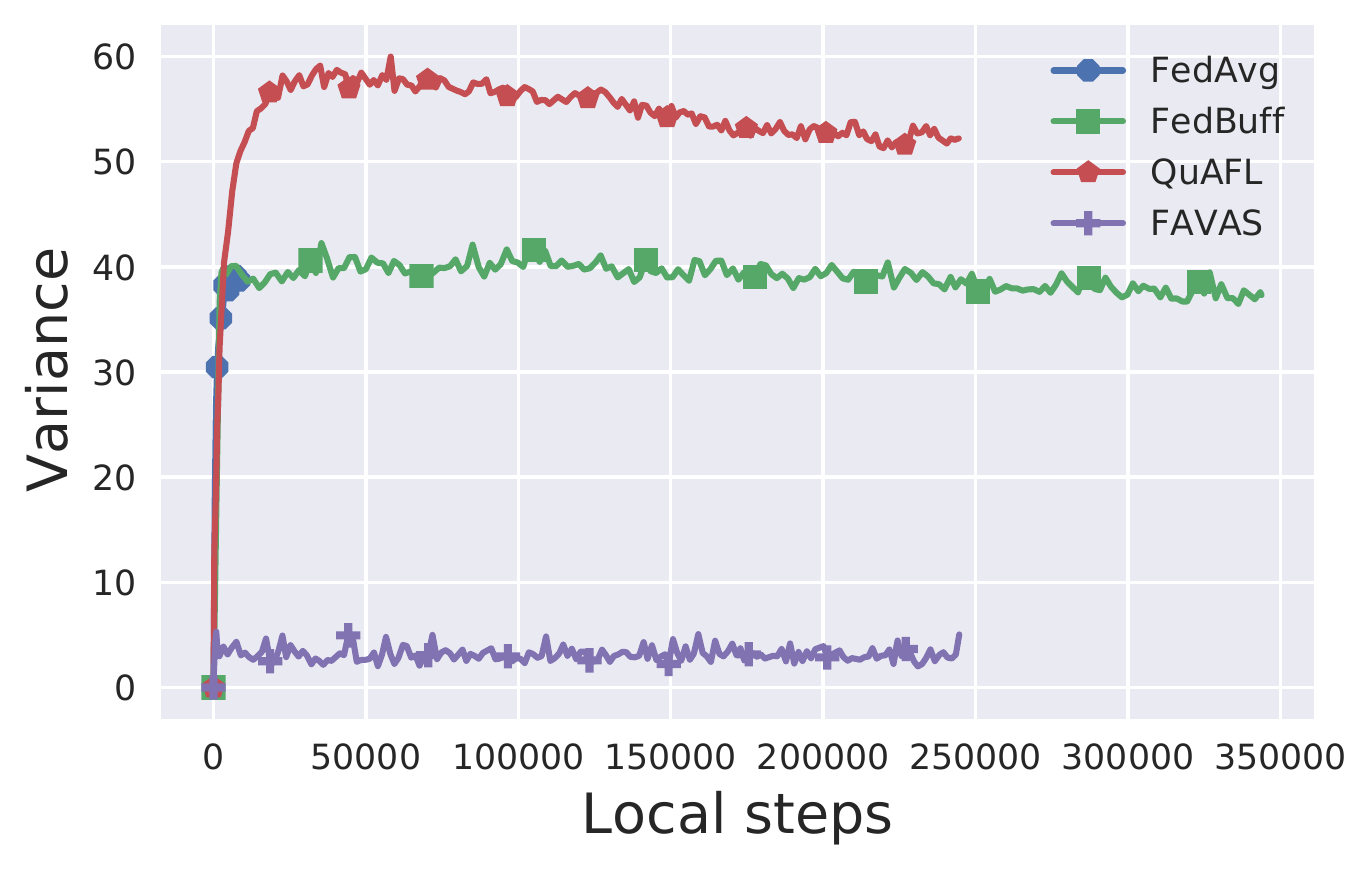}
    \caption{Validation loss/accuracy and variance on the CIFAR-10 dataset with a non-iid split in between $n=100$ total nodes.}
    \label{fig:appcifar10_noniid}
\end{figure}
We refer to \Algo[QNN] when training a neural network with low bit precision arithmetic. We ran the code \footnote{https://openreview.net/forum?id=clwYez4n8e8} from LUQ \cite{chmiel2021logarithmic} and adapted it to our datasets and the FL framework. During \Algo[QNN] training, 3-bits quantization  for weights and activation are used, 4 bits quantization for neural gradients is used.
\begin{figure}[H]
    \centering
    \includegraphics[scale=0.5]{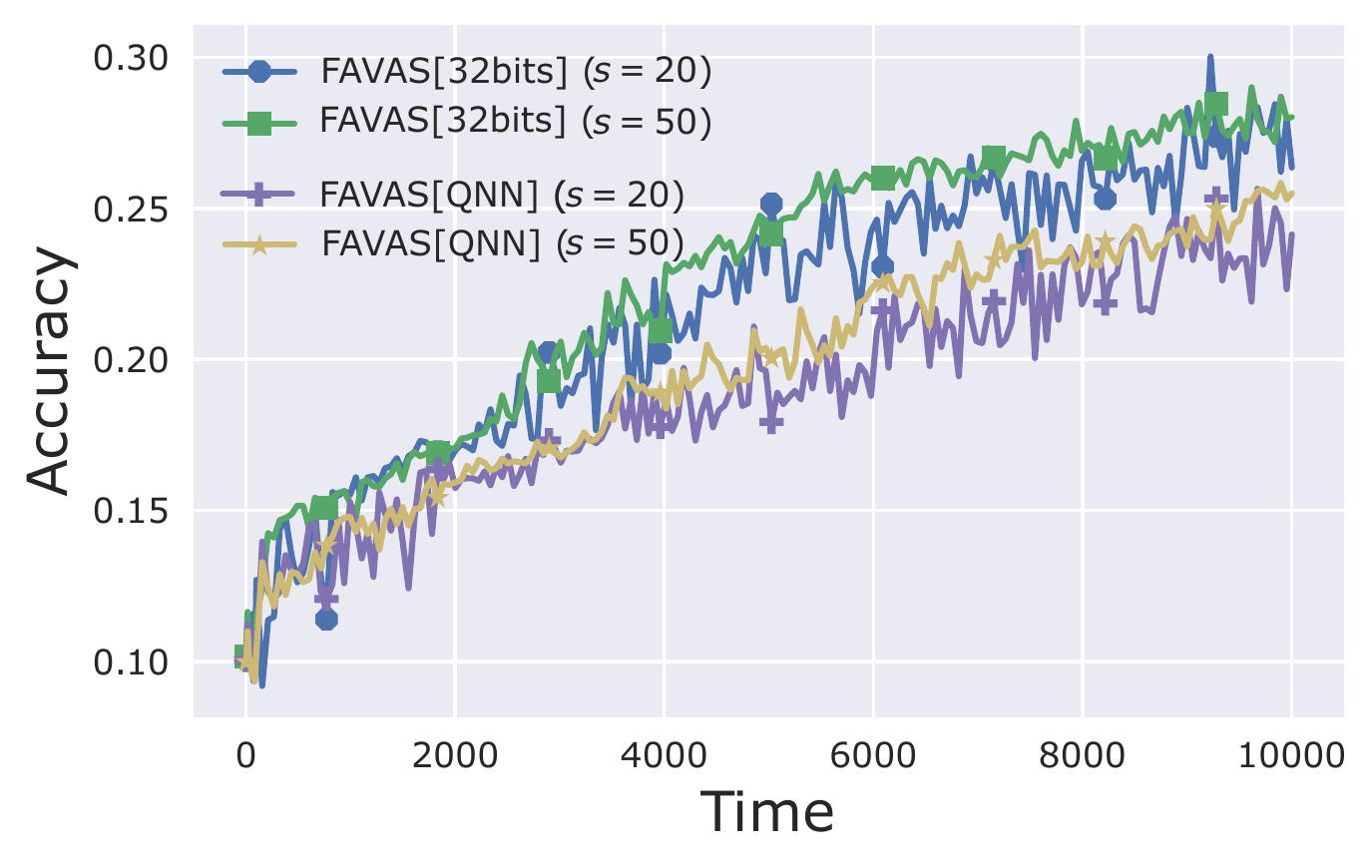}
    \caption{Validation accuracy on the CIFAR-10 dataset with a non-iid split in between $n=100$ total nodes. The amount $s$ of selected clients at each round is varied. \Algo[QNN] is the quantized version of \Algo[32bits].}
    \label{fig:quantiz}
\end{figure}
In \Cref{fig:quantiz} we analyse the effects of quantization and the influence of the number of randomly selected clients $s$ on the convergence behaviour. As expected, we find that higher $s$ improve the performance of \Algo. Quantizing the neural network degrades  the convergence behaviour of the algorithm, but, even if the weights and activation functions are highly quantized - as in the scenario we are considering-, the results are close to their full-precision counterpart.

\end{document}